\documentclass{article}

\PassOptionsToPackage{numbers, compress}{natbib}

\usepackage[preprint]{neurips_2025}

\usepackage[utf8]{inputenc} 
\usepackage[T1]{fontenc}    
\usepackage{textcomp}
\usepackage{hyperref}       
\usepackage{url}            
\usepackage{booktabs}       
\usepackage{amsfonts}       
\usepackage{nicefrac}       
\usepackage{microtype}      
\usepackage{xcolor}         
\usepackage[most]{tcolorbox}

\usepackage{microtype}
\usepackage{graphicx}
\usepackage{subfigure}
\usepackage{booktabs} 
\usepackage{longtable}

\usepackage{amsmath}
\usepackage{amssymb}
\usepackage{mathtools}
\usepackage{amsthm}
\usepackage{color}
\usepackage{bm}
\usepackage{enumitem}
\usepackage{float}
\usepackage{wrapfig}
\usepackage{newunicodechar}
\newunicodechar{Ȧ}{\=A}

\newtcolorbox{takeawaybox}{
  colback=gray!10,    
  colframe=gray!80,   
  arc=2mm,            
  boxrule=0.5pt,     
  left=1mm, right=1mm, top=1mm, bottom=1mm,  
  fonttitle=\bfseries,
  title=Takeaway,
}

\makeatletter
\renewcommand{\@fnsymbol}[1]{\ifcase#1\or*\else\@arabic{#1}\fi} 
\makeatother

\title{The Other Side of the Coin: Unveiling the Downsides of Model Aggregation in Federated Learning from a Layer-peeled Perspective}

\author{
Guogang Zhu\textsuperscript{1}, 
Xuefeng Liu\textsuperscript{1 *}, 
Jianwei Niu\textsuperscript{1}, 
Shaojie Tang\textsuperscript{2},
Xinghao Wu\textsuperscript{1}\\
\textsuperscript{1} State Key Laboratory of Virtual Reality Technology and Systems, \\ School of Computer Science and Engineering, Beihang University, Beijing, China \\ 
\textsuperscript{2} Center for AI Business Innovation, Department of Management Science and Systems, \\ University at Buffalo, Buffalo, New York, USA. \\ 
\texttt{\{buaa\_zgg, liu\_xuefeng, niujianwei, wuxinghao\}@buaa.edu.cn} \\
\texttt{shaojiet@buffalo.edu}
}

\begin{document}

\maketitle

\vspace{-1.0cm}

\begin{abstract}
In federated learning (FL), model aggregation plays a central role in enabling decentralized knowledge sharing.
However, it is often observed that the aggregated model underperforms on local data until after several rounds of local training.
This temporary performance drop can potentially slow down the convergence of the FL model.
Prior work regards this performance drop as an inherent cost of knowledge sharing among clients and does not give it special attention. 
While some studies directly focus on designing techniques to alleviate the issue, its root causes remain poorly understood.
To bridge this gap, we construct a framework that enables layer-peeled analysis of how feature representations evolve during model aggregation in FL. 
It focuses on two key aspects: (1) the intrinsic quality of extracted features, and (2) the alignment between features and their subsequent parameters---both of which are critical to downstream performance.
Using this framework, we first investigate \textbf{how} model aggregation affects internal feature extraction process.
Our analysis reveals that aggregation degrades feature quality and weakens the coupling between intermediate features and subsequent layers, both of which are well shaped during local training.
More importantly, this degradation is not confined to specific layers but progressively accumulates with network depth---a phenomenon we term Cumulative Feature Degradation (CFD).
CFD significantly impairs the quality of penultimate-layer features and weakens their coupling with the classifier, ultimately degrading model performance.
We further revisit several widely adopted solutions through the lens of layer-peeled feature extraction to understand \textbf{why} they are effective in addressing aggregation-induced performance drop.
Our results show that their effectiveness lies in mitigating the feature degradation described above, which is well aligned with our observations.
These findings provide a more interpretable understanding of model aggregation and can potentially offer insight into the design of more effective FL algorithms.

\end{abstract}

\vspace{-0.6cm}
\section{Introduction}
\begin{wrapfigure}{r}{0.6\textwidth}
	\centering
	\vspace{-1.1cm}
	\includegraphics[width=3.3in]{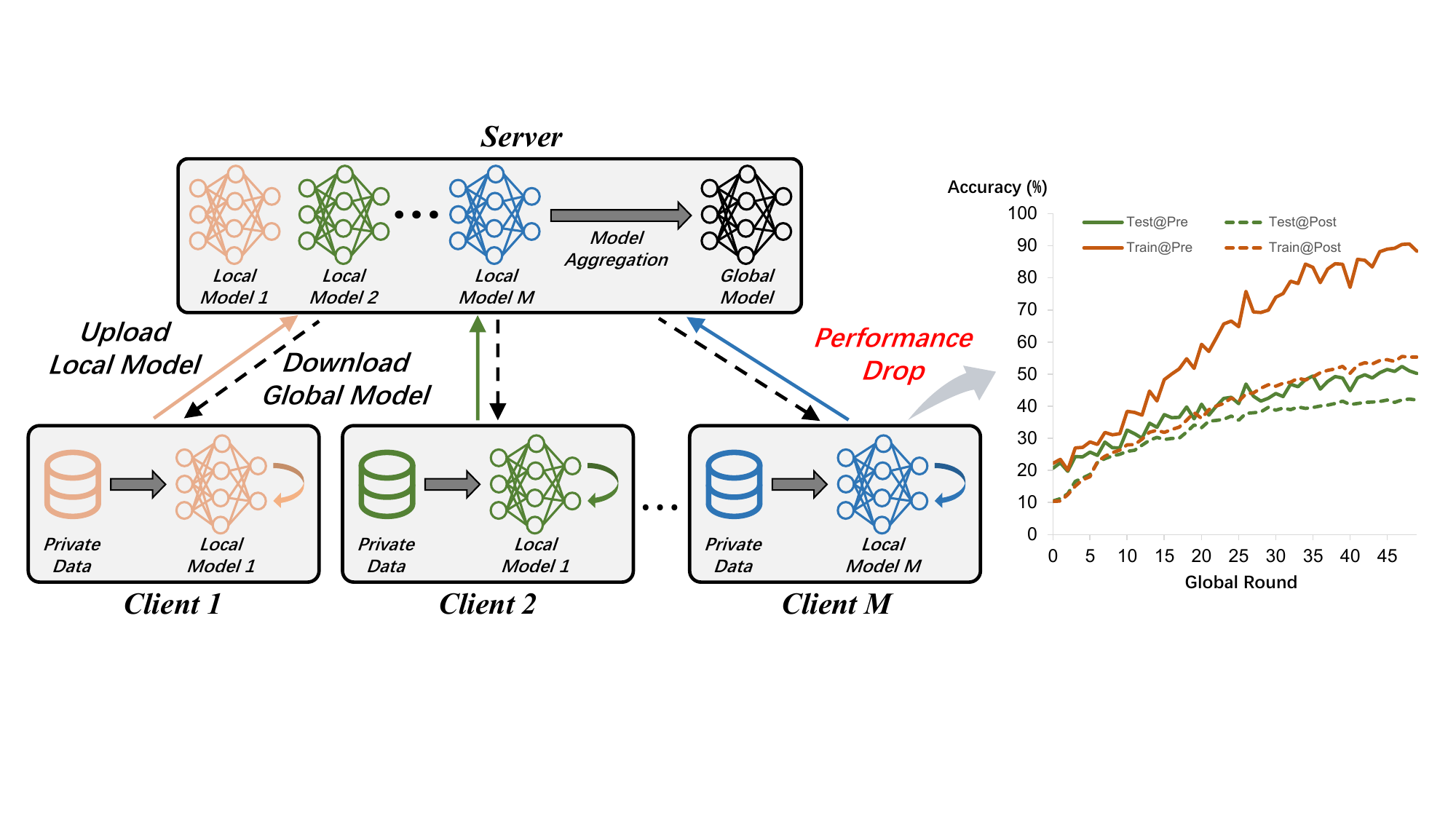}
	\vspace{-0.4cm}
	\caption{\textbf{Left:} Local model training and global model aggregation during FL training. 
		\textbf{Right:} Performance comparison when evaluating models on local data, *@Pre refers to the evaluation results of the model before aggregation, while *@Post indicates the results after aggregation.}
	\vspace{-0.6cm}
	\label{Model_Aggregation}
\end{wrapfigure}

\renewcommand{\thefootnote}{}
\footnotetext{* Corresponding Author.}
\vspace{-0.2cm}
%
%
%
%
Recently, federated learning (FL) has gained increasing interests \citep{kairouz2021advances} since it can enable multiple clients to collaboratively train models without sharing their private data.  
A standard FL process involves iterative cycles in which local models are trained on each client, followed by aggregation of these locally updated models on a central server \citep{FedAvg}, as presented in Figure \ref{Model_Aggregation}. 
During local training, each client performs multiple rounds of model updates using its private data. 
Once local training is complete, the updated model is uploaded to the server. 
The server then aggregates the uploaded models via parameter-wise averaging, with each model weighted based on factors such as the number of training samples on each client \citep{FedDyn,FedProx,SCAFFOLD}. 
The aggregated model is then sent back to each client for next round of local training.

In the above process, model aggregation is a key step that facilitates knowledge sharing among clients in FL.  
However, it is well known that the model aggregation often leads to a significant performance drop compared to the model before aggregation \citep{pFedSD,FedNTD,pFedAMF}.
This phenomenon is particularly pronounced when data distributions across clients are heterogeneous, a common scenario in practical applications due to factors such as variations in data acquisition conditions across different clients \citep{zhu2021federated,FedBN}.
To further investigate this phenomenon, we conduct preliminary experiments in a typical data-heterogeneous FL setting.
Figure \ref{Model_Aggregation} presents a performance comparison of the model before and after aggregation, evaluated on the local data from each client.
As shown, the performance of the aggregated model significantly deteriorates compared to the model before aggregation.

Although this temporary performance drop can be mitigated after several rounds of local updates, its impact persists and continues to pose a challenge.
The suboptimal initialization in each local update, caused by model aggregation, undermines the progress made in the previous round, potentially slowing the convergence rate of FL training.
Most FL research treats this performance drop as an inherent cost of knowledge sharing among clients, giving it limited attention \citep{FedAvg, FedCAC, MOON}.
Recently, several studies attributes this phenomenon to `client drift' \citep{FedProx, SCAFFOLD} and `knowledge forgetting' \citep{pFedSD, FedNTD}.
To address this issue, methods such as partial parameter personalization \citep{FedPer,FedBN,LG-FedAvg,PartialFed}, pre-trained initialization \citep{nguyen2023where,chen2023on}, and classifier fine-tuning \citep{FedBABU,FedETF} are proposed.

While the above methods have achieved notable success in mitigating the performance degradation caused by model aggregation---mainly as measured by metrics such as accuracy or loss—the fundamental causes of this issue remain insufficiently understood.
This gap can be largely attributed to the dominant use of deep neural networks (DNNs) in FL \cite{FedBN,chen2023on,FedETF,FedBABU}, which are often treated as black-box models \cite{montavon2018methods,samek2021explaining} composed of stacked layers performing hierarchical feature extraction\cite{yosinski2014transferable,olah2017feature,masarczyk2024tunnel,wang2023understanding,zeiler2014visualizing,rangamani2023feature}. 
Such architectures hinder the interpretation of internal feature representation dynamics that potentially underlie aggregation-induced degradation.
Although some recent studies have attempted to analyze FL from a layer-peeled perspective \citep{CCVR,chan2024internal,adilova2024layerwise}, they either focus on parameter-space analysis using loss-based metrics \citep{chan2024internal}, or rely on feature similarity metrics that require two models for comparison \cite{CCVR,adilova2024layerwise}.
To the best of our knowledge, no existing work analyzes the dynamics of layer-peeled feature extraction using metrics that can be directly computed using a single model to quantify either feature quality or feature-parameter alignment.

To bridge this gap, we construct a framework to investigate model aggregation from a layer-peeled feature extraction perspective. 
We hypothesize that the model performance fundamentally depends on two key factors: (1) the intrinsic quality of the extracted features, and (2) the degree of alignment between features and the parameters of subsequent layers. 
The first factor determines whether the features are semantically meaningful and discriminative, while the second determines whether they can be effectively exploited by the model’s decision-making or downstream feature extraction modules.
Accordingly, our framework integrates quantitative metrics to evaluate both aspects, providing a more interpretable understanding of aggregation effects in FL.

Based on this framework, we perform a layer-peeled feature analysis of aggregation across multiple datasets and model architectures.
In doing so, we aim to answer the following key questions:
\vspace{-.3cm}
\begin{itemize}
    \item \textbf{How model aggregation affects feature representations?}  
    We observe that model aggregation generally compromises the quality of extracted features and weakens the alignment between features and subsequent parameters.
    More importantly, this degradation is not confined to specific layers but accumulates progressively with network depth---a phenomenon we term Cumulative Feature Degradation (CFD).
    We identify CFD as a fundamental factor contributing to the performance drop when aggregating DNNs.
    \item \textbf{Why are common solutions effective?} We revisit several widely used solutions for mitigating the effects of model aggregation, including personalizing specific parameters, initializing models with pretrained parameters, and fine-tuning the classifier.
    Our analysis shows that these methods are effective because they can address the feature degradation issues describe above.
    This further validates the utility of our layer-peeled analysis framework.
\end{itemize}

This study provide a comprehensive study of model aggregation in FL from a layer-peeled feature extraction perspective.
The key findings can inspire the design of more effective FL algorithms.

\section{Problem Formulation of FL from a Layer-Peeled Perspective}
\vspace{-0.4cm}
In this paper, we consider a standard FL system consisting of a central server and $M$ distributed clients. 
We assume that each client $m$ contains $N_m$ training samples, which are drawn from the data distribution $\mathcal{D}_m$. 
In practice, the underlying data distribution $\mathcal{D}_m$ for each client $m$ is typically different from one another due to the variations in data collection conditions. 
Formally, the training samples on client $m$ can be represented as $(\bm{x}_m^i, \bm{y}_m^i)_{i = 1}^{N_m}$, where $\bm{x}_m^i \in \mathcal{X}_m \subseteq \mathbb{R}^n$ denotes the raw input data for the DNNs, and $\bm{y}_m^i \in \mathcal{Y}_m \subseteq \{0, 1\}^C$ represents the corresponding ground truth labels used to optimize the DNNs, with $C$ denoting the number of classes.

We denote the DNN for client $m$ as $\bm{\psi}_m(\cdot)$, with parameters represented by $\bm{\Theta}_m$.
The optimization objective for an FL system can then be formulated as:
\begin{equation}
\label{FL}
\text{arg} \underset{\bm{\Theta}_1, \ldots, \bm{\Theta}_M}{\text{min}} \ \mathcal{L}(\bm{\Theta}_1, \ldots, \bm{\Theta}_M) \triangleq \text{arg} \underset{\bm{\Theta}_1, \ldots, \bm{\Theta}_M}{\text{min}} \frac{1}{M} \sum_{m=1}^{M}\mathcal{L}_m(\bm{\Theta}_m),
\end{equation}
where $\mathcal{L}_m(\bm{\Theta}_m)$ represents the empirical risk for client \( m \), which is  computed based on its private data samples, as shown in the following equation:
\begin{equation}
\label{empirical-risk}
\mathcal{L}_m(\bm{\Theta}_m)  := \frac{1}{N_m} \sum_{i=1}^{N_m} \ell (\bm{y}_m^i, \hat{\bm{y}}_m^i),
\end{equation}
where $\hat{\bm{y}}_m^i = \psi_m(\bm{x}_m^i;\bm{\Theta}_m)$ represents the predicted output of $\bm{x}_m^i$ given the model $\psi_m(\cdot)$ for $\bm{x}_m^i$, $\ell:\mathcal{Y} \times \mathcal{Y} \rightarrow \mathbb{R}$ is the loss function used to measure the prediction error. 

To optimize Equation (\ref{FL}) in a privacy-preserving manner, FL is typically carried out in two iterative stages: local model training and global model aggregation.
During the local model training phase, each client optimizes its model for $E$ epochs by minimizing the loss function defined in Equation (\ref{empirical-risk}).
Once local training is complete, each client uploads its updated model to the server.
The server then performs model aggregation to generate the global model. A common aggregation strategy involves applying a weighted average for each parameter within the model based on the number of training samples per client, which is expressed as follows:
\begin{equation} \label{global-aggregation} \tilde{\bm{\Theta}} = \sum_{m=1}^{M} \frac{N_m}{\sum_{m=1}^M N_m}\bm{\Theta}_m \end{equation}
Here, $\tilde{\bm{\Theta}}$ represents the aggregated global model. By repeating the above procedures for several rounds, the model eventually converges, resulting in the final FL model. For simplicity, we will sometimes refer to the locally updated model $\bm{\Theta}_m$ as the \textbf{\textit{pre-aggregated model}}, and $\tilde{\bm{\Theta}}$ as the \textit{\textbf{post-aggregated model}}. Additionally, we may omit the client and sample indices for simplicity.

It is well known that post-aggregated models often suffer a significant performance drop when evaluated on local data.
We provide detailed evidence of this phenomenon in Appendix~\ref{performance_drop_appendix}.
To better understand its underlying causes, we propose a layer-peeled feature analysis framework to investigate how feature extraction evolves across model depth during aggregation.
Within this framework,  we reformulate the parameters of the FL model as $\boldsymbol{\Theta}=\left\{\boldsymbol{W}^{\ell}\right\}_{\ell=1}^L$, where $L$ denotes the total number of layers, and the depth increases with $\ell$. 
This stacked layers progressively transforms the input data into prediction outputs, from the shallow layers to the deeper layers. 
The features of the $\ell$-th layer for input $\bm{x}$ can be formulated as: 
\begin{equation}
    \boldsymbol{z}^{\ell}=\boldsymbol{W}^{\ell} \ldots \boldsymbol{W}^{1} \boldsymbol{x}=\boldsymbol{W}^{\ell: 1} \bm{x}, \forall \ell=1, \ldots, L-1
\end{equation}
where $\boldsymbol{z}^{\ell}$ denotes the intermidiated features of $\ell$-th layer and we define $\bm{z}^0=\bm{x}$. 

\vspace{-0.5cm}
\section{Evaluation Setup}\label{evaluation_setup}
\vspace{-0.3cm}

\subsection{Implementation Details}
\vspace{-0.3cm}
    \textbf{Datasets.} We conduct experiments on several cross-domain datasets, including Digit-Five, PACS \citep{PACS}, and DomainNet \citep{DomainNet}.
    For each dataset, samples from a single domain are assigned to an individual client.
    Details on dataset preprocessing and partitioning are provided in Appendix~\ref{appendix_dataset_description}.

    \textbf{Model Architectures.}
    Our evaluation are conducted on various architectures, including convolutional networks (ConvNet), three variants of ResNet (ResNet18, ResNet34, and ResNet50), VGG13\_BN \citep{VGG} \citep{ResNet}, and ViT\_B-16 \citep{ViT}. 
    All models are trained from random initialization unless otherwise specified. 
    Detailed descriptions of architectures are provided in Appendix \ref{appendix_model_architecture}.

    \textbf{Training Settings.}
    We adopt the standard FedAvg algorithm \citep{FedAvg} for federated training.
    Model optimization is performed using stochastic gradient descent (SGD) with a learning rate of 0.01, momentum of 0.5, and a batch size of 64.
    Each client trains locally for 10 epochs per global round, and the training runs for a total of 50 global rounds.
    On the server side, model aggregation is conducted via weighted averaging of parameters, with weights proportional to the number of training samples on each client.
    To account for randomness, all experiments are repeated three times with different random seeds.
    Further implementation details are provided in Appendix~\ref{implementation_details_appendix}.

\vspace{-0.3cm}
\subsection{Feature Evaluation Metrics}
\vspace{-0.3cm}
   
    Our feature evaluation framework incorporates a set of metrics designed to evaluate both the quality of extracted features (feature vairance \citep{wang2023understanding,rangamani2023feature} and linear probing accuracy \citep{chen2020simple, he2022masked, wang2023does}) and their alignment \citep{rangamani2023feature, jordan1875essai, bjorck1973numerical} with subsequent model parameters at a given layer $\ell$.
    To quantify changes in feature extraction before and after aggregation, we additionally introduce the pairwise distance between features or models and the relative change of evaluated metrics.
    A brief overview of these metrics is provided below.
    The detailed definitions can be found in Appendix \ref{appendix_metric}.

\begin{enumerate}[label=\textbullet] 
\item \textbf{Feature Variance.}  We calculate the normalized within-class and between-class variance to assess the discrimination of extracted features. 
The within-class variance quantifies the compactness of features belonging to the same class, where a lower value indicates higher intra-class consistency. 
In contrast, the between-class variance reflects the degree of separation among different classes, with higher values suggesting better inter-class separability.

\item \textbf{Accuracy of Linear Probing.}  This metric evaluates the generalization of features across different distributions. We randomly initialize a linear classifier on top of the extracted features, train it on the target dataset, and report the average test accuracy.

\item \textbf{Alignment between Features and Parameters.}  This metric is used to evaluate the degree of coupling between features and their subsequent parameters.
A higher value indicates stronger alignment between the features and the corresponding parameters.

\item \textbf{Relative Change of Evaluated Metrics.} It quantifies the metric variations during aggregation, providing insights into the sensitivity of the extracted features to aggregation. 
\item \textbf{Pairwise Distance of Features or Models.} It evaluates the differences between the parameters or extracted features of pre-aggregated and post-aggregated models.
\end{enumerate}

\vspace{-0.6cm}
\section{How Model Aggregation Affects Feature Representations?}
\vspace{-0.3cm}

\subsection{Model Aggregation Disrupts Feature Variance Suppression}\label{feature_suppression}
\vspace{-0.2cm}

\textbf{Motivation} In this section, we investigate how feature discrimination evolves during model aggregation, as it plays a crucial role in determining model performance on local client data. To this end, we measure both within-class and between-class variances to quantify the degree of intra-class feature compactness and inter-class featre separability.

\begin{figure}[!htbp]
\centering
\includegraphics[width=\textwidth]{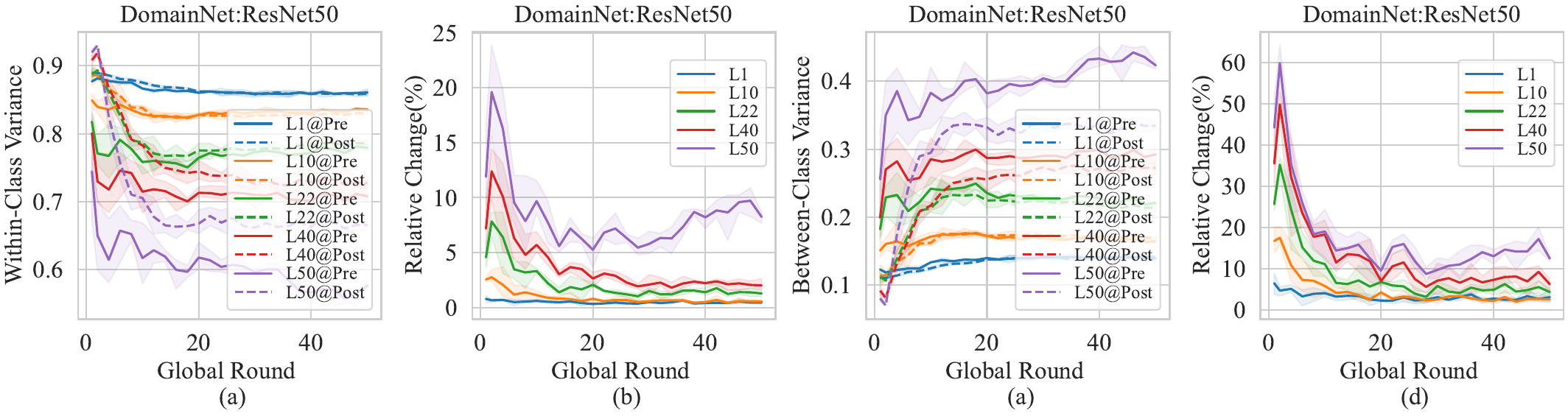}
\vspace{-0.8cm}
\caption{Normalized within-class and between-class feature variances at different layers during FL training. The model is trained on DomainNet using ResNet50: (a) normalized within-class variance, (b) relative change of within-class variance, (c) normalized between-class variance, (d) relative change of between-class variance.}
\vspace{-0.4cm}
\label{FedAvg_ResNet50_Layerwise_NC1_NC1_Between}
\end{figure}

\textbf{Experimental Results} Figures \ref{FedAvg_ResNet50_Layerwise_NC1_NC1_Between} and \ref{FedAvg_ResNet50_Epochwise_NC1_NC1_Between} presents the evaluation resutls on DomainNet using ResNet50 as the backbone. 
These figures reveal several key observations:

\begin{figure}[!htbp]
\centering
\includegraphics[width=\textwidth]{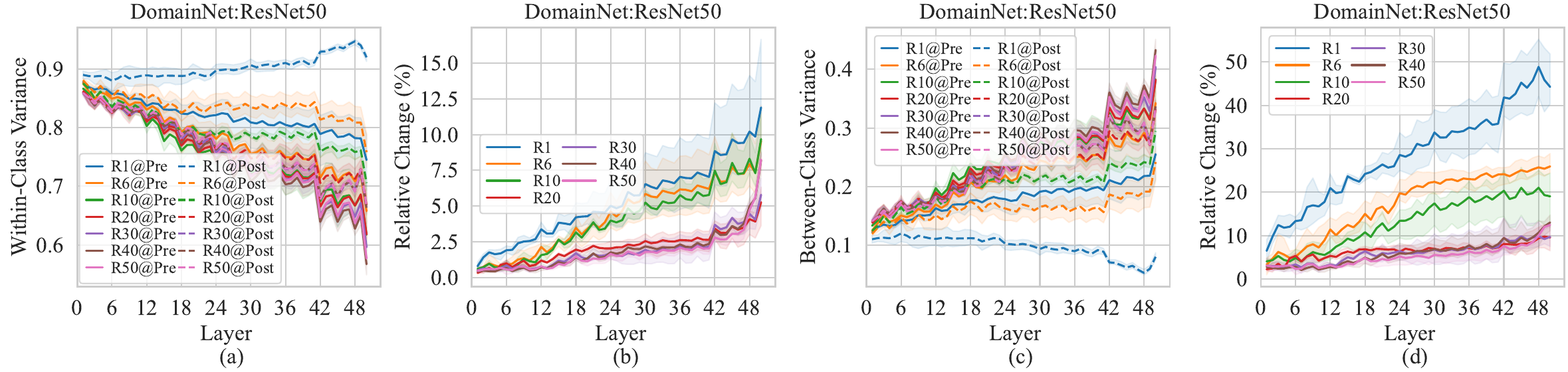}
\vspace{-0.8cm}
\caption{Normalized within-class and between-class feature variances across layers for specific global rounds. The model is trained on DomainNet using ResNet50: (a) normalized within-class variance, (b) relative change of within-class variance, (c) normalized between-class variance, (d) relative change of between-class variance.}
\vspace{-0.8cm}
\label{FedAvg_ResNet50_Epochwise_NC1_NC1_Between}
\end{figure}

\textbf{(1) Features become increasingly compressed within the same class as training progresses and layer depth increases.} 
As shown in Figure \ref{FedAvg_ResNet50_Layerwise_NC1_NC1_Between} (a), from a temporal perspective, the within-class feature variance at a given layer progressively decreases as federated training proceeds.
Moreover, as depicted in Figure \ref{FedAvg_ResNet50_Epochwise_NC1_NC1_Between} (a), from a spatial perspective, at a given training epoch, the within-class feature variance decreases with increasing network depth.
These observations collectively indicate that the within-class feature variance decreases consistently as training progresses and the model layer goes deeper.
\vspace{-0.1cm}
\begin{wrapfigure}{r}{0.5\textwidth}
\centering
\includegraphics[width=2.7in]{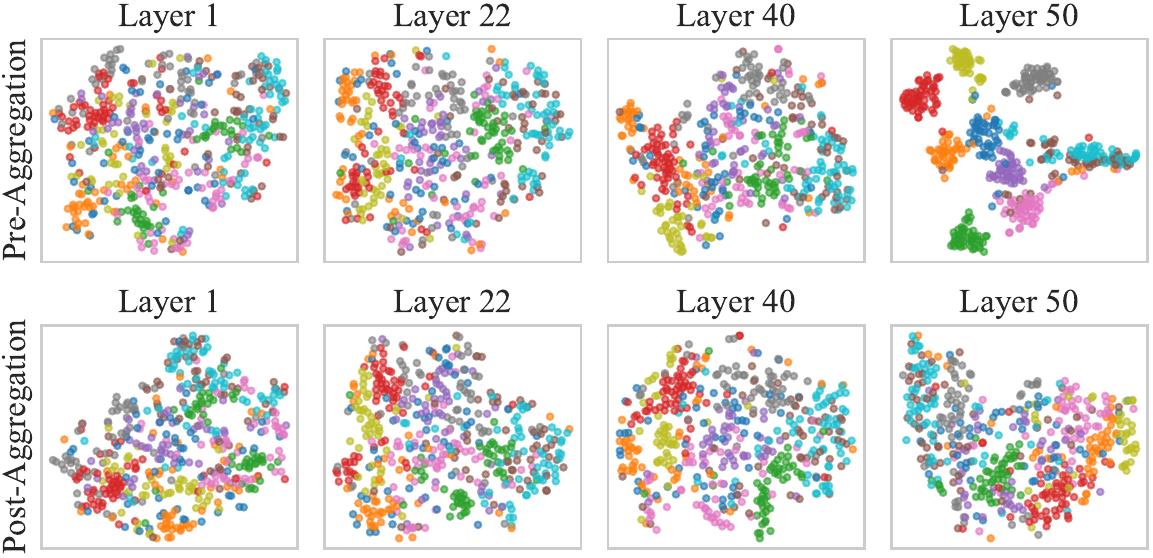}
\vspace{-0.2cm}
\caption{T-SNE \cite{TSNE} visualization of features at different layers on the `Quickdraw' domain of DomainNet. The features are extracted from ResNet50 in the final global round.}
\vspace{-0.5cm}
\label{TSNE_DomainNet_quickdraw_ResNet50_4Layers}
\end{wrapfigure}

\textbf{(2) Features become increasingly discriminative across different class as training progresses and layer depth increases.} 
As shown in Figure \ref{FedAvg_ResNet50_Layerwise_NC1_NC1_Between} (a), from a temporal perspective, the normalized between-class feature variance at a given layer increases progressively over global training rounds.
Similarly, as depicted in Figure \ref{FedAvg_ResNet50_Epochwise_NC1_NC1_Between} (a), from a spatial perspective, at a given global round, deeper layers exhibit higher between-class variance.
These observations collectively indicate that the model progressively enhances feature separability across classes over both training time and network depth.
Furthermore, prior work in centralized learning (CL) show that features become increasingly compact within classes and more separable across classes as training progresses and depth increases \citep{wang2023understanding,rangamani2023feature}.
Together with Observation (1), our findings reveal that FL exhibits similar training dynamics from a layer-peeled feature extraction perspective, despite its decentralized nature.

\textbf{(3) Model aggregation disrupts feature variance suppression during FL training.} As shown in Figures \ref{FedAvg_ResNet50_Layerwise_NC1_NC1_Between} (a) and \ref{FedAvg_ResNet50_Epochwise_NC1_NC1_Between} (a), both temporally and spatially, model aggregation leads to increased within-class variance and decreased between-class variance. 
This opposes the training tendency of DNNs---namely, to reduce within-class variance and increase between-class variance---which has been empirically verified in Observations 1 and 2. Further supporting evidence is provided in Figure \ref{TSNE_DomainNet_quickdraw_ResNet50_4Layers}, where the visualizations reveal degraded feature discrimination after aggregation across layers.
These findings suggest that, from the perspective of feature variance on local data, model aggregation inherently contradicts the training objective and can hinder the convergence rate of FL.

\begin{takeawaybox}
    While FL shares similar training dynamics with CL---promoting within-class feature compactness and between-class feature separability---model aggregation can counteract this progression and ultimately impede FL convergence.
\end{takeawaybox}

\vspace{-0.45cm}
\subsection{Feature Vairance Disruption Accumulate as Model Depth Increases}\label{feature_variation_accumulation}
\vspace{-0.25cm}
\textbf{Motivation} In this section, we investigate the relative change in feature variance induced by model aggregation during FL training. The goal is to assess the sensitivity of different layers to aggregation and to reveal how the stacked architecture of DNNs affects feature variance in this process.

\textbf{Experimental Results} 
The relative changes in normalized within-class and between-class variance are shown in the Figures \ref{FedAvg_ResNet50_Layerwise_NC1_NC1_Between} and \ref{FedAvg_ResNet50_Epochwise_NC1_NC1_Between}. From these figures, we make the following observations:

\begin{wrapfigure}{r}{0.5\textwidth}
\centering
\includegraphics[width=2.8in]{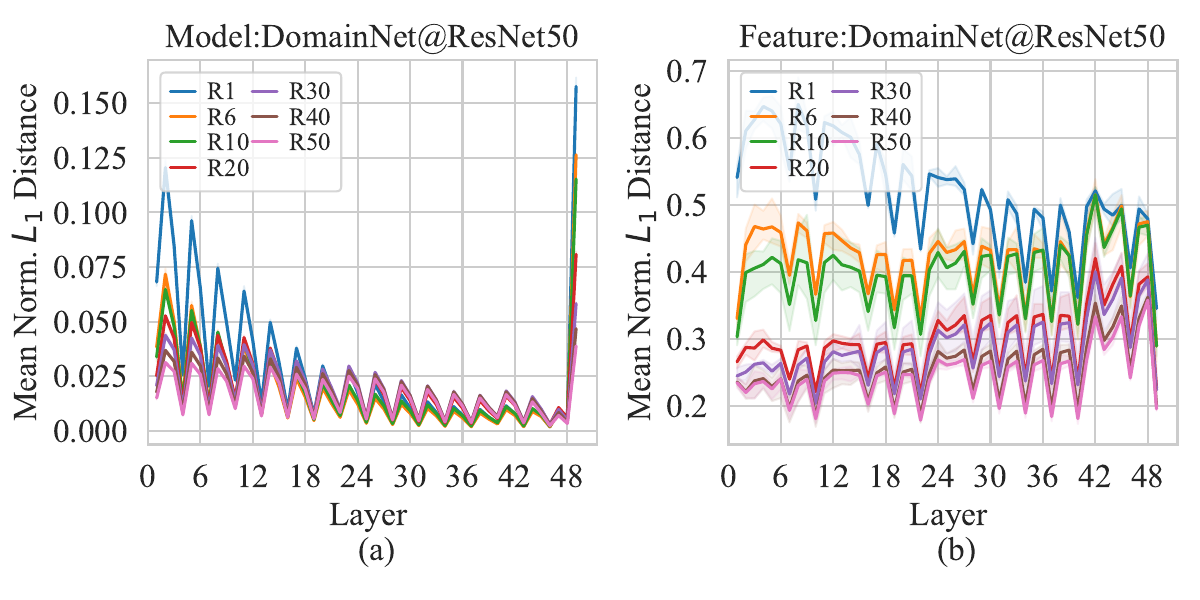}
\vspace{-0.8cm}
\caption{Mean normalized $L_1$ distance of the features and parameters between pre-aggregated and post-aggregated models across model layers for specific global rounds. The model is trained on DomainNet using ResNet50: (a) distance of model parameters (b) distance of features.
\vspace{-0.4cm}
}
\label{FedAvg_Distance_DomainNet_ResNet50_norm_abs_dis_Epochwise}
\end{wrapfigure}

\textbf{(1) Feature variance degradation accumulates with increasing model depth.}
The relative changes in both normalized within-class and between-class variance increase with network depth, indicating that feature variance degradation becomes more severe in deeper layers.
We refer to this phenomenon as Cumulative Feature Degradation (CFD).
Since this degradation is specifically defined in terms of feature variance, we denote it as CFD-V to distinguish it from CFD observed on other metrics.
CFD-V can be attributed to the stacked architecture of DNNs, where degraded features in early layers propagate to subsequent layers, compounding the disruption.
To further validate this, we compute layer-wise parameter and feature distances between pre-aggregated and post-aggregated models.
As shown in Figure \ref{FedAvg_Distance_DomainNet_ResNet50_norm_abs_dis_Epochwise}, the magnitude of parameter distance is consistently smaller than that of feature distance. Moreover, parameter distance decreases with depth---except for the final classifier—while feature distance increases steadily across layers.
This suggests that performance degradation may not stem solely from parameter divergence that has been explored previously \cite{FedProx}, but is more closely associated with CFD-V. 

\textbf{(2) Deeper features begin to compress only after aggregation stabilizes earlier layers.}
As shown in Figures \ref{FedAvg_ResNet50_Layerwise_NC1_NC1_Between} - \ref{FedAvg_ResNet50_Epochwise_NC1_NC1_Between}, during FL training, feature variance in shallower layers first converges to a stable level, after which deeper layers begin to exhibit compression.
If this condition is not satisfied, model aggregation can instead destabilize features in deeper layers.
This behavior is closely related to the aforementioned CFD-V and further complicates the convergence of FL models under aggregation.

\begin{takeawaybox}
    Feature variance degradation can accumulate due to the stacked architecture of DNNs, leading to severe disruption in penultimate-layer features despite minimal parameter divergence.
\end{takeawaybox}

\vspace{-0.4cm}
\subsection{Feature-Parameter Mismatch after Model Aggregation}
\vspace{-0.2cm}
\textbf{Motivation} In the previous sections, we analyze how the features themselves are influenced by model aggregation.
However, due to the stacked architecture of DNNs, both the progressive feature extraction process and the final decision stage depend not only on the quality of the features, but also on their coupling with the parameters of subsequent layers.
In this section, we examine the alignment between features and their subsequent parameters to understand how model aggregation influences the consistency between features and the parameters that transform them across layers.

\begin{figure}[!htbp]
\centering
\vspace{-0.4cm}
\includegraphics[width=\textwidth]{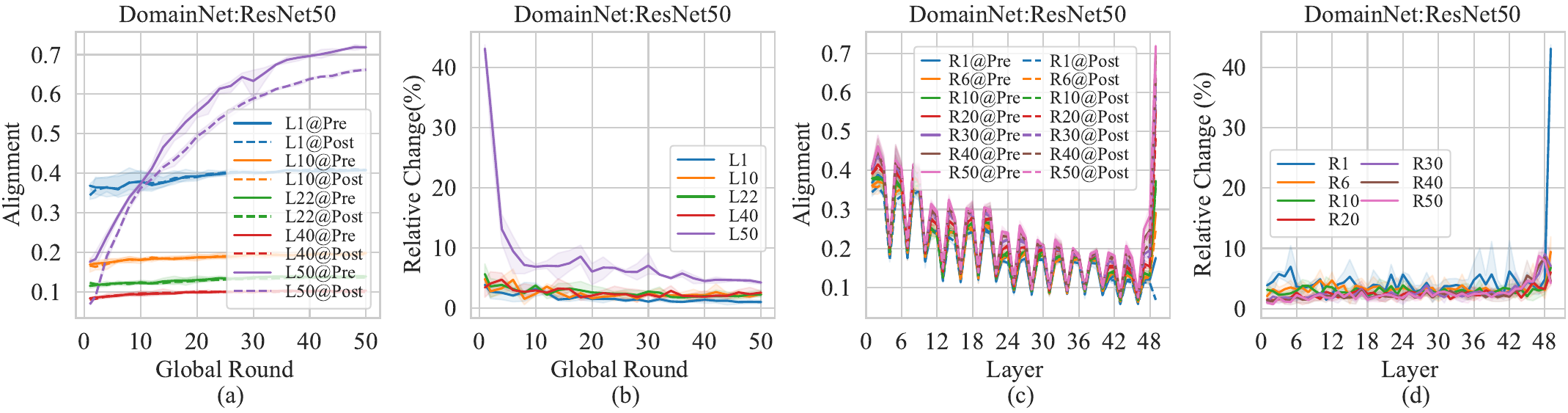}
\vspace{-0.8cm}
\caption{Evolution in feature-parameter alignment. The model is trained on DomainNet using ResNet50.
Subfigures (a) and (b) show the alignment and their relative changes across global rounds.
Subfigures (c) and (d) show the same metrics across model depth.}
\vspace{-0.4cm}
\label{FedAvg_ResNet50_Layerwise_Epochwise_NC3}
\end{figure}

\textbf{Experimental Results}
Figure \ref{FedAvg_ResNet50_Layerwise_Epochwise_NC3} presents the results on DomainNet using Reset50.
Additional results are provided in Appendix \ref{appendix_alignment}.
Based on Figure \ref{FedAvg_ResNet50_Layerwise_Epochwise_NC3}, we can make the following observations:

\textbf{(1) Feature-parameter alignment increases as training progresses.}
From Figure \ref{FedAvg_ResNet50_Layerwise_Epochwise_NC3} (a) and (c), it can be observed as training proceeds, the alignment between features at a give layer and their subsequent parameters gradually increases.
This suggests that during the training process, the transformation between layers becomes progressively more coherent, allowing downstream parameters to better accommodate upstream feature representations.
At the early stages of training, the feature-parameter alignment decreases with increasing model depth, reflecting weaker coupling in deeper layers.
However, the alignment between the penultimate layer (L50) and the classifier improves at a much faster rate than other layers, eventually surpassing all others after several global rounds.

\textbf{(2) Feature-parameter alignment is disrupted by model aggregation and exhibits a cumulative degradation trend.}
As shown in Figure \ref{FedAvg_ResNet50_Layerwise_Epochwise_NC3} (a) and (c), model aggregation clearly disrupts the alignment between features and their subsequent parameters.
Furthermore, as shown in Figures~\ref{FedAvg_ResNet50_Layerwise_Epochwise_NC3} (b) and (d), the relative change in feature-parameter alignment also exhibits a CFD trend, which we denote as CFD-A.
This results in a pronounced mismatch between the penultimate-layer features and the classifier.
Notably, unlike feature variance, the relative change in alignment shows a sharp spike specifically at the interface between the penultimate layer and the classifier---substantially greater than the changes observed in earlier layers.
This pronounced mismatch, together with the previously observed decline in feature discrimination, jointly accounts for the significant performance degradation observed in the post-aggregated model.

\begin{takeawaybox}
    Model aggregation induces progressive misalignment between features and subsequent parameters, with the most severe disruption occurring between the penultimate layer and the classifier.
\end{takeawaybox}

\vspace{-0.4cm}
\subsection{Model Aggregation Improves Model Generalization}
\vspace{-0.4cm}
\textbf{Motivation} Previous experiments show that model aggregation disrupts locally formed feature structures, reducing feature discrimination and degrading performance on local client data.
This raises a key question: if aggregation consistently harms local representations, what is the value of client collaboration in FL?
To address this, we shift focus from feature discrimination to generalization, aiming to uncover the potential benefits of aggregation across diverse data distributions.

\begin{wrapfigure}{r}{0.5\textwidth}
\vspace{-0.4cm}
\centering
\includegraphics[width=2.8in]{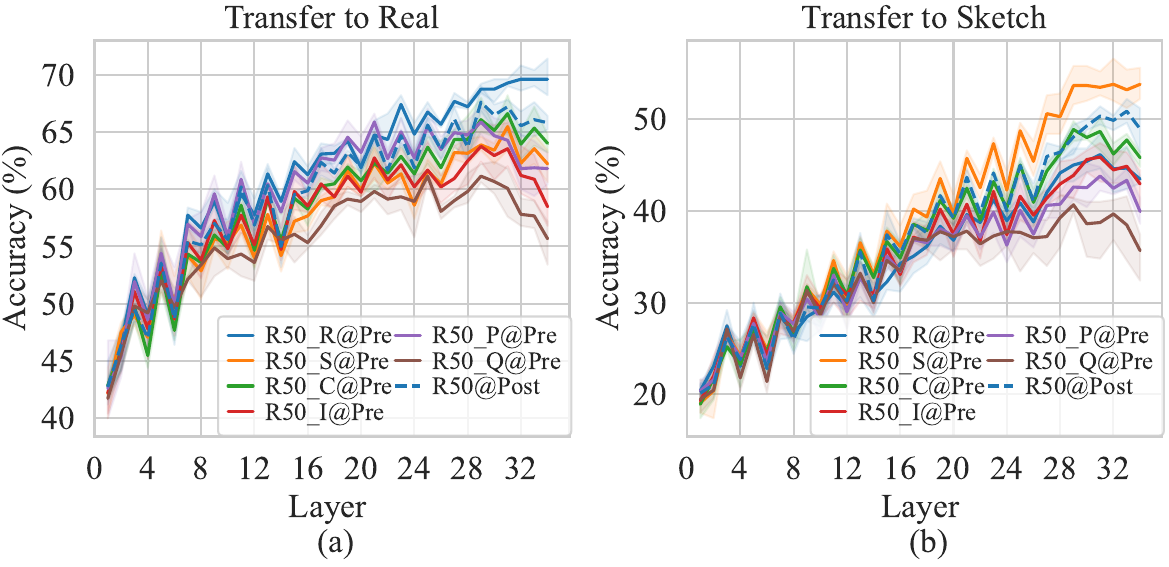}
\vspace{-0.8cm}
\caption{Linear probing accuracy when transferring to `Real' and `Sketch' domains in DomainNet.
Experiments are conducted on ResNet34.
R50@Post denotes the results of post-aggregated model, while R50\_@Pre* indicates the results of pre-aggregated models from domain *.}
\vspace{-0.5cm}
\label{Linear_Probing_DomainNet_E50_ResNet34_Epochwise}
\end{wrapfigure}

\textbf{Experimental Results}
We use linear probing accuracy to evaluate feature generalization.
Figure \ref{Linear_Probing_DomainNet_E50_ResNet34_Epochwise} reports results on `Real' and `Sketch' domains using ResNet34.
Detailed settings and additional results are provided in Appendix \ref{appendix_linear_probing}.
For clarity, we refer to the linear probing accuracy of the pre-aggregated models on its own local data as in-distribution (ID) accuracy, and its performance on other clients' data as out-of-distribution (OOD) accuracy.
Based on Figure \ref{Linear_Probing_DomainNet_E50_ResNet34_Epochwise}, we can draw the following observations:

\textbf{(1) As network depth increases, ID accuracy improves and then plateaus, while OOD accuracy rises initially but eventually declines.}
We observe that the ID accuracy of the pre-aggregated model increases with depth and stabilizes at the final few layers, indicating that the model progressively extracts more discriminative features from its local data.
In contrast, the OOD accuracy exhibits a different trend---it first increases, then decreases---indicating reduced generalization to other clients' data at deeper layers.
This observation aligns with prior findings that DNNs tend to learn generalizable features in shallow layers and task-specific features in deepper layers \citep{yosinski2014transferable,masarczyk2024tunnel}.
These results suggest that, under data heterogeneity in FL, clients may benefit more from collaboratively training on intermediate-layer features rather than relying solely on the final features.

\textbf{(2) Post-aggregated model produces more generalizable features than the pre-aggregated model.}
As shown in Figure \ref{Linear_Probing_DomainNet_E50_ResNet34_Epochwise}, although the pre-aggregated model performs well on ID data, it exhibits a significant drop in OOD accuracy.
This indicates that the pre-aggregated model tends to overfit its local data and fails to learn features that generalize to other clients.
In contrast, the post-aggregated model maintains relatively high performance across diverse data distributions, indicating that aggregation effectively fuses knowledge from clients and yields more generalizable features.

\begin{takeawaybox}
    Model aggregation mitigates the overfitting of local models by fusing locally learned knowledge and facilitates the extraction of more generalizable features across diverse data distributions.
\end{takeawaybox}

\section{Why are Common Solutions Effective?}
In this section, we revisit several common yet effective solutions to address the accuracy drop caused by model aggregation, through the lens of our layer-peeled feature analysis framework.
These solutions include personalizing specific parameters, initializing models with pre-trained weights, and fine-tuning classifier using local data.
Our analysis offers a deeper understanding of why these methods are effective, by linking their impact to the feature-level disruptions and alignment issues identified in the previous sections.

\subsection{Personalizing Specific Parameters} 
\textbf{Motivation} Parameter personalization has been shown to be effective in mitigating the performance degradation caused by model aggregation \citep{FedPer,FedBN,LG-FedAvg,PartialFed}.
In this section, we investigate how personalization affects the feature extraction process from a layer-peeled perspective.

\begin{figure}[ht]
\centering
\includegraphics[width=\textwidth]{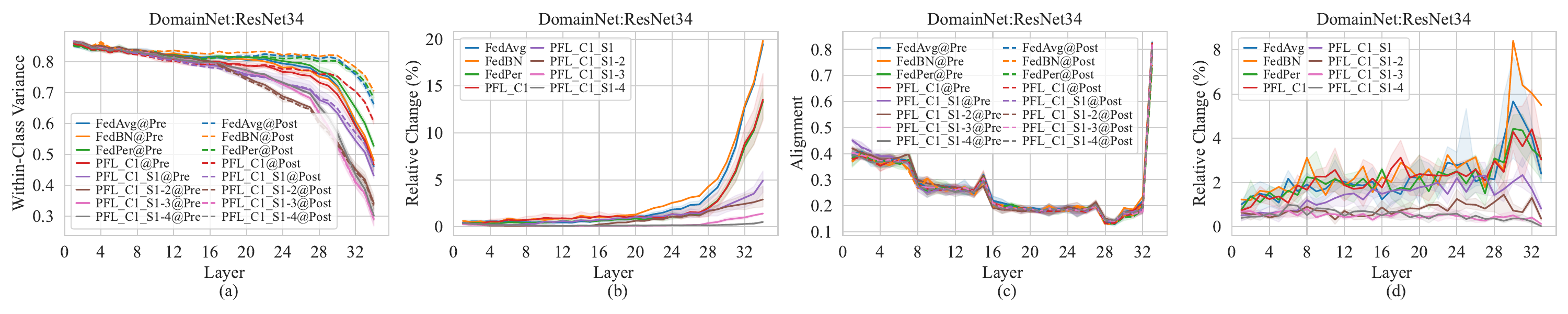}
\vspace{-0.8cm}
\caption{Changes in the normalized within-class feature variance and the alignment between features and parameters when personalizing different parts of the model. The results are obtained by the model in global round $30$, using ResNet34 as the backbone and is trained on DomainNet. PFL\_* denotes the PFL methods that personalize different parts, with C1 referring to the first convolutional layer and S* referring to the stage block.}
\label{PFL_ResNet34_DomainNet_Epochwise_NC3_NC3_R29_4Cols}
\end{figure}

\textbf{Experimental Results} 
We first examine two representative PFL approaches: FedPer \citep{FedPer}, which personalizes the classifier, and FedBN \citep{FedBN}, which personalizes the batch normalization (BN) layers.
Additionally, inspired by PartialFed \citep{PartialFed}, we explore a progressive personalization strategy that successively personalizes multiple layers starting from the input.
Figure \ref{PFL_ResNet34_DomainNet_Epochwise_NC3_NC3_R29_4Cols} presents the results on DomainNet using ResNet34.
Additional results are provided in Appendix \ref{appendix_parameter_personalization}.
It can be observed that personalizing more parameters within the feature extractor generally leads to more compact within-class features and smaller relative changes in feature variance after model aggregation.
This benefit arises because personalizing shallow layers helps prevent the CFD effect---preserving the locally adapted feature extraction capability that can be disrupted by aggregation.

\vspace{-0.4cm}

\begin{takeawaybox}
    Personalizing shallow layers preserves locally adapted feature representations and mitigates the CFD effects introduced by model aggregation. 
\end{takeawaybox}

\subsection{Initializing with Pre-trained Parameters on Large-scaled Dataset}
\textbf{Motivation}
Pre-trained parameters have been widely adopted to initialize FL models and have been shown to accelerate convergence \citep{nguyen2023where,chen2023on}.
However, these studies primarily rely on loss or accuracy to assess the effects of pre-trained parameters.
In this section, we investigate how initializing with pre-trained parameters affects the layer-peeled feature extraction during FL training.

\begin{figure}[ht]
\centering
\vspace{-0.4cm}
\includegraphics[width=\textwidth]{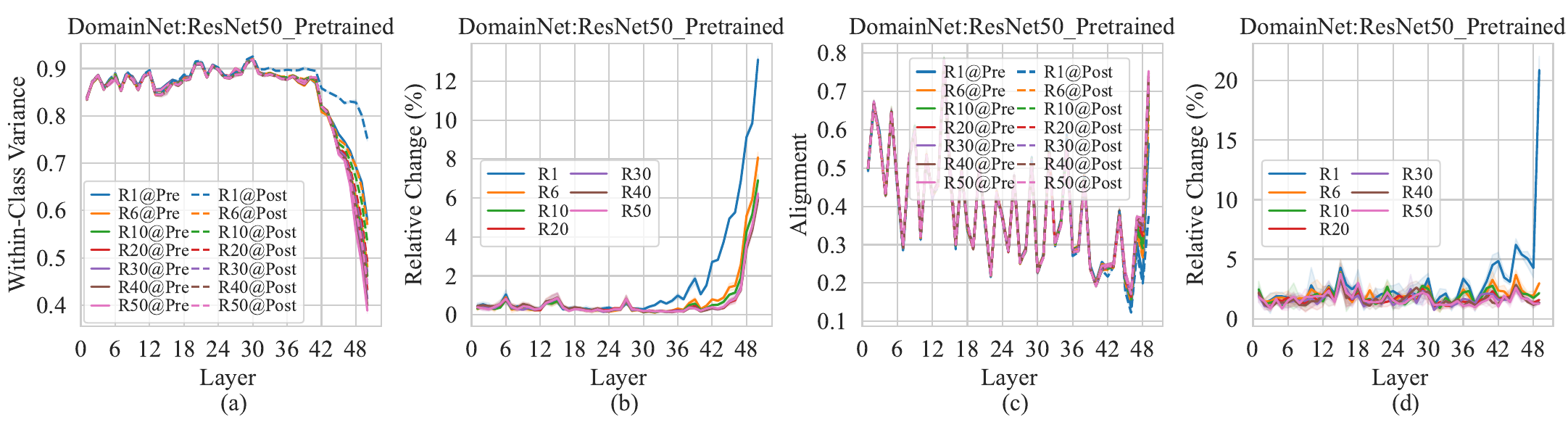}
\vspace{-0.8cm}
\caption{Changes in normalized within-class feature variance and feature-parameter aligment across model layers for specific global rounds. The model is trained on DomainNet using ResNet50 pre-trained on ImageNet: (a) normalized within-class variance, (b) relative change of within-class variance, (c) feature-parameter alignment, (d) relative change of feature-parameter alignment.}
\vspace{-0.4cm}
\label{FedAvg_ResNet50_Pretrained_DomainNet_Epochwise_NC1_NC1_NC3_NC3}
\end{figure}

\textbf{Experimental Results}
Figure \ref{FedAvg_ResNet50_Pretrained_DomainNet_Epochwise_NC1_NC1_NC3_NC3} presenst the results on DomainNet using ResNet50 pretrained on ImageNet. 
Additional experimental results are provided in Appendix \ref{appendix_pretrained}.
It can be observed that both the within-class variance and feature-parameter alignment exhibit lower sensitivity to model aggregation when models are initialized with pre-trained parameters compared to random initialization.
This is because pre-training enables the shallow layers to extract meaningful features early on, allowing the model to concentrate training efforts on deeper layers.
Such initialization helps mitigate the CFD phenomenon discussed in Section \ref{feature_variation_accumulation}, where deeper layers become increasingly sensitive to model aggregation and only begin to converge after sufficient compression in preceding layers.
By accelerating the stabilization of shallow-layer features, pre-training effectively reduces the effective path length of CFD, thereby significantly lowering the relative changes in feature variance and alignment introduced by model aggregation.

\begin{takeawaybox}
    Pre-trained initialization shortens the effective path length of CFD, thereby alleviating the negative impact of model aggregation on feature stability. 
\end{takeawaybox}

\subsection{Fine-tuning Classifier Using Local Data}
\textbf{Motivation} Fine-tuning the classifier using local data has been shown to effectively improve a model’s adaptation to local distributions \citep{FedBABU,FedETF}.
However, the underlying mechanism behind this improvement has not been thoroughly explored.
In this section, we analyze the effect of classifier fine-tuning using the feature-parameter alignment metric within our evaluation framework.

\begin{wrapfigure}{r}{0.5\textwidth}
\centering
\vspace{-0.6cm}
\includegraphics[width=2.7in]{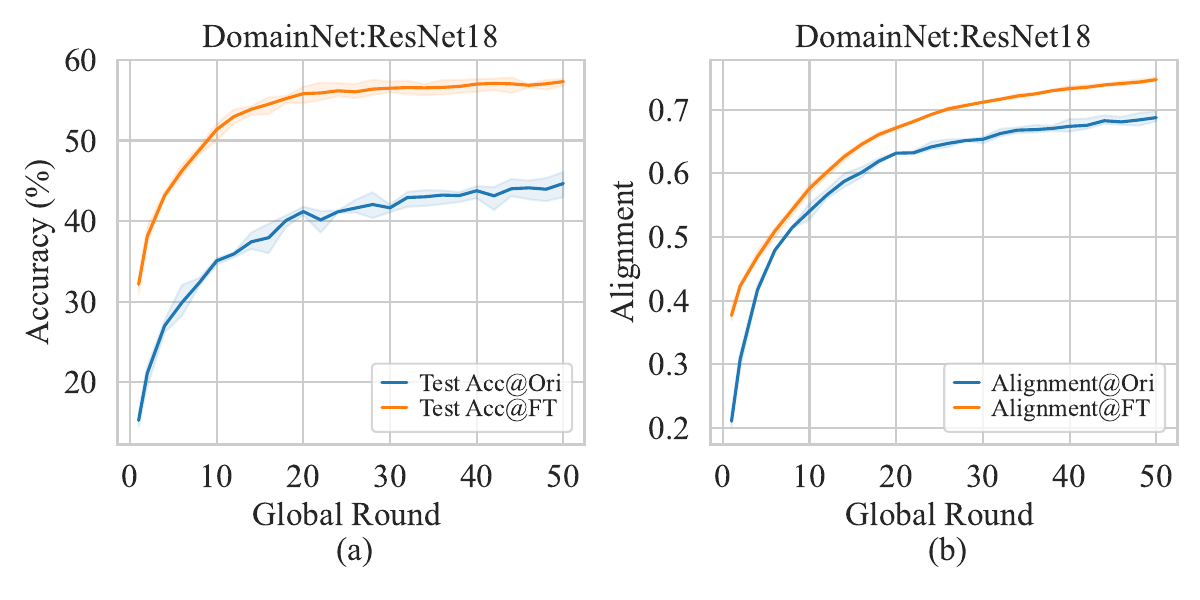}
\vspace{-0.4cm}
\caption{Accuracy and feature-parameter alignment when fine-tuning classifier. The experiments are conducted on DomainNet using ResNet18.}
\vspace{-0.5cm}
\label{FedAvg_ResNet18_DomainNet_Finetune_Classifier}
\end{wrapfigure}

\textbf{Experimental Results} Figure \ref{FedAvg_ResNet18_DomainNet_Finetune_Classifier} shows both accuracy and feature-parameter alignment on DomainNet when fine-tuning the classifier at various global rounds, using ResNet18 as the backbone. 
Additional experiments are provided in Appendix \ref{appendix_finetune_classifier}.
We observe that the alignment between penultimate layer features and the classifier consistently improves with fine-tuning as FL training progresses.
This suggests that fine-tuning strengthens the coupling between penultimate-layer features and the classifier, ultimately enhancing the model performance.

\begin{takeawaybox}
    Fine-tuning classifier align global classifier with local features, thereby improving model adaptation and performance on local data.  
\end{takeawaybox}

\vspace{-0.4cm}
\section{Conclusion}
\vspace{-0.4cm}
In this paper, we construct a layer-peeled feature analysis framework to investigate the impact of model aggregation on feature extraction in FL.
The framework includes metrics that quantify both the quality of extracted features and their alignment with subsequent model parameters.
Using this framework, we conduct a comprehensive analysis of the feature extraction process in FL.
Our findings reveal that although the training dynamics of FL are generally consistent with those of CL---promoting within-class feature compression, between-class separation, and feature-parameter alignment---the aggregation process disrupts these objectives.
More importantly, this disruption accumulates across network layers, leading to substantial degradation in the quality of penultimate-layer features.
We further apply our framework to reinterpret several widely used solutions---such as parameter personalization, pre-trained initialization, and classifier fine-tuning---and demonstrate that they effectively mitigate the feature degradation and accumulation effects observed in our analysis.
Overall, our work offers a new perspective for understanding model aggregation in FL and provides insights for the design of more robust and interpretable FL algorithms.




\bibliographystyle{unsrtnat}
\bibliography{reference.bib}

\newpage
\appendix

\section*{Appendix}\label{Appendix}
\addcontentsline{toc}{section}{Appendix} 

\renewcommand{\contentsname}{Contents} 
\tableofcontents

\clearpage

\section{Related Work}\label{appendix_related_work}
\textbf{Federated Learning.} 
FedAvg \citep{FedAvg} is widely recognized as the pioneering work in FL. It facilitates collaborative model training by iteratively performing independent training on each client and aggregating the models on the server. While FedAvg demonstrates satisfactory performance in IID settings \citep{stich2018local, woodworth2020local}, a performance drop is commonly observed when data distributions are non-IID. This phenomenon has led to multiple explanations for the degradation, such as `client drift' \citep{zhao2018federated,SCAFFOLD} and `knowledge forgetting' \citep{pFedSD}.
To address the challenges posed by non-IID data, various solutions have been proposed, including local model regularization \citep{FedProx}, correction techniques \citep{SCAFFOLD,pFedGF}, knowledge distillation \citep{pFedSD,FedNTD}, and partial parameter personalization \citep{FedBN,FedRep,FedPer,LG-FedAvg,PartialFed,ChannelFed,FedCAC,FedDecomp}. However, these methods generally adhere to the standard framework of local training and global aggregation introduced by FedAvg. There remains a lack of in-depth exploration into how model aggregation intuitively impacts model training from a feature extraction perspective.

\textbf{Federated Learning within Feature Space.} 
Numerous studies have recently observed that heterogeneous data can lead to suboptimal feature extraction in FL models and are working on improving FL models by directly calibrating the resulting suboptimal feature spaces. 
A common line of research attributes performance degradation to inconsistent feature spaces across clients. 
To tackle this problem, CCVR \citep{CCVR} introduces a post-calibration strategy that fine-tunes the classifier after FL training using virtual features generated by an approximate Gaussian Mixture Model (GMM).
Several methods \citep{FedProto, AlignFed, AlignFed1, FedFA, FPL} propose using prototypes to align feature distributions across different clients, ensuring a consistent feature space. Studies such as FedBABU \citep{FedBABU}, SphereFed \citep{SphereFed}, and FedETF \citep{FedETF} utilize various fixed classifiers (e.g., random or orthogonal initialization) as targets to align features across clients.
In addition to inconsistent feature spaces, FedDecorr \citep{FedDecorr_ICLR,FedDecorr_TPAMI} observed that heterogeneous data leads to severe dimensional collapse in the global model. 
To combat this, it applies a regularization term based on the Frobenius norm of the correlation matrix during local training, encouraging the different dimensions of features to remain uncorrelated. 
FedPLVM \citep{FedPLVM} discovers differences in domain-specific feature variance in cross-domain FL. 
Consequently, they propose dual-level prototype clustering that adeptly captures variance information and addresses the aforementioned problem.
However, these studies primarily focus on calibrating features in the penultimate layer while overlooking intermediate features during feature extraction.
In this paper, we present a layer-peeled analysis of how model aggregation affects the feature extraction process and identify several issues related to the hierarchical topology of DNNs.

\textbf{Feature Learning in DNNs.} Advanced DNNs are typically structured with hierarchical layers, enabling them to efficiently and automatically extract informative features from raw data \citep{krizhevsky2012imagenet, allen2023backward, wang2023understanding,DualFed}. Numerous efforts have been made to understand how DNNs transform raw data from shallow to deep layers. A commonly accepted view is that DNNs initially extract transferable universal features and then progressively filter out irrelevant information to form task-specific features \citep{yosinski2014transferable, zeiler2014visualizing, evci2022head2toe, kumar2022finetuning}. Another line of studies demonstrates that DNNs progressively learn features that are compressed within classes and discriminative between classes. Specifically, \citep{alain2017understanding} observes that the linear separability of features increases as the layers become deeper, using a linear probe. \citep{masarczyk2024tunnel} proposes the tunnel effect hypothesis, which states that the initial layers create linearly separable features, while the subsequent layers (referred to as the tunnel) compress these features. Meanwhile, certain research has extended the feature analysis of neural collapse (NC)—originally observed in the penultimate layers \citep{NC}—to intermediate layers \citep{ansuini2019intrinsic, rangamani2023feature, li2024understanding}, showing that the features of each layer exhibit gradual collapse as layer depth increases. \citep{wang2023understanding} provides a theoretical analysis of feature evolution across depth based on deep linear networks (DLNs). These studies focus on centralized training where there is no model aggregation during training, they provide solid support for us to effectively analyze feature evolution in FL, especially the influence of model aggregation on feature extraction. 
It should be noted that several recent studies have begun to explore FL from a layer-wise or layer-peeled perspective \citep{CCVR,chan2024internal,adilova2024layerwise}.
However, these works differ significantly from ours in both methodology and analytical focus.
For example, Chan et al.~\citep{chan2024internal} investigate aggregation behavior primarily in the parameter space and use loss-based metrics to characterize layer-wise changes, without directly analyzing intermediate feature representations.
Other studies such as \citep{CCVR,adilova2024layerwise} rely on feature similarity metrics that require pairwise comparisons between multiple models, which limits their applicability in single-model diagnostics or online evaluation.
In contrast, our work proposes a unified feature-level evaluation framework that can be applied directly to a single model and its corresponding data.
To the best of our knowledge, this is the first approach to characterize the dynamics of layer-wise feature extraction in FL using metrics that independently quantify feature quality and feature-parameter alignment—without the need for auxiliary models or downstream tasks.

\section{Dataset Description and Partition}\label{appendix_dataset_description}
In the experiments, we primarily focus on the data-heterogeneous FL setting, where the distribution of raw input data differs across clients. This data heterogeneity, often referred to as cross-domain FL, is commonly observed in practical FL applications due to variations in data collection conditions across clients.
Building on previous studies \citep{FedBN, AlignFed, AlignFed1, FedPLVM}, we use three widely used public cross-domain datasets: Digit-Five, PACS \citep{PACS}, and DomainNet \citep{DomainNet}. These datasets contain multiple domains, with each domain consisting of images with different backgrounds and styles, which effectively simulate the data heterogeneity caused by variations in raw input.

\begin{figure*}[ht]
    \centering
\includegraphics[width=5.5in]{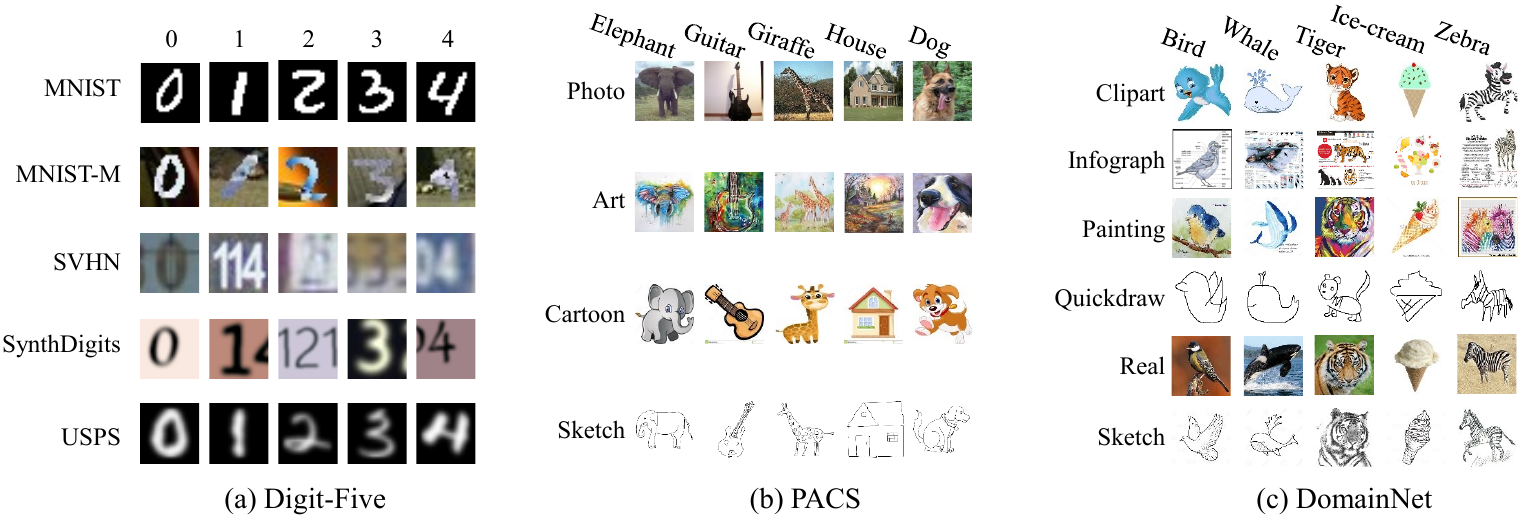}
    \caption{Visualization of example samples within the datasets used for layer-wise feature evaluation: (a) Digit-Five, (b) PACS, (c) DomainNet. For each domain within these adopted datasets, we show the representative samples from 5 classes.}   \label{dataset_visualization}
\end{figure*}

The Digit-Five dataset includes images across 10 classes and 5 domains, namely: MNIST-M \citep{SynthDigits}, MNIST \citep{MNIST}, USPS \citep{USPS}, SynthDigits \citep{SynthDigits}, and SVHN \citep{SVHN}. The PACS dataset includes 4 distinct domains with a total of 7 classes: Photo (P), Art (A), Cartoon (C), and Sketch (S). The DomainNet dataset comprises six domains: Clipart (C), Infograph (I), Painting (P), Quickdraw (Q), Real (R), and Sketch (S). Initially, the DomainNet dataset includes 345 classes per domain. Based on prior research \citep{FedBN, AlignFed}, we reduce the number of classes to 10 commonly used ones for our layer-peeled feature analysis. Figure \ref{dataset_visualization} shows some example images from these three datasets. The representative images demonstrate significant variations across different domains, as observed in Figure \ref{dataset_visualization}.

To create the data-heterogeneous setting across different clients, we assign the images from a single domain to each client. As a result, there are 5 clients for the Digit-Five dataset, 4 clients for PACS, and 6 clients for DomainNet in our analysis.
For the Digit-Five dataset, the number of training and testing samples are set to 500 and 1000, respectively. For both the PACS and DomainNet datasets, the number of training and testing samples is set to 500. The images in the Digit-Five dataset are scaled to $32 \times 32$ for both the training and testing datasets. For PACS and DomainNet, the images are scaled to $224 \times 224$ and we apply data augmentations such as random flipping and rotation for the training samples. No data augmentations are applied to the Digit-Five or test datasets across all experiments.

\section{Model Architectures}\label{appendix_model_architecture}
We utilize multiple models for the selected datasets to perform layer-wise feature extraction analysis, including both Convolutional Neural Networks (CNN) and Vision Transformers (ViT). Specifically, the CNN models we employ are ConvNet, VGG13\_BN \citep{VGG}, and three variants of ResNet \citep{ResNet} (ResNet18, ResNet34, and ResNet50). For the ViT architecture, we apply ViT\_B/16 \citep{ViT}.

The ConvNet model consists of several convolutional layers followed by fully connected (FC) layers. The detailed architecture of ConvNet is presented in Table \ref{Arch_ConvNet}, with the output size calculated using the input scale of the Digit-Five dataset as an example. For the other models, we modify only the classifiers by adjusting the number of output classes to match the requirements of our dataset, while leaving the backbone architecture unchanged.

For the Digit-Five dataset, we use ConvNet, ResNet18, and ResNet34 for FL training. For both PACS and DomainNet datasets, we utilize ConvNet, VGG1\_BN, ResNet18, ResNet34, ResNet50, and ViT\_B/16 to accomplish the FL training.

\begin{longtable}{|c|c|c|} 
\caption{Detailed Architecture of ConvNet} \label{Arch_ConvNet}\\
\hline
\textbf{Layer} & \textbf{Output Size} & \textbf{Description} \\
\hline
\endfirsthead
\hline
\textbf{Layer} & \textbf{Output Size} & \textbf{Description} \\
\hline
\endhead
\hline
\endfoot
\hline
\endlastfoot

Input & 32x32x3 & Input image \\
\hline
Conv1\_1 & 32x32x64 & 5x5 Convolution, 64 filters, stride 1, padding 2 \\
\hline
BN1\_1 & 32x32x64 & Batch Normalization \\
\hline
ReLU1\_1 & 32x32x64 & ReLU activation \\
\hline
MaxPool1\_1 & 16x16x64 & 2x2 Max Pooling, stride 2 \\
\hline
Conv1\_2 & 16x16x64 & 5x5 Convolution, 64 filters, stride 1, padding 2 \\
\hline
BN1\_2 & 16x16x64 & Batch Normalization \\
\hline
ReLU1\_2 & 16x16x64 & ReLU activation \\
\hline
MaxPool1\_2 & 8x8x64 & 2x2 Max Pooling, stride 2 \\
\hline
Conv1\_3 & 8x8x128 & 5x5 Convolution, 128 filters, stride 1, padding 2 \\
\hline
BN1\_3 & 8x8x128 & Batch Normalization \\
\hline
ReLU1\_3 & 8x8x128 & ReLU activation \\
\hline
Flatten & 8192 & Flatten layer for fully connected input \\
\hline
Linear2\_1 & 2048 & Fully connected layer, 2048 units \\
\hline
BN2\_1 & 2048 & Batch Normalization \\
\hline
ReLU2\_1 & 2048 & ReLU activation \\
\hline
Linear2\_2 & 512 & Fully connected layer, 512 units \\
\hline
BN2\_2 & 512 & Batch Normalization \\
\hline
ReLU2\_2 & 512 & ReLU activation \\
\hline
Output (C) & $C$ & Output layer (number of classes) \\
\hline
\end{longtable}

\section{Implementation Details of Experiments} \label{implementation_details_appendix}
In our experiments, we train the model using the standard FL training process introduced by FedAvg \citep{FedAvg}. This process involves iterative local model updating and global model aggregation.
During the local updating phase, the model is optimized using each client’s private data for $E$ epochs, after which the local models are uploaded to the server for aggregation.
In the model aggregation stage, we apply parameter-wise averaging, using the number of samples as weights for each client's local model.
The above procedures are repeated for $R$ global rounds until the global model converges.
 
For local model updating, we use stochastic gradient descent (SGD) with momentum for model optimization, where the learning rate is set to 0.01 and the momentum is set to 0.5. The batch size for local updates is set to 64.
Unless otherwise specified, the number of local update epochs $E$ is set to 10, and the number of global rounds $R$ is set to 50.

All of our experiments are conducted using the PyTorch framework \citep{Pytorch} and implemented on a four-card Nvidia V100 (32G) cluster. During the layer-peeled feature analysis, we use the \textit{forward hook} in PyTorch to extract input features from each evaluated layer as we move from shallow to deeper layers.

For ConvNet, the features preceding each convolutional and linear layer are used for evaluation.
For ResNet, the features before each convolutional layer (excluding the first convolutional layer), the global average pooling layer, and the classifier are used for evaluation.
For VGG13\_BN, the features before each convolutional layer, the linear layers, and the classifier are used for evaluation. For ViT\_B/16, we evaluate the features passed to the multilayer perceptron (MLP), self-attention layers, layer normalization layers, and the final classifier.

To reduce computational cost during feature evaluation, we apply average pooling to the intermediate features, thereby lowering their dimensionality. For CNNs, let the intermediate features at the $\ell$-th layer be represented as $B \times H^{\ell} \times W^{\ell} \times C^{\ell}$, where $B$ denotes the batch size, and $H^{\ell}$, $W^{\ell}$, and $C^{\ell}$ are the height, width, and channel dimensions, respectively. Following previous studies \citep{sarfi2023simulated, harun2024what}, we apply $2 \times 2$ \textit{adaptive average pooling} to the height and width dimensions ($H^{\ell} \times W^{\ell}$). After this adaptive average pooling operation, the intermediate features within the convolutional layers are reduced to $B \times 2 \times 2 \times C^{\ell}$. We then flatten this tensor into a one-dimensional vector and compute the feature metrics, where the features of one sample within the batch has a dimension of $4C^{\ell}$, and the total dimension of the samples within the batch is $B \times 4C^{\ell}$. For ViTs, following previous studies \citep{raghu2021vision,harun2024what}, we apply \textit{global average pooling} to aggregate the image tokens, excluding the class token. For features input to the linear layer with a two-dimensional shape of $B \times D^{\ell}$, we directly use these features to perform evaluation.

During feature evaluation, we process the features in a batch-wise manner, concatenating them across all batches in the evaluated dataset. The corresponding feature metrics are then computed based on these concatenated features.
For all experiments, we evaluate the features every 20 local updating epochs, which corresponds to 2 global rounds when the local update epoch is set to 10.
To minimize the impact of randomness, each experiment is repeated three times with different random seeds.

\section{Feature Evaluation Metrics}\label{appendix_metric}
We apply the following metrics to evaluate the features generated by the pre-aggregated model and post-aggregated model, including the \textit{feature variance, alignment between features and parameters, accuracy of linear probing, pairwise distance of features and models, relative change of evaluated metrics}. For simplicity, we omit the client and sample indices and focus solely on the computation process of these metrics. As previously stated, the features in our experiments are first stacked into a two-dimensional tensor, and then used to compute the feature metrics.
We assume that the stacked features at $\ell$-th layer is denoted as $\bm{Z}^{\ell} \in R^{N \times D^{\ell}}$, where $N$ is the total number of samples used for feature evaluation, and $D^{\ell}$ denote the feature dimension for one sample at $\ell$-th layer.  
The metrics used to evaluate features in this paper are computed as follows.

\subsection{Feature Variance}
Building on previous studies \citep{rangamani2023feature, wang2023understanding}, we use within-class feature variance and between-class feature variance to evaluate the feature structure. Specifically, the within-class variance quantifies the degree of feature compression within the same class, while the between-class variance measures the degree of feature discrimination between different classes.

Before calculating the within-class and between-class variances, we first compute the within-class, between-class, and total covariance matrices of the features at $\ell$-th layer as follows:
\begin{equation}
\begin{aligned}
\Sigma^{\ell}_W &= \frac{1}{N} \sum_{c=1}^C \sum_{i=1}^{N_c} \left( \bm{z}^{\ell}_{c, i} - \bm{\mu}^{\ell}_c \right) \left( \bm{z}^{\ell}_{c, i} - \bm{\mu}^{\ell}_c \right)^{\top} \\
\Sigma^{\ell}_B &= \frac{1}{C} \sum_{c=1}^C \left( \bm{\mu}^{\ell}_c - \bm{\mu}^{\ell}_G \right) \left( \bm{\mu}^{\ell}_c - \bm{\mu}^{\ell}_G \right)^{\top} \\
\Sigma^{\ell}_T &= \frac{1}{N} \sum_{c=1}^C \sum_{i=1}^{N_c} \left( \bm{z}^{\ell}_{c,i} - \bm{\mu}^{\ell}_G \right) \left( \bm{z}^{\ell}_{c,i} - \bm{\mu}^{\ell}_G \right)^{\top},
\end{aligned}
\end{equation}

where

\begin{equation}
    \bm{\mu}^{\ell}_c := \frac{1}{N_c} \sum_{i=1}^{N_c} \bm{z}^{\ell}_{c,i} \quad \bm{\mu}^{\ell}_G := \frac{1}{N} \sum_{c=1}^C \sum_{i=1}^{N_c} \bm{z}^{\ell}_{c,i}.
\end{equation}

In the above equation, \(\bm{\mu}^{\ell}_c\) is the mean class feature computed from the samples within the \(c\)-th class, and \(\bm{\mu}^{\ell}_G\) as the global mean feature computed from all samples. \(N_c\) represents the number of samples within the \(c\)-th class. It should be noted that the total covariance matrix can be decomposed as the sum of the within-class and between-class covariances, i.e.,

\begin{equation}
\Sigma^{\ell}_T = \Sigma^{\ell}_W + \Sigma^{\ell}_B.
\end{equation}

Based on the computed covariance matrices, we then use the total variance \( \text{Tr}(\Sigma^{\ell}_T) \) as the normalization factor and compute the normalized within-class variance and between-class variance as follows:

\begin{equation}
    \Bar{\sigma}^{\ell}_W = \frac{\text{Tr}(\Sigma^{\ell}_W)}{\text{Tr}(\Sigma^{\ell}_T)},
\end{equation}
\begin{equation}
    \quad \Bar{\sigma}^{\ell}_B = \frac{\text{Tr}(\Sigma^{\ell}_B)}{\text{Tr}(\Sigma^{\ell}_T)},
\end{equation}

where \(\text{Tr}(\cdot)\) denotes the trace of the covariance matrices. These two normalized variances are used to measure the within-class feature compression and between-class feature discrimination, respectively, in this paper.

\subsection{Alignment between Features and Subsequent Parameters}
Following previous studies, we use the principal angles between subspaces (PABS) \citep{rangamani2023feature, jordan1875essai, bjorck1973numerical}---denoted as $\theta_1, \ldots, \theta_C$—to measure the alignment between the range space of class-wise feature means $\Bar{\bm{Z}}^{\ell}$ and the top $C$-rank subspace of the subsequent input layer parameters $\bm{W}^{\ell + 1}$.

Specifically, for a linear layer, we first apply singular value decomposition (SVD) to $\bm{W}^{\ell+1}$ and $\Bar{\bm{Z}}^{\ell}$, i.e., $\bm{W}^{\ell+1} = \bm{U}^{\ell+1}_{\bm{W}}\bm{S}^{\ell+1}_{\bm{W}}(\bm{V}_{\bm{W}}^{\ell+1})^{T}$ and $\Bar{\bm{Z}}^{\ell} = \bm{U}^{\ell}_{\Bar{\bm{Z}}}\bm{S}^{\ell}_{\Bar{\bm{Z}}}(\bm{V}_{\Bar{\bm{Z}}}^{\ell})^{T}$. We then compute the PABS between $\bm{V}^{\ell+1}_{\bm{W}}$ and $\bm{U}^{\ell}_{\Bar{\bm{Z}}}$, which represent the basis for the input subspace $\bm{W}^{\ell+1}$ and the range space of $\Bar{\bm{Z}}^{\ell}$, respectively. The alignment is finally computed by the mean of the singular values of $(\bm{V}^{\ell+1}_{W})^{T} \bm{U}^{\ell}_{\Bar{\bm{Z}}}$.

For the convolutional layer, assume the filter kernel has shape $\bm{W}^{\ell+1} \in R^{C^{\ell+1}_{\text{out}} \times C^{\ell}_{\text{in}} \times k^{\ell+1}_H \times k^{\ell+1}_W}$,
and the class-wise feature means have shape 
$\Bar{\bm{Z}}^{\ell} \in \mathbb{R}^{C \times C^{\ell}_{\text{in}} \times H^{\ell} \times W^{\ell}}$.
As previously stated, \(\Bar{\bm{Z}}^{l}\) can be reshaped into
$C \times (C^{\ell}_{\text{in}} \times 4)$.
We begin by flattening the features and parameters along the 
\(C^{\ell}_{\text{in}}\) dimension, resulting in shapes 
$C^{\ell}_{\text{in}} \times (C \times 4)$
for the features and
$C^{\ell}_{\text{in}} \times (C^{\ell + 1}_{\text{out}} \times k^{\ell + 1}_H \times k^{\ell + 1}_W)$
for the parameters. We then compute the alignment along the 
\(C^{\ell}_{\text{in}}\) dimension as described above.

For the self-attention layer in a ViT, we compute the alignments of 
the features and the \(QKV\) matrices separately, and then take the 
average of these alignments as the final metric.

\subsection{Accuracy of Linear Probing}
Linear probing is a technique used in transfer learning to evaluate the quality of learned features by training a simple linear classifier on top of the features extracted from a pre-trained model \citep{chen2020simple, he2022masked, wang2023does}. In this paper, we employ the linear probing technique to assess feature generalization across diverse data distributions. Specifically, after extracting features from different layers, we apply a randomly initialized linear classifier on top of these features. This classifier is then trained using the training subset of the evaluated datasets, and we compute accuracy by testing on the corresponding samples from the test datasets. The testing accuracy serves as the metric for evaluating feature generalization on each dataset.

\subsection{Pairwise Distance of Features and Models}
We use four metrics to evaluate the distance between two models or the features extracted by them: \textit{mean normalized $L_1$ distance, mean squared distance, mean $L_1$ distance, and cosine similarity}. In this section, we focus on the computation of the distance between features, since the model parameters are reshaped into a single vector, which can be treated as a feature with a sample size of 1. Thus, the distance computation is applied to these features can be directly transferred to models. Let the features of the pre-aggregated and post-aggregated models be denoted as $\bm{Z}^{\ell}_{pre}$ and $\bm{Z}^{\ell}_{post}$, respectively. The corresponding distance can then be computed as follows.

\begin{enumerate}[label=\textbullet] 
\item \textbf{Mean Normalized $L_1$ Distance.}
This measure computes the mean normalized $L_1$ distance between the pre-aggregated and post-aggregated feature matrices, and then averages the distances across all elements, as shown below.
\begin{equation}
    D^{\ell}_{\hat{L}_1} = \frac{1}{ND} \sum_{i=1}^{N} \sum_{j = 1}^{D} \frac{|\bm{Z}^{\ell}_{pre}(i, j) - \bm{Z}^{\ell}_{post}(i, j)|}{|\bm{Z}^{\ell}_{pre}(i, j)| + |\bm{Z}^{\ell}_{post}(i, j)|}
\end{equation}
\item \textbf{Mean Squared Error.}
This distance measure computes the average squared differences between corresponding elements of the feature vectors, as shown below.
\begin{equation}
    D^{\ell}_{s} = \frac{1}{ND} \sum_{i=1}^{N} \sum_{j = 1}^{D} (\bm{Z}^{\ell}_{pre}(i, j) - \bm{Z}^{\ell}_{post}(i, j))^2
\end{equation}
\item \textbf{Mean $L_1$ Distance.}
This distance computes the average $L_1$ distances between corresponding elements of the pre-aggregated and post-aggregated feature matrices, as shown below.
\begin{equation}
    D^{\ell}_{L_1} = \frac{1}{ND} \sum_{i=1}^{N} \sum_{j = 1}^{D} |\bm{Z}^{\ell}_{pre}(i, j) - \bm{Z}^{\ell}_{post}(i, j)|
\end{equation}

\item \textbf{Mean Cosine Similarity.} This measure computes the cosine of the angle between the pre-aggregated and post-aggregated features of the same samples, and then averages these values across all clients. It quantifies the cosine of the angle between the pre-aggregated and post-aggregated features, where a value of 1 indicates identical directions and a value of -1 indicates opposite directions.
The formulation of the mean cosine similarity is shown below.
\begin{equation}
    D^{\ell}_{cos} = \frac{1}{N} \sum_{i=1}^{N} \frac{Z^{\ell}_{pre}(i, :)\cdot Z^{\ell}_{post}(i, :)}{||Z^{\ell}_{pre}(i, :)||||Z^{\ell}_{post}(i, :)||}
\end{equation}
where $Z^{\ell}_{pre}(i, :) \cdot Z^{\ell}_{post}(i, :)$ denotes the dot product of the pre-aggregated and post-aggregated features of sample $i$ at $\ell$-th layer, respectively, and $||Z^{\ell}_{pre}(i, :)||$ and $||Z^{\ell}_{post}(i, :)||$ denote the Euclidean norms of the pre-aggregated and post-aggregated features at $\ell$-th layer, respectively.
\end{enumerate}

\subsection{Relative Change of Evaluated Metrics}
Since the original metrics of features at different layers can vary in magnitude, we use the relative change in the evaluated metrics to measure the ratio of change before and after aggregation. 
Let $V^{\ell}_{pre}$ and $V^{\ell}_{post}$ represent the metrics of features generated by the models before and after aggregation at $\ell$-th layer, respectively. The relative change in the evaluated metrics is then defined as:

\begin{equation}
    \Delta^{\ell}(V) = \frac{|V^{\ell}_{post} - V^{\ell}_{pre}|}{|V^{\ell}_{pre}| + |V^{\ell}_{post}|} * 100\%.
\end{equation}

\section{Detailed Results of Performance Drop in Model Aggregation}\label{performance_drop_appendix}
\begin{figure}[ht]
\centering
\includegraphics[width=\textwidth]{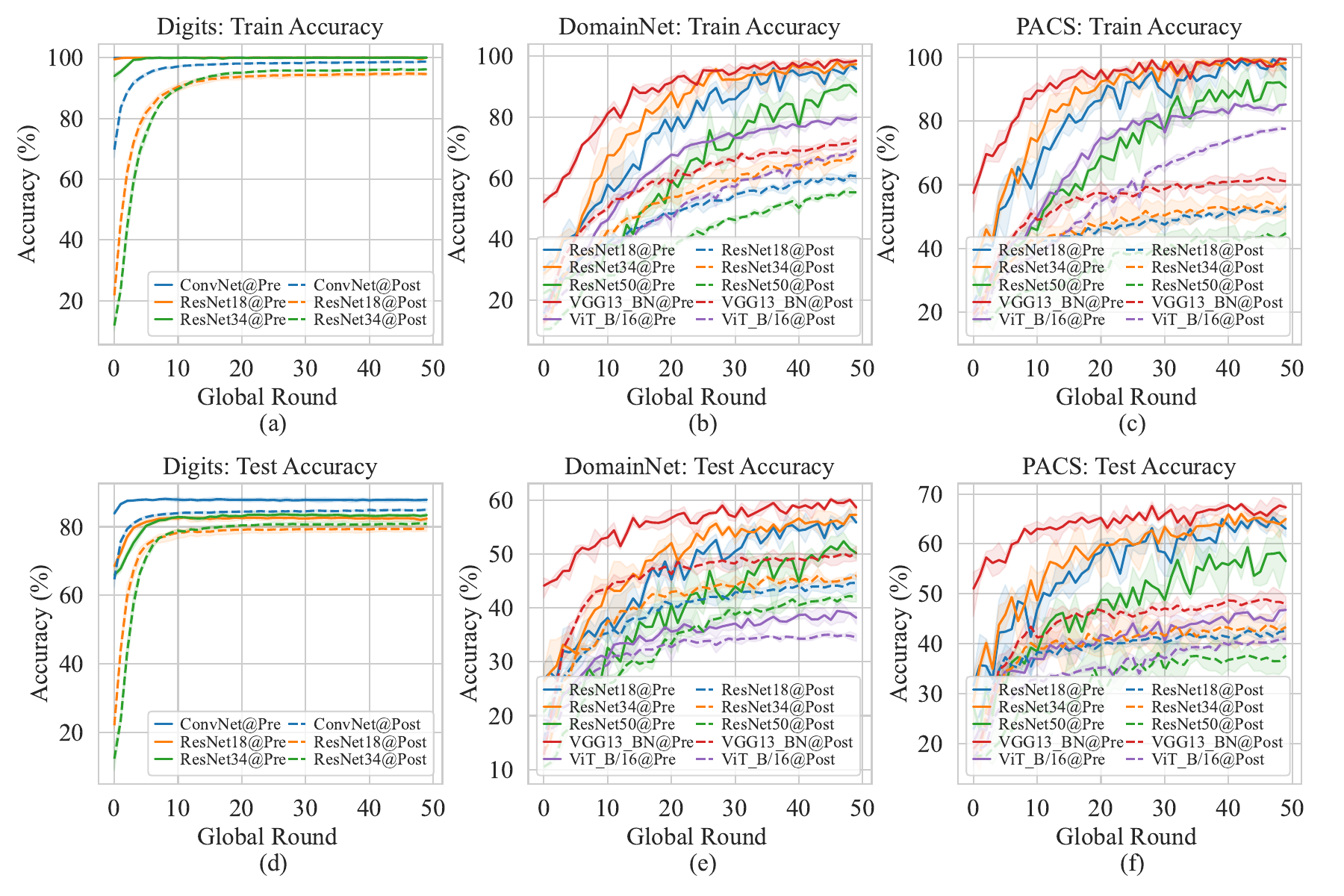}
\caption{Training and testing accuracy curves of the model before and after aggregation, evaluated on the local dataset during FL training. The experiments are conducted on Digit-Five, DomainNet, and PACS, using multiple model architectures as the backbone.}
\label{FedAvg_Acc_DomainNet_PACS_Digits}
\end{figure}

In this section, we provide more detailed results that demonstrate the performance drop during model aggregation.
In these experiments, we perform inference on both the training and testing datasets using the pre-aggregated and post-aggregated models.
The experimental results are shown in Figure \ref{FedAvg_Acc_DomainNet_PACS_Digits}.
These experiments are conducted on different datasets, including Digit-Five, PACS, and DomainNet, and on various model architectures.
From Figure \ref{FedAvg_Acc_DomainNet_PACS_Digits}, we can observe that the performance drop introduced by model aggregation is consistent across all adopted datasets and model architectures, on both training and testing dataset.
The performance drop consistently occurs throughout the entire training procedure of FL.
These results indicate that the performance drop during model aggregation is a common phenomenon in FL.

\vspace{5.0cm}

\section{Detailed Results of Feature Variance}\label{appendix_variance}
In this section, we provide a detailed analysis of the layer-wise feature variance during FL training. Our experiments are conducted on three datasets: Digit-Five, PACS, and DomainNet, using various model architectures, as previously described.
For each dataset, we presents four types of feature variances: normalized within-class feature variance, normalized between-class feature variance, original unnormalized within-class feature variance, and original unnormalized between-class feature variance.

To better visualize feature evolution over time (across different epochs of FL training) and space (across different layers), we employ two types of visualizations. The first visualizes feature changes across different layers while keeping the training global round fixed. The second focuses on visualizing feature evolution across training rounds while fixing specific layers.

The experimental results are presented in the following sections. From these results, we observe that both the original within-class and between-class feature variances increase as the layer depth increases. In contrast, the normalized within-class feature variance decreases with both layer depth and training rounds, which is in contrast to the normalized between-class feature variance. This suggests that features within the same class become more compressed, while features across different classes become more discriminative.

However, after model aggregation, the normalized within-class variance increases while the normalized between-class variance decreases. This indicates that model aggregation disrupts the feature compression objective during DNN training.
More importantly, this disruption progressively accumulates across model layers, causing the features in the penultimate layer (which are used for final decision-making) to degrade more significantly.

\subsection{Changes of Feature Variance Across Layers}

\begin{figure}[H]
\centering
\includegraphics[width=4.5in]{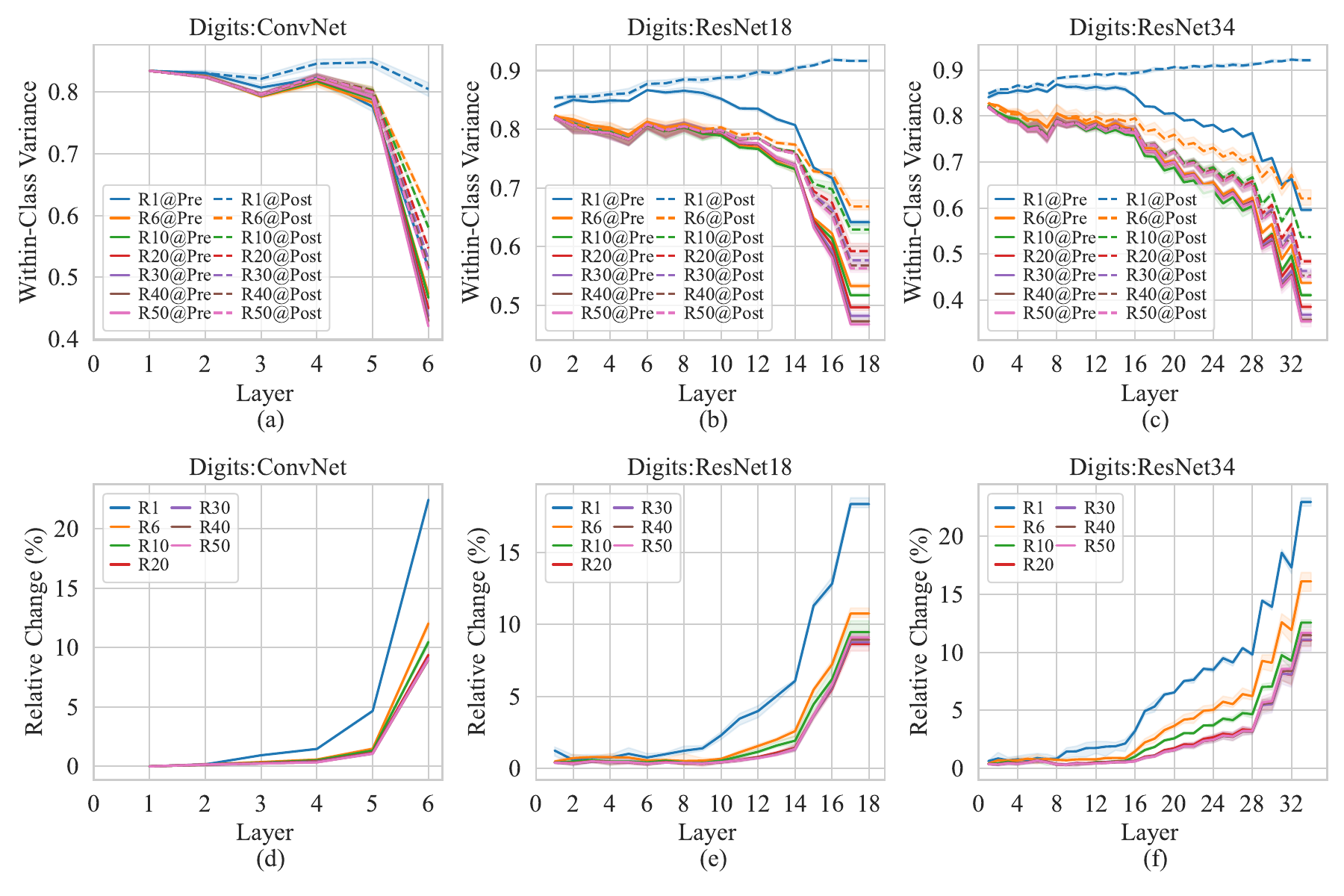}
\caption{Changes in the normalized within-class variance of features across model layers for specific global rounds, with larger X-axis values indicating deeper layers. The model is trained on Digit-Five with multiple models that are randomly initialized. The top half of the figure shows the normalized within-class variance, while the bottom half displays the relative change in variance before and after model aggregation.}
\label{FedAvg_AllModel_Digits_Epochwise_NC1_NC1}
\end{figure}

\begin{figure}[H]
\centering
\includegraphics[width=4.5in]{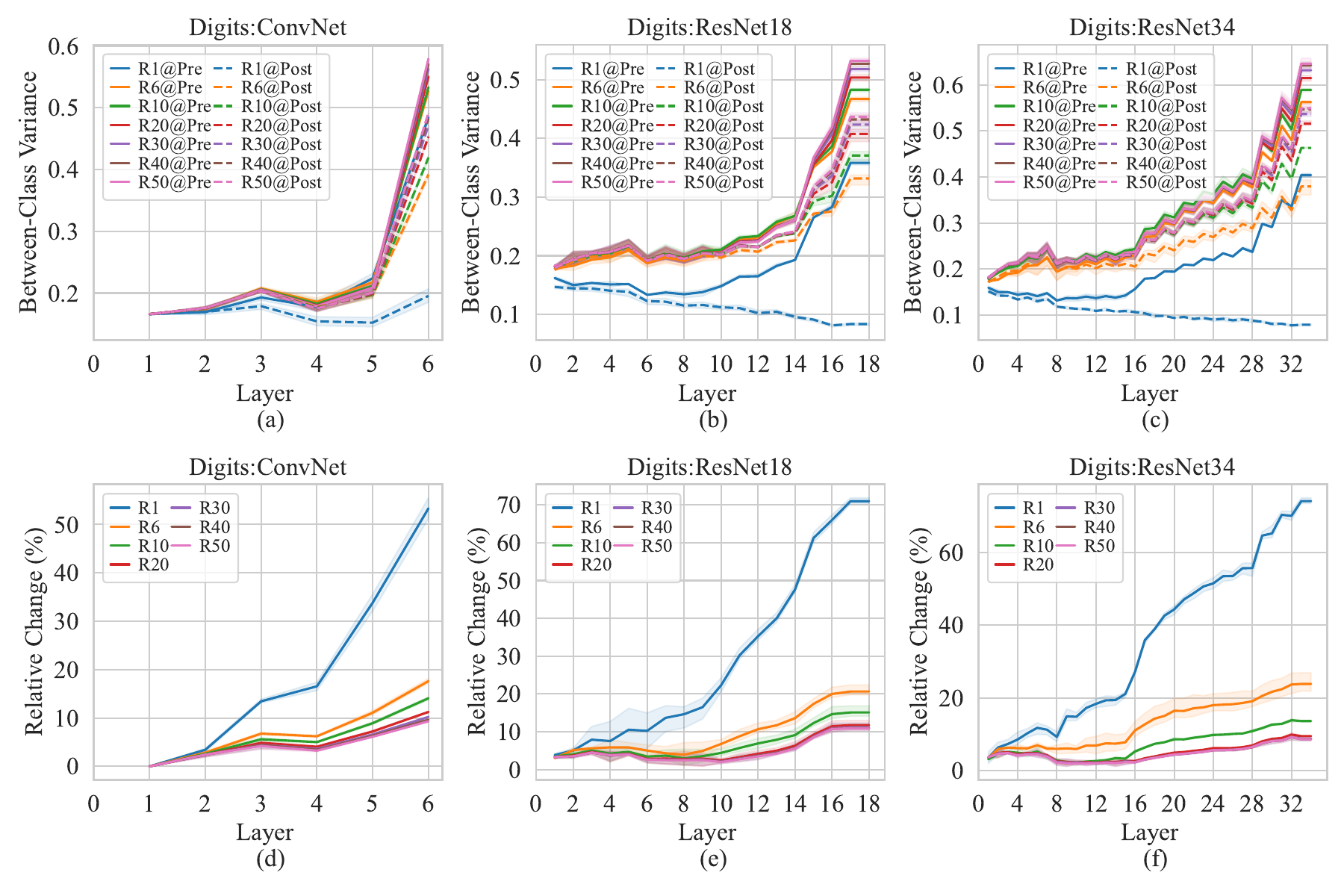}
\caption{Changes in the normalized between-class variance of features across model layers for specific global rounds, with larger X-axis values indicating deeper layers. The model is trained on Digit-Five with multiple models that are randomly initialized. The top half of the figure shows the normalized between-class variance, while the bottom half displays the relative change in variance before and after model aggregation.}
\label{FedAvg_AllModel_Digits_Epochwise_NC1_between}
\end{figure}

\begin{figure}[H]
\centering
\includegraphics[width=4.5in]{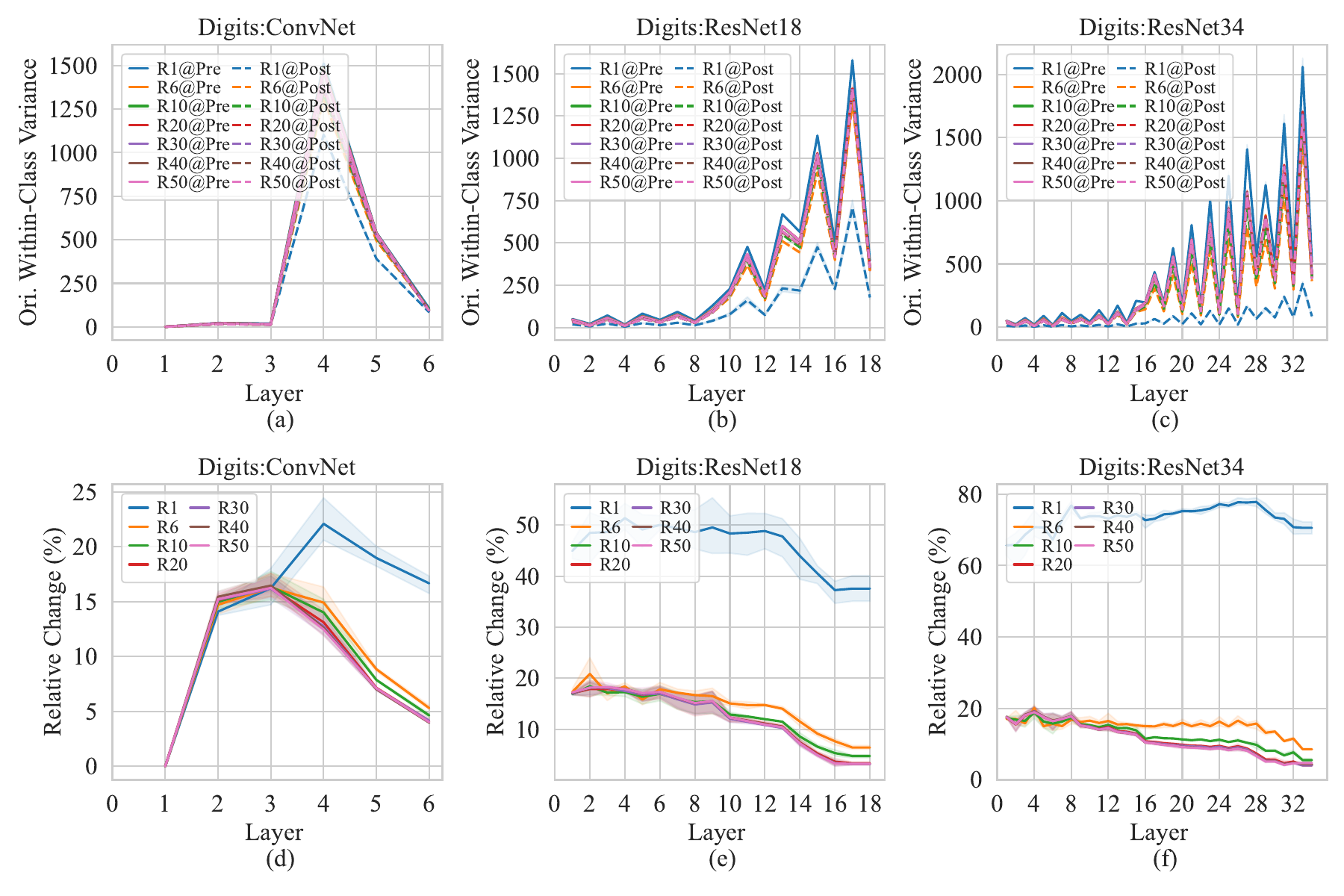}
\caption{Changes in the original unnormalized within-class variance of features across model layers for specific global rounds, with larger X-axis values indicating deeper layers. The model is trained on Digit-Five with multiple models that are randomly initialized. The top half of the figure shows the original unnormalized within-class variance, while the bottom half displays the relative change in variance before and after model aggregation.}
\label{FedAvg_AllModel_Digits_Epochwise_NC1_trace_within}
\end{figure}

\begin{figure}[H]
\centering
\includegraphics[width=4.5in]{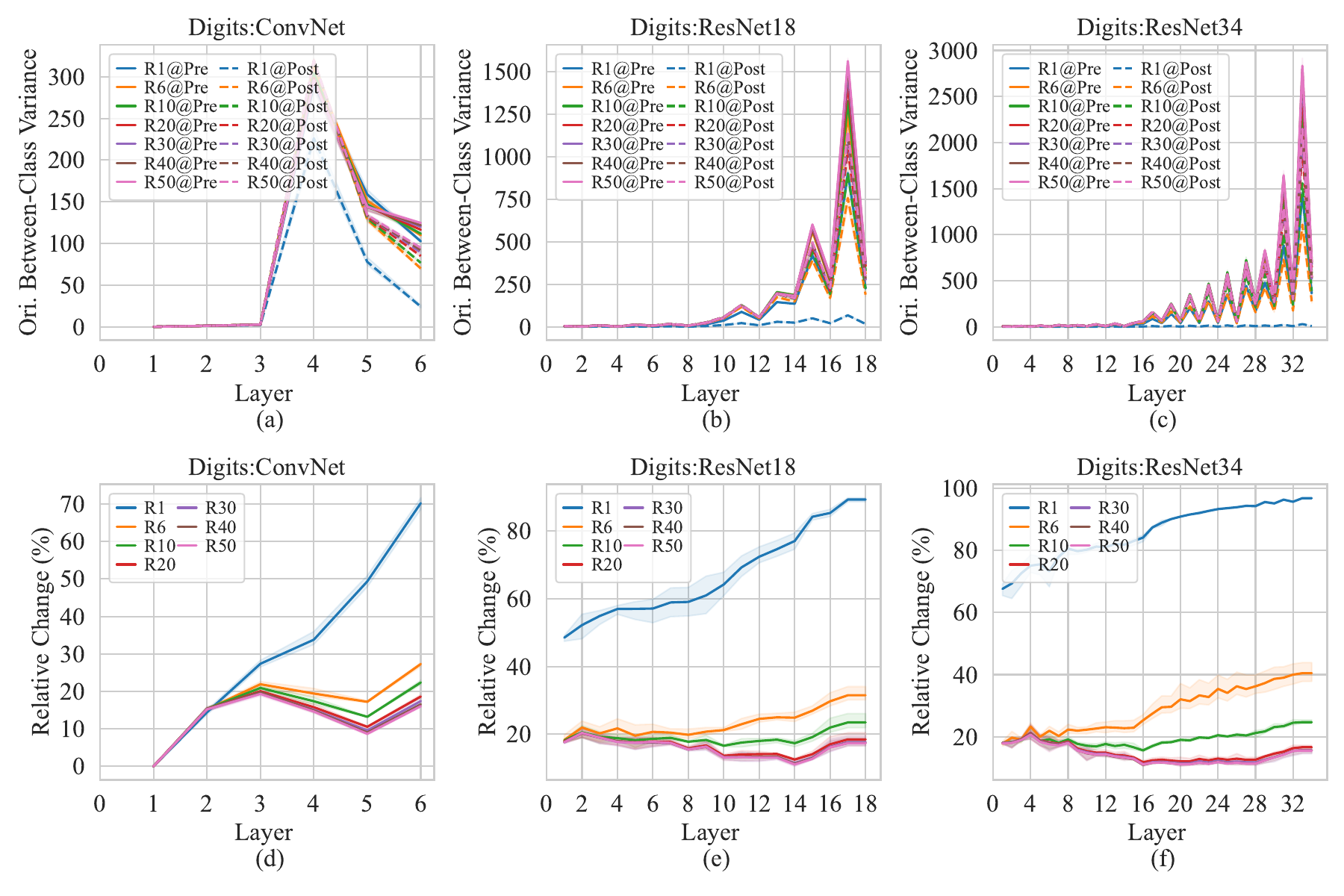}
\caption{Changes in the original unnormalized between-class variance of features across model layers for specific global rounds, with larger X-axis values indicating deeper layers. The model is trained on Digit-Five with multiple models that are randomly initialized. The top half of the figure shows the original unnormalized between-class variance, while the bottom half displays the relative change in variance before and after model aggregation.}
\label{FedAvg_AllModel_Digits_Epochwise_NC1_trace_between}
\end{figure}

\begin{figure}[H]
\centering
\hspace*{-1.8cm}
\includegraphics[width=6.8in]{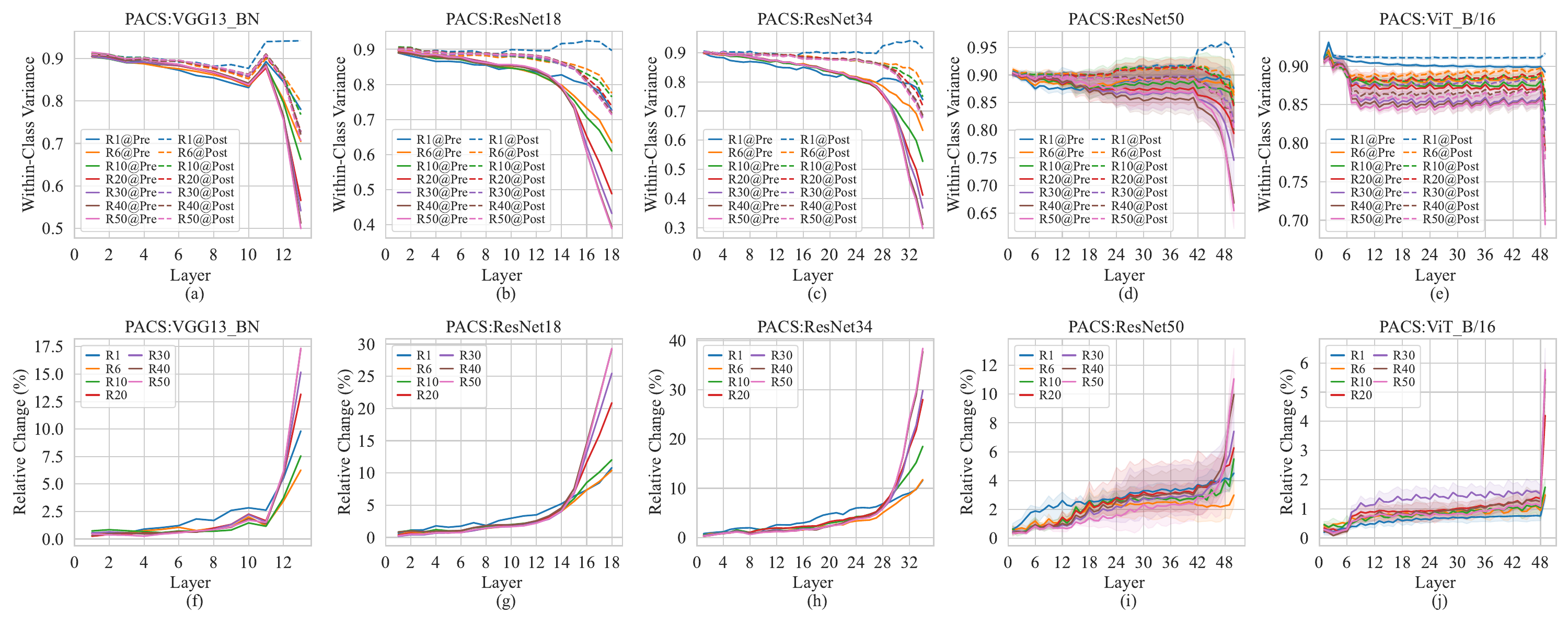}
\caption{Changes in the normalized within-class variance of features across model layers for specific global rounds, with larger X-axis values indicating deeper layers. The model is trained on PACS with multiple models that are randomly initialized. The top half of the figure shows the normalized within-class variance, while the bottom half displays the relative change in variance before and after model aggregation.}
\label{FedAvg_AllModel_PACS_Epochwise_NC1_NC1}
\end{figure}

\begin{figure}[H]
\centering
\hspace*{-1.8cm}
\includegraphics[width=6.8in]{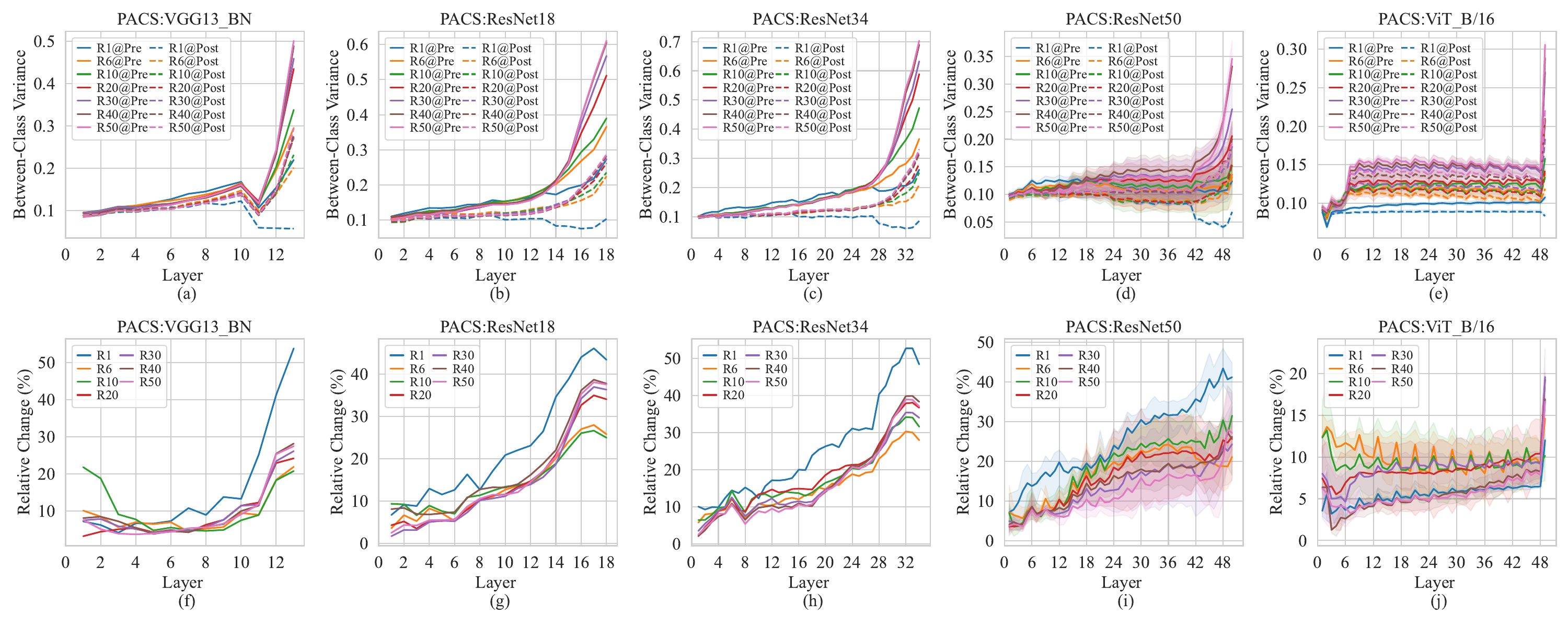}
\caption{Changes in the normalized between-class variance of features across model layers for specific global rounds, with larger X-axis values indicating deeper layers. The model is trained on PACS with multiple models that are randomly initialized. The top half of the figure shows the normalized between-class variance, while the bottom half displays the relative change in variance before and after model aggregation.}
\label{FedAvg_AllModel_PACS_Epochwise_NC1_between}
\end{figure}

\begin{figure}[H]
\centering
\hspace*{-1.8cm}
\includegraphics[width=6.8in]{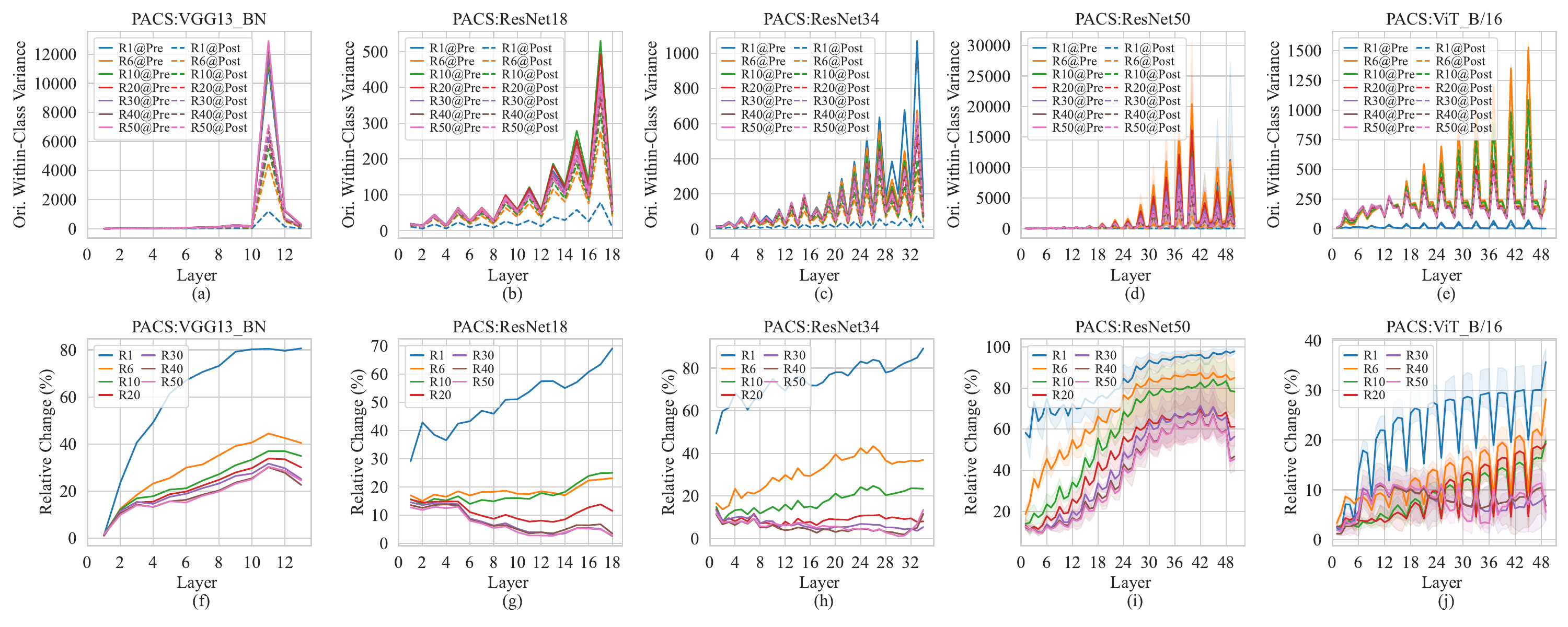}
\caption{Changes in the original unnormalized within-class variance of features across model layers for specific global rounds, with larger X-axis values indicating deeper layers. The model is trained on PACS with multiple models that are randomly initialized. The top half of the figure shows the original unnormalized within-class variance, while the bottom half displays the relative change in variance before and after model aggregation.}
\label{FedAvg_AllModel_PACS_Epochwise_NC1_trace_within}
\end{figure}

\begin{figure}[H]
\centering
\hspace*{-1.8cm}
\includegraphics[width=6.8in]{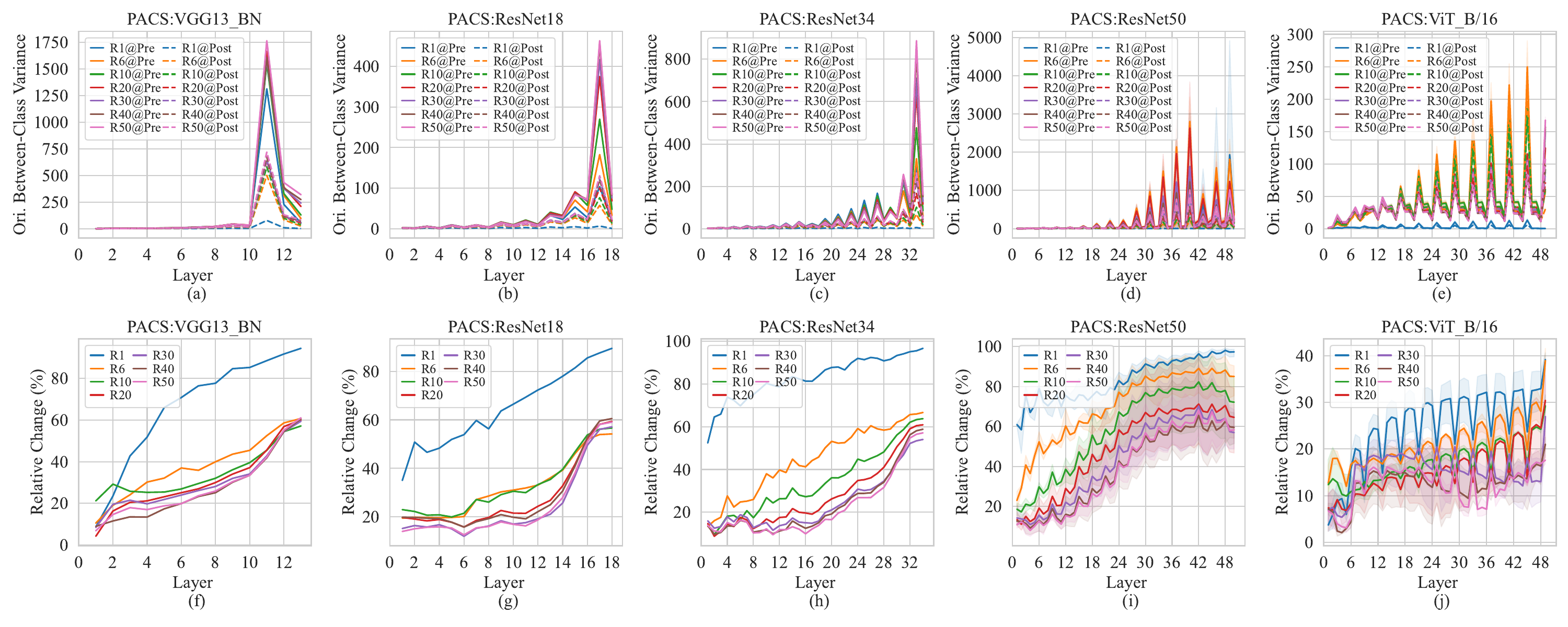}
\caption{Changes in the original unnormalized between-class variance of features across model layers for specific global rounds, with larger X-axis values indicating deeper layers. The model is trained on PACS with multiple models that are randomly initialized. The top half of the figure shows the original unnormalized between-class variance, while the bottom half displays the relative change in variance before and after model aggregation.}
\label{FedAvg_AllModel_PACS_Epochwise_NC1_trace_between}
\end{figure}

\begin{figure}[H]
\centering
\hspace*{-1.8cm}
\includegraphics[width=6.8in]{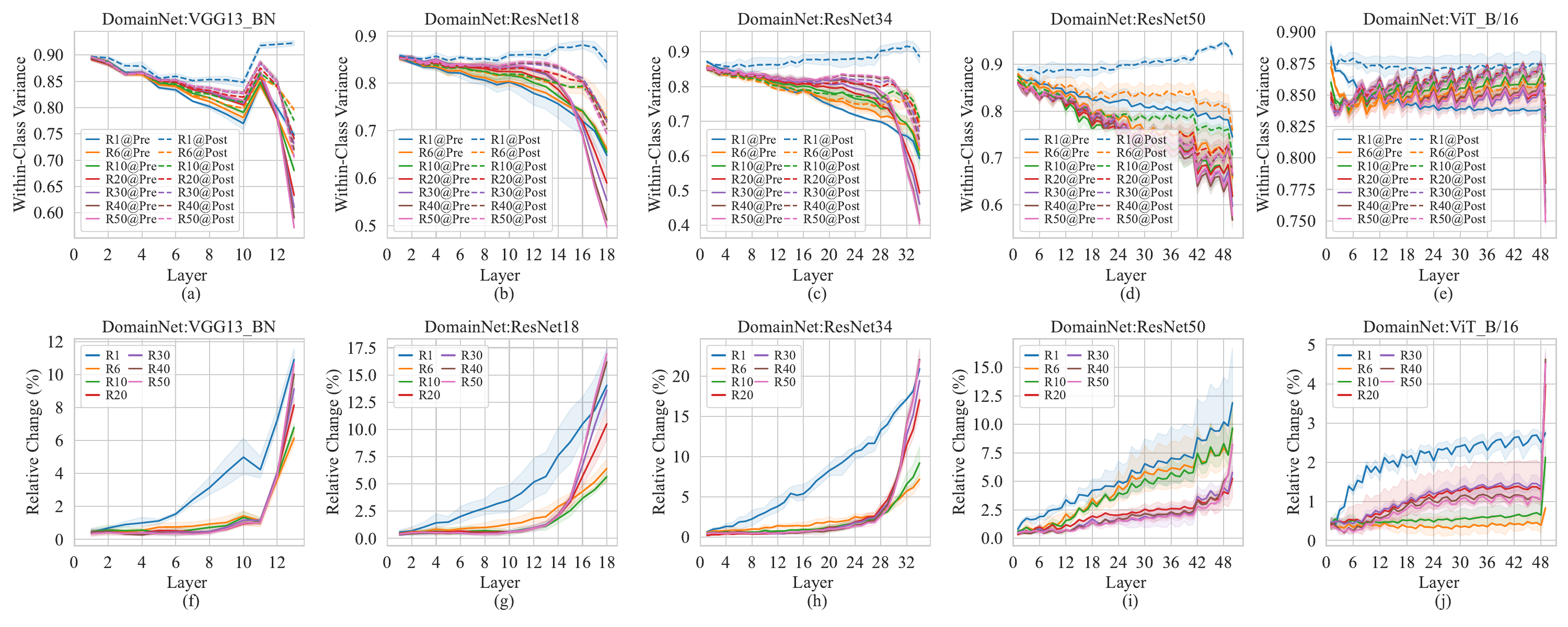}
\caption{Changes in the normalized within-class variance of features across model layers for specific global rounds, with larger X-axis values indicating deeper layers. The model is trained on DomainNet with multiple models that are randomly initialized. The top half of the figure shows the normalized within-class variance, while the bottom half displays the relative change in variance before and after model aggregation.}
\label{FedAvg_AllModel_DomainNet_Epochwise_NC1_NC1}
\end{figure}

\begin{figure}[H]
\centering
\hspace*{-1.8cm}
\includegraphics[width=6.8in]{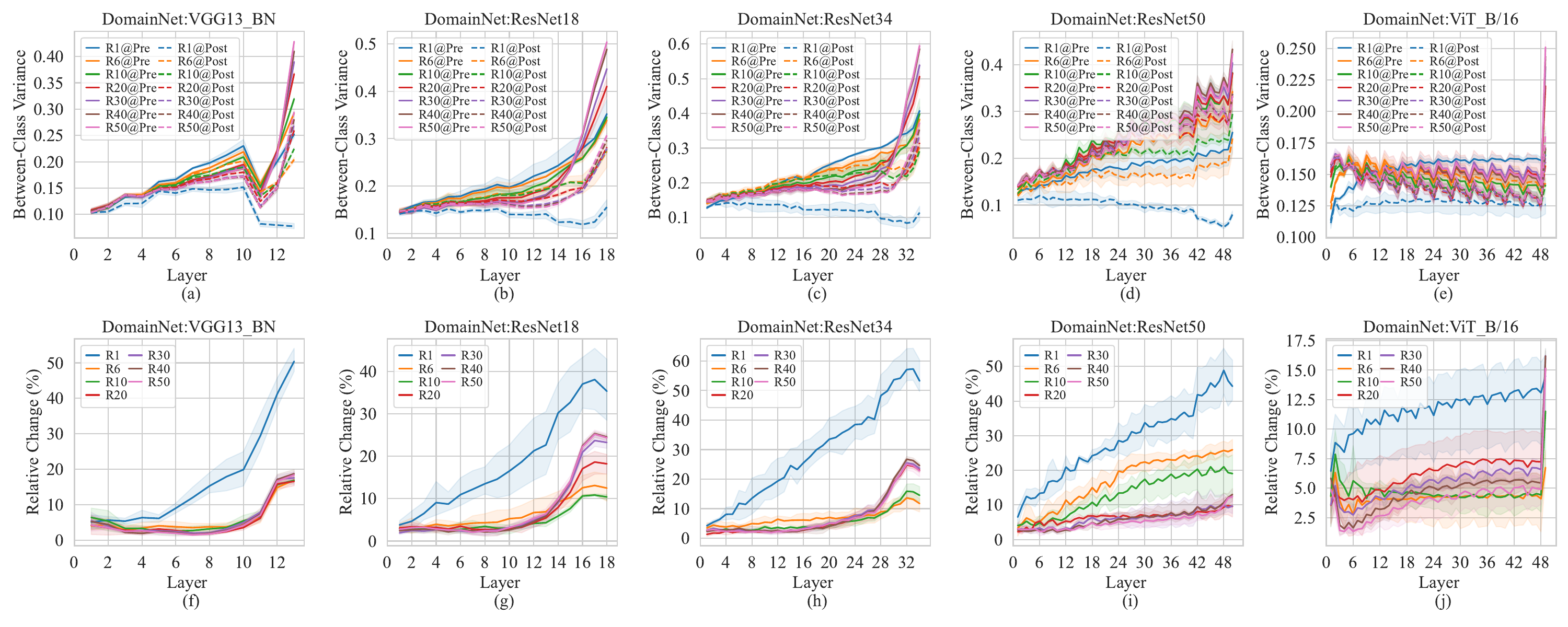}
\caption{Changes in the normalized between-class variance of features across model layers for specific global rounds, with larger X-axis values indicating deeper layers. The model is trained on DomainNet with multiple models that are randomly initialized. The top half of the figure shows the normalized between-class variance, while the bottom half displays the relative change in variance before and after model aggregation.}
\label{FedAvg_AllModel_DomainNet_Epochwise_NC1_between}
\end{figure}

\begin{figure}[H]
\centering
\hspace*{-1.8cm}
\includegraphics[width=6.8in]{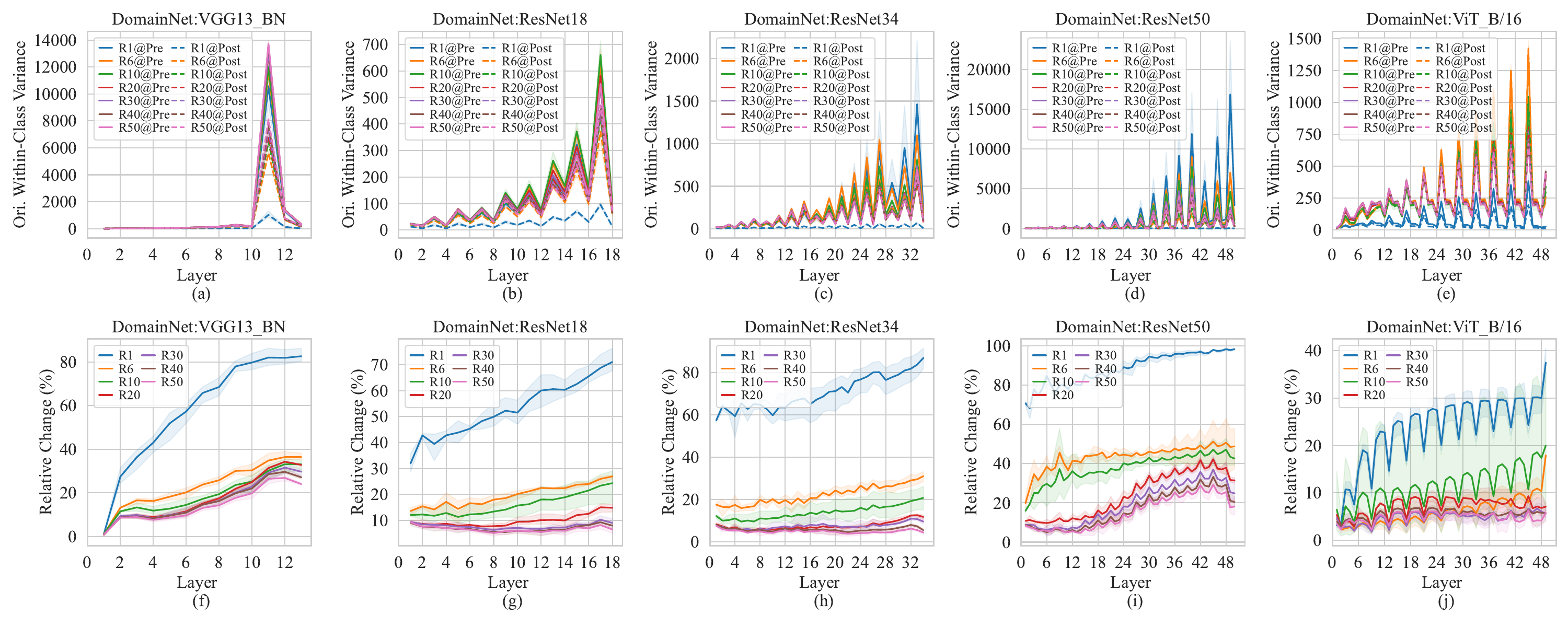}
\caption{Changes in the unnormalized within-class variance of features across model layers for specific global rounds, with larger X-axis values indicating deeper layers. The model is trained on DomainNet with multiple models that are randomly initialized. The top half of the figure shows the original unnormalized within-class variance, while the bottom half displays the relative change in variance before and after model aggregation.}
\label{FedAvg_AllModel_DomainNet_Epochwise_NC1_trace_within}
\end{figure}

\begin{figure}[H]
\centering
\hspace*{-1.8cm}
\includegraphics[width=6.8in]{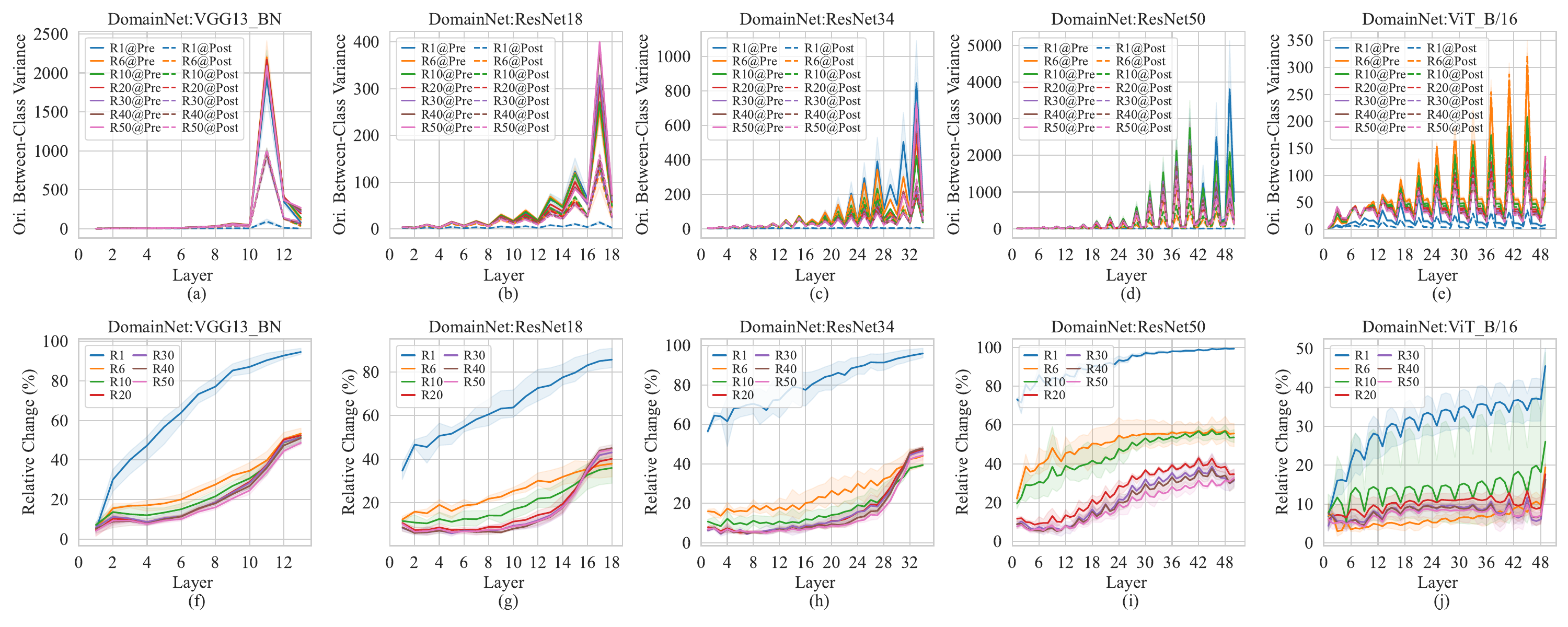}
\caption{Changes in the unnormalized between-class variance of features across model layers for specific global rounds, with larger X-axis values indicating deeper layers. The model is trained on DomainNet with multiple models that are randomly initialized. The top half of the figure shows the original unnormalized between-class variance, while the bottom half displays the relative change in variance before and after model aggregation.}
\label{FedAvg_AllModel_DomainNet_Epochwise_NC1_trace_between}
\end{figure}

\subsection{Changes of Feature Variance Across Training Rounds}

\begin{figure}[H]
\centering
\includegraphics[width=4.5in]{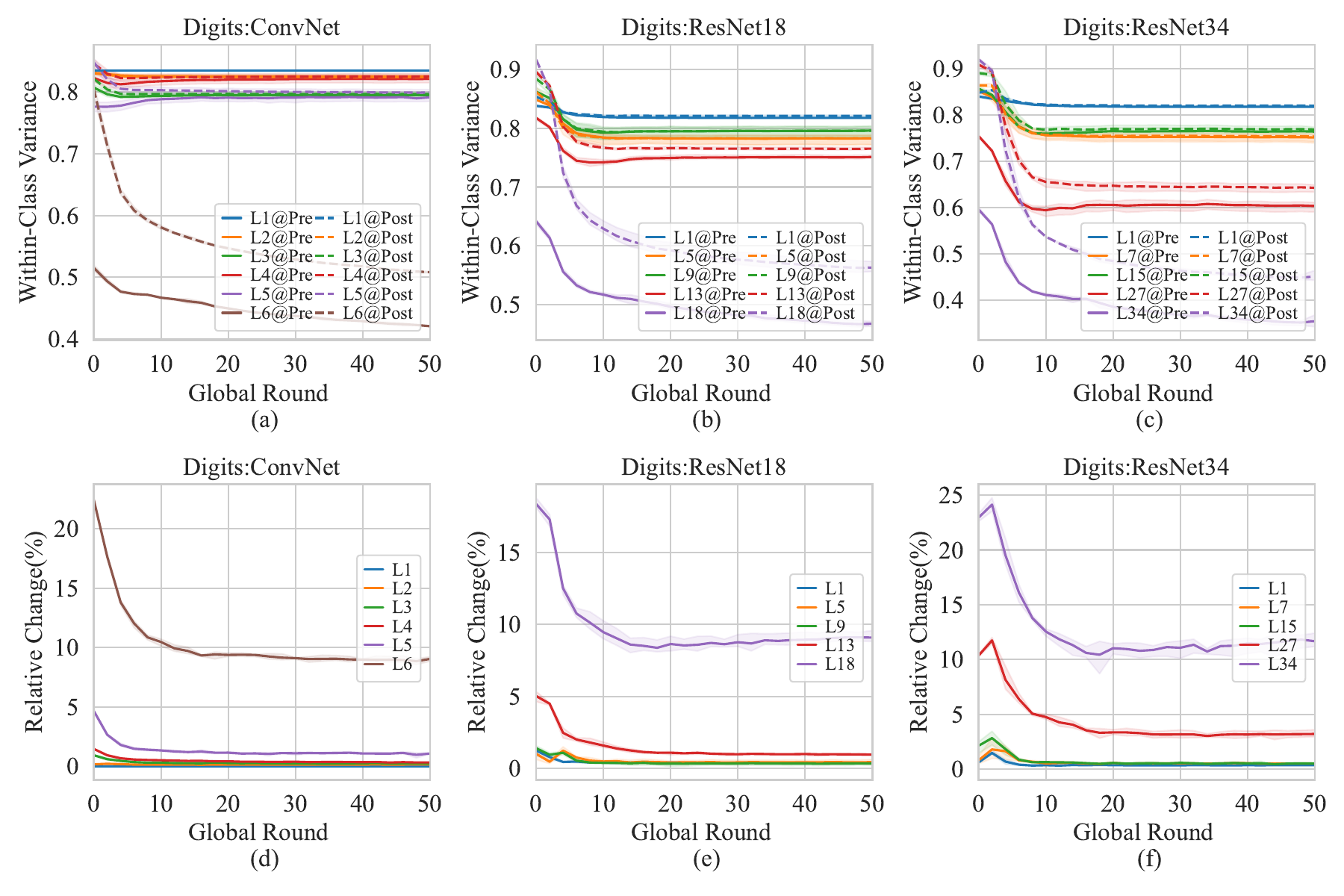}
\caption{Changes in the normalized within-class variance of features across FL training at specific model layers. The model is trained on Digit-Five with multiple models that are randomly initialized. The top half of the figure shows the normalized within-class variance, while the bottom half displays the relative change in variance before and after model aggregation.}
\label{FedAvg_AllModel_Digits_Layerwise_NC1_NC1}
\end{figure}

\begin{figure}[H]
\centering
\includegraphics[width=4.5in]{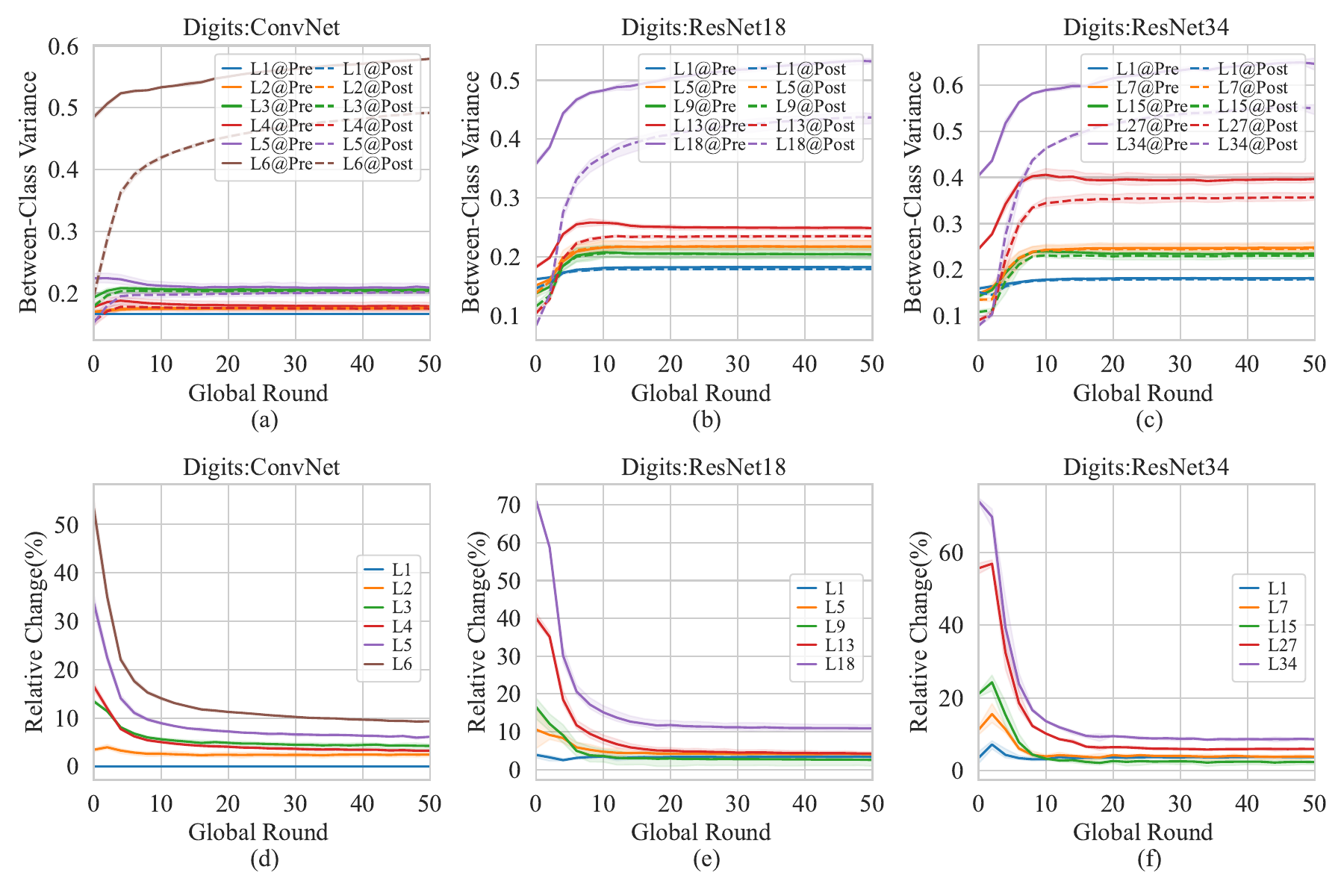}
\caption{Changes in the normalized between-class variance of features across FL training at specific model layers. The model is trained on Digit-Five with multiple models that are randomly initialized. The top half of the figure shows the normalized between-class variance, while the bottom half displays the relative change in variance before and after model aggregation.}
\label{FedAvg_AllModel_Digits_Layerwise_NC1_between}
\end{figure}

\begin{figure}[H]
\centering
\includegraphics[width=4.5in]{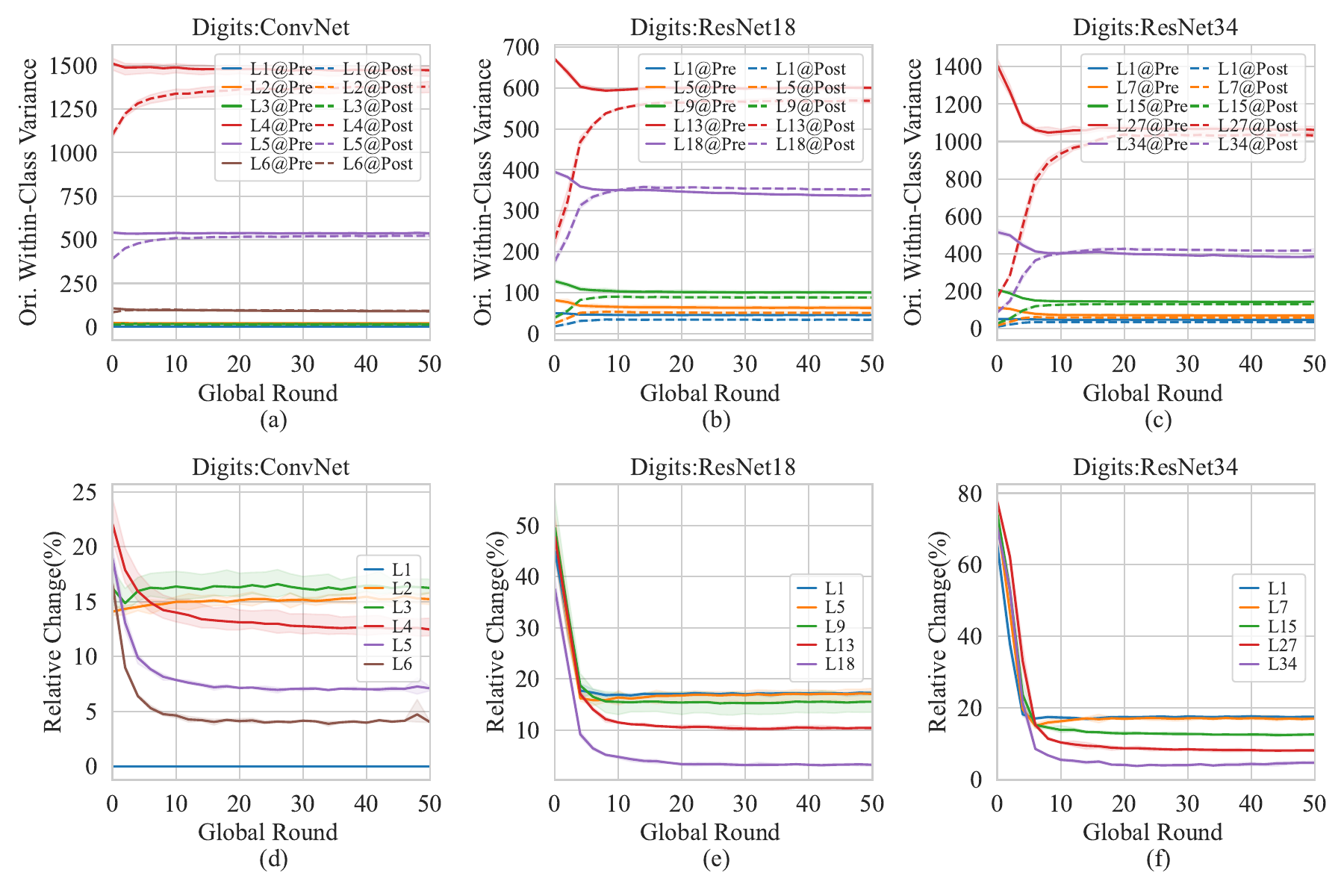}
\caption{Changes in the original unnormalized within-class variance of features across FL training at specific model layers. The model is trained on Digit-Five with multiple models that are randomly initialized. The top half of the figure shows the original unnormalized within-class variance, while the bottom half displays the relative change in variance before and after model aggregation.}
\label{FedAvg_AllModel_Digits_Layerwise_NC1_trace_within}
\end{figure}

\begin{figure}[H]
\centering
\includegraphics[width=4.5in]{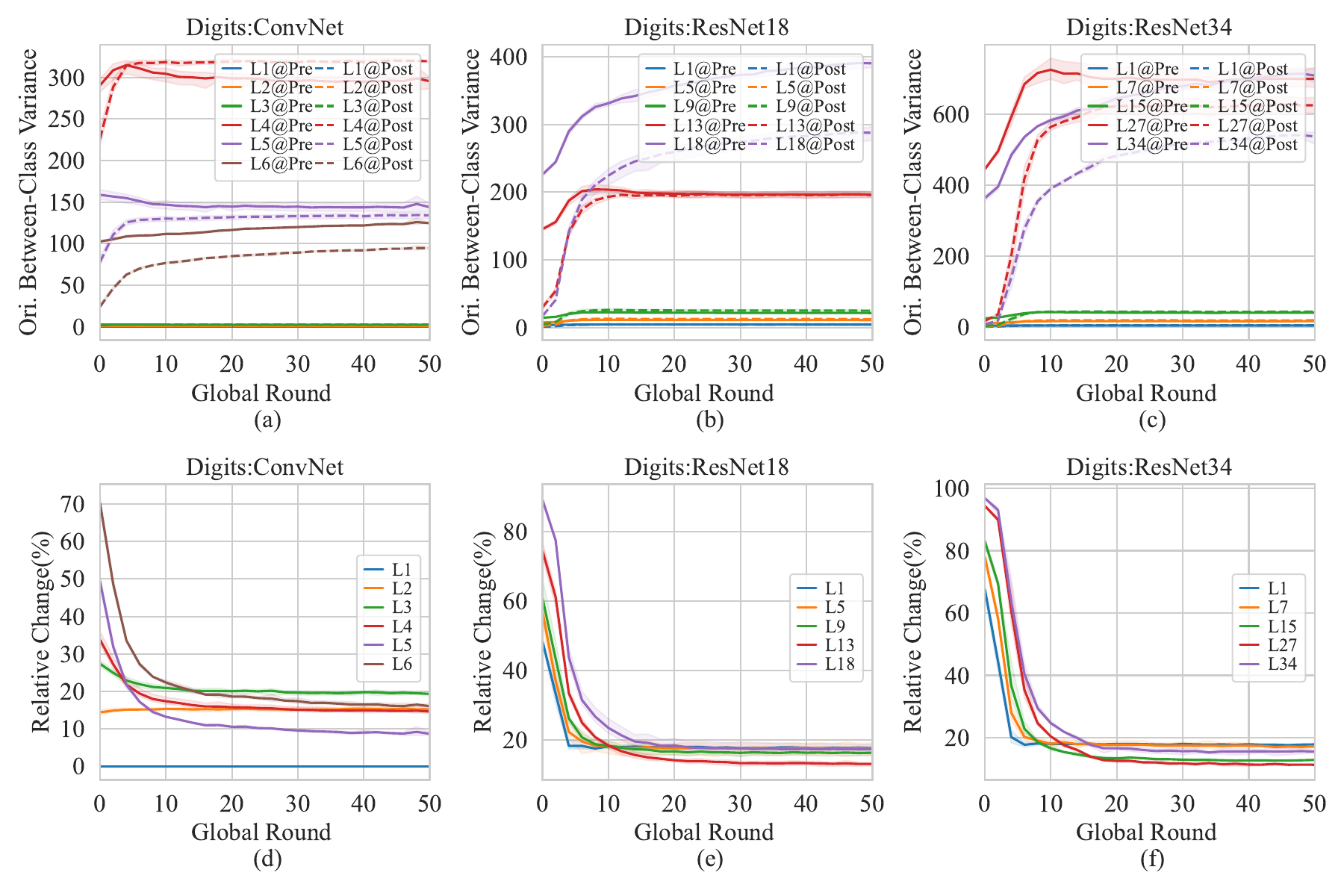}
\caption{Changes in the original unnormalized between-class variance of features across FL training at specific model layers. The model is trained on Digit-Five with multiple models that are randomly initialized. The top half of the figure shows the original unnormalized between-class variance, while the bottom half displays the relative change in variance before and after model aggregation.}
\label{FedAvg_AllModel_Digits_Layerwise_NC1_trace_between}
\end{figure}

\begin{figure}[H]
\centering
\hspace*{-1.8cm}
\includegraphics[width=6.8in]{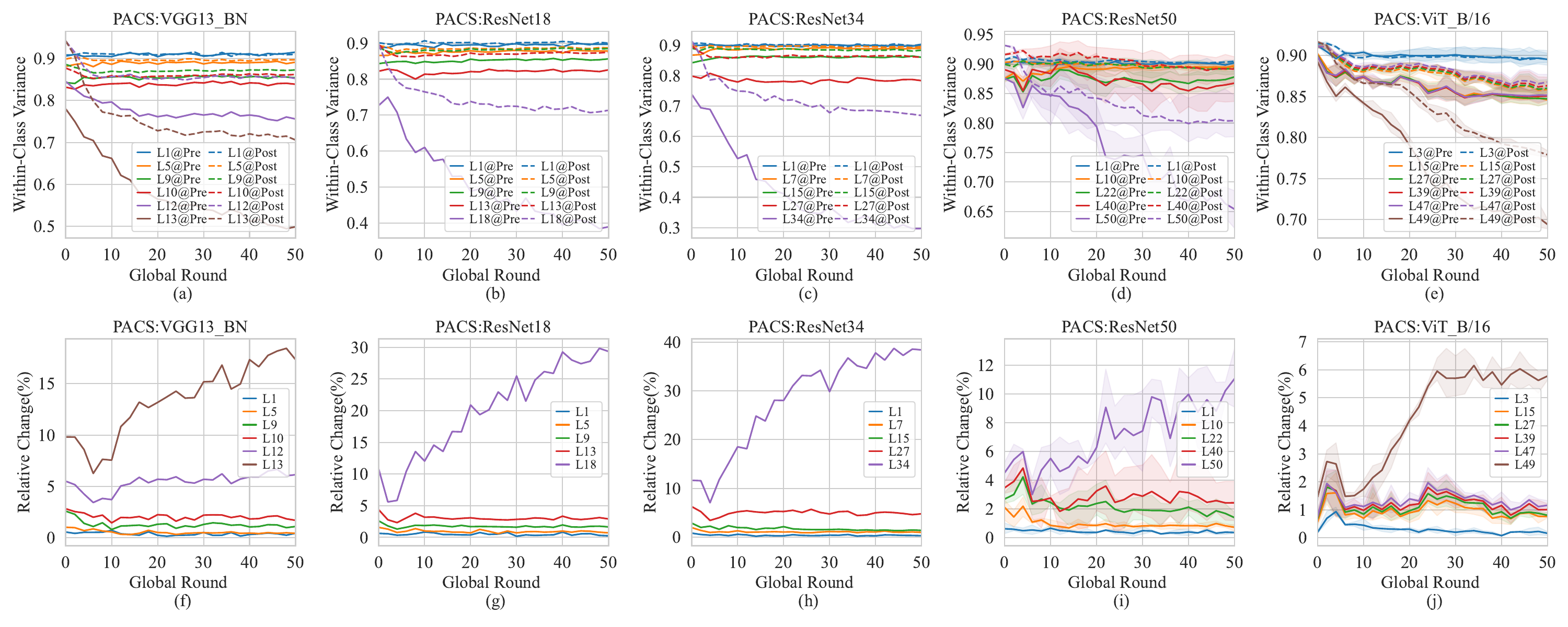}
\caption{Changes in the normalized within-class variance of features across FL training at specific model layers. The model is trained on PACS with multiple models that are randomly initialized. The top half of the figure shows the normalized within-class variance, while the bottom half displays the relative change in variance before and after model aggregation.}
\label{FedAvg_AllModel_PACS_Layerwise_NC1_NC1}
\end{figure}

\begin{figure}[H]
\centering
\hspace*{-1.8cm}
\includegraphics[width=6.8in]{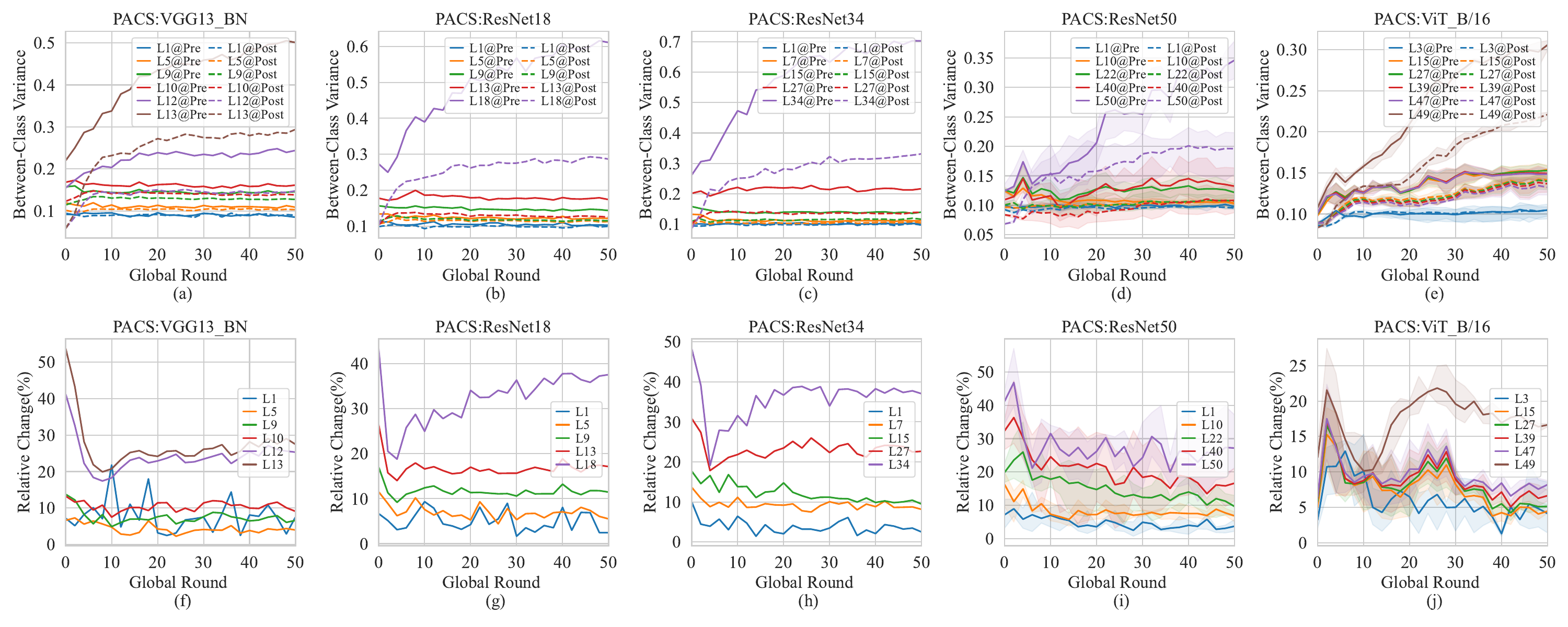}
\caption{Changes in the normalized between-class variance of features across FL training at specific model layers. The model is trained on PACS with multiple models that are randomly initialized. The top half of the figure shows the normalized between-class variance, while the bottom half displays the relative change in variance before and after model aggregation.}
\label{FedAvg_AllModel_PACS_Layerwise_NC1_between}
\end{figure}

\begin{figure}[H]
\centering
\hspace*{-1.8cm}
\includegraphics[width=6.8in]{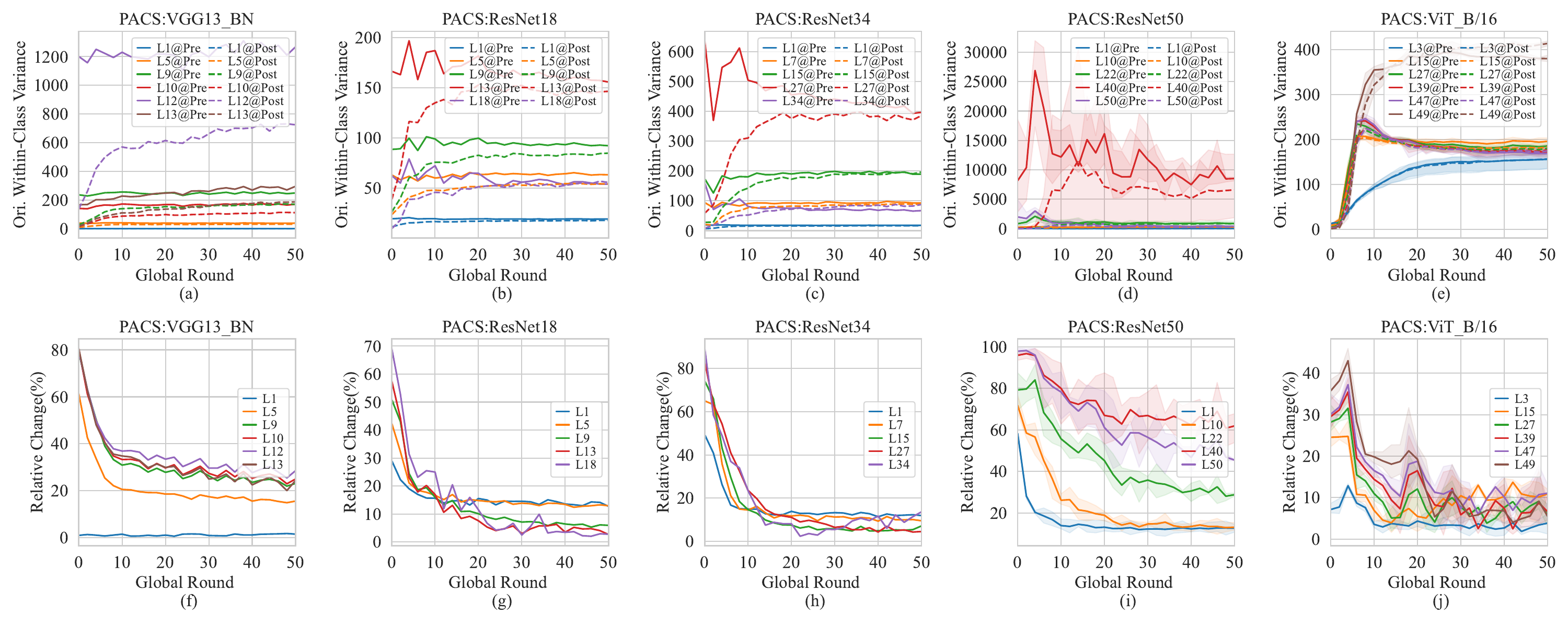}
\caption{Changes in the original unnormalized within-class variance of features across FL training at specific model layers. The model is trained on PACS with multiple models that are randomly initialized. The top half of the figure shows the original unnormalized within-class variance, while the bottom half displays the relative change in variance before and after model aggregation.}
\label{FedAvg_AllModel_PACS_Layerwise_NC1_trace_within}
\end{figure}

\begin{figure}[H]
\centering
\hspace*{-1.8cm}
\includegraphics[width=6.8in]{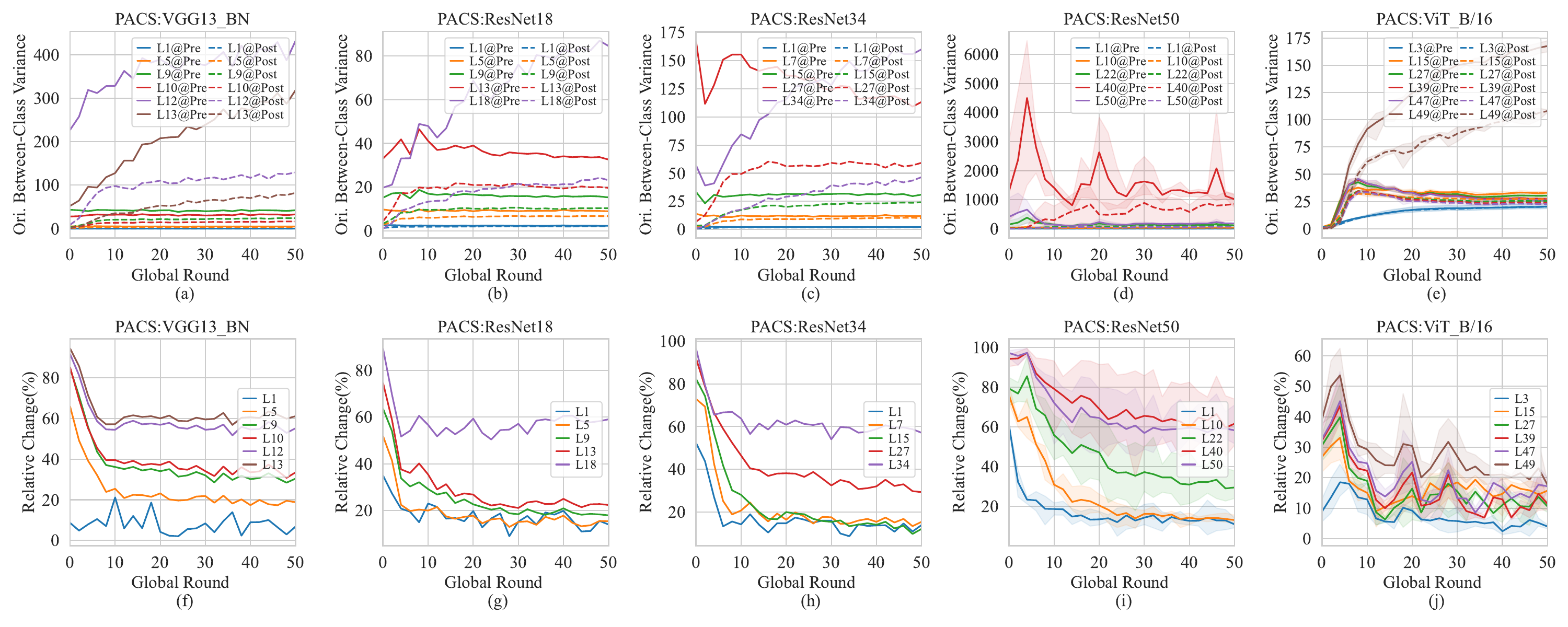}
\caption{Changes in the original unnormalized between-class variance of features across FL training at specific model layers. The model is trained on PACS with multiple models that are randomly initialized. The top half of the figure shows the original unnormalized between-class variance, while the bottom half displays the relative change in variance before and after model aggregation.}
\label{FedAvg_AllModel_PACS_Layerwise_NC1_trace_between}
\end{figure}

\begin{figure}[H]
\centering
\hspace*{-1.8cm}
\includegraphics[width=6.8in]{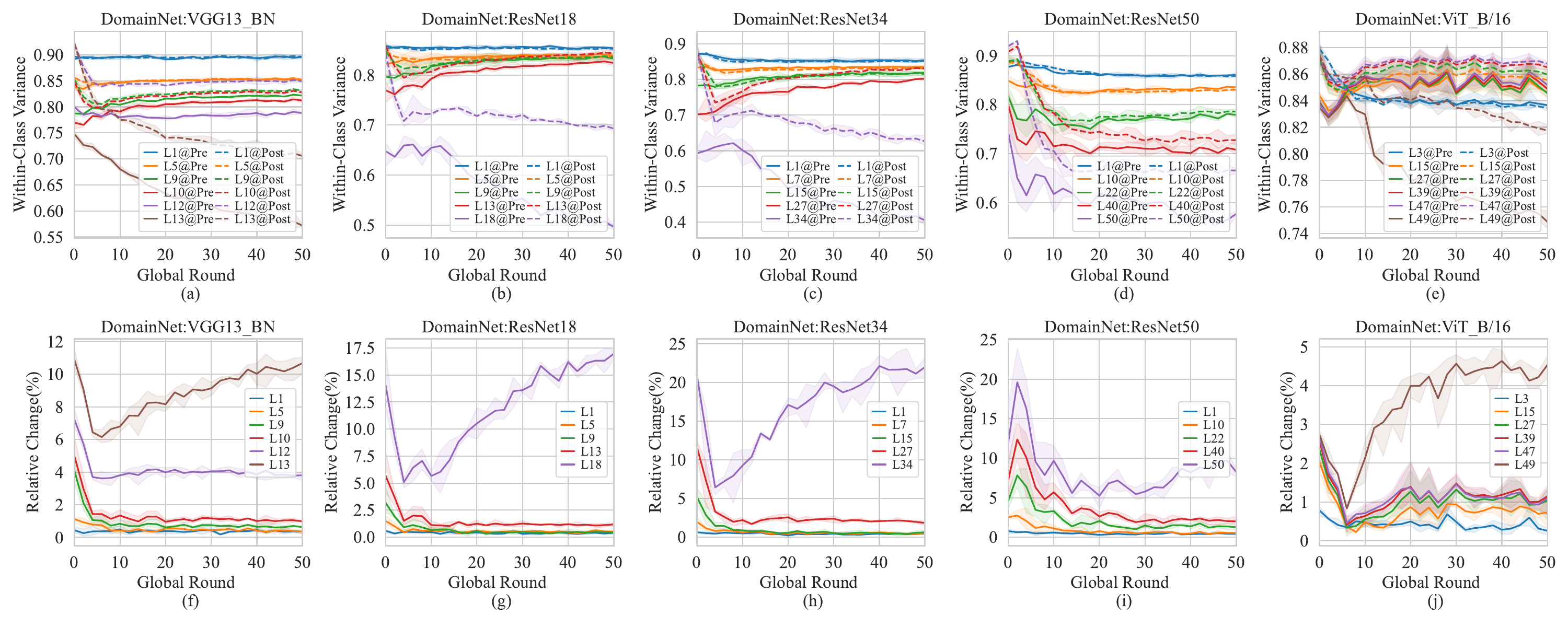}
\caption{Changes in the normalized within-class variance of features across FL training at specific model layers. The model is trained on DomainNet with multiple models that are randomly initialized. The top half of the figure shows the normalized within-class variance, while the bottom half displays the relative change in variance before and after model aggregation.}
\label{FedAvg_AllModel_DomainNet_Layerwise_NC1_NC1}
\end{figure}

\begin{figure}[H]
\centering
\hspace*{-1.8cm}
\includegraphics[width=6.8in]{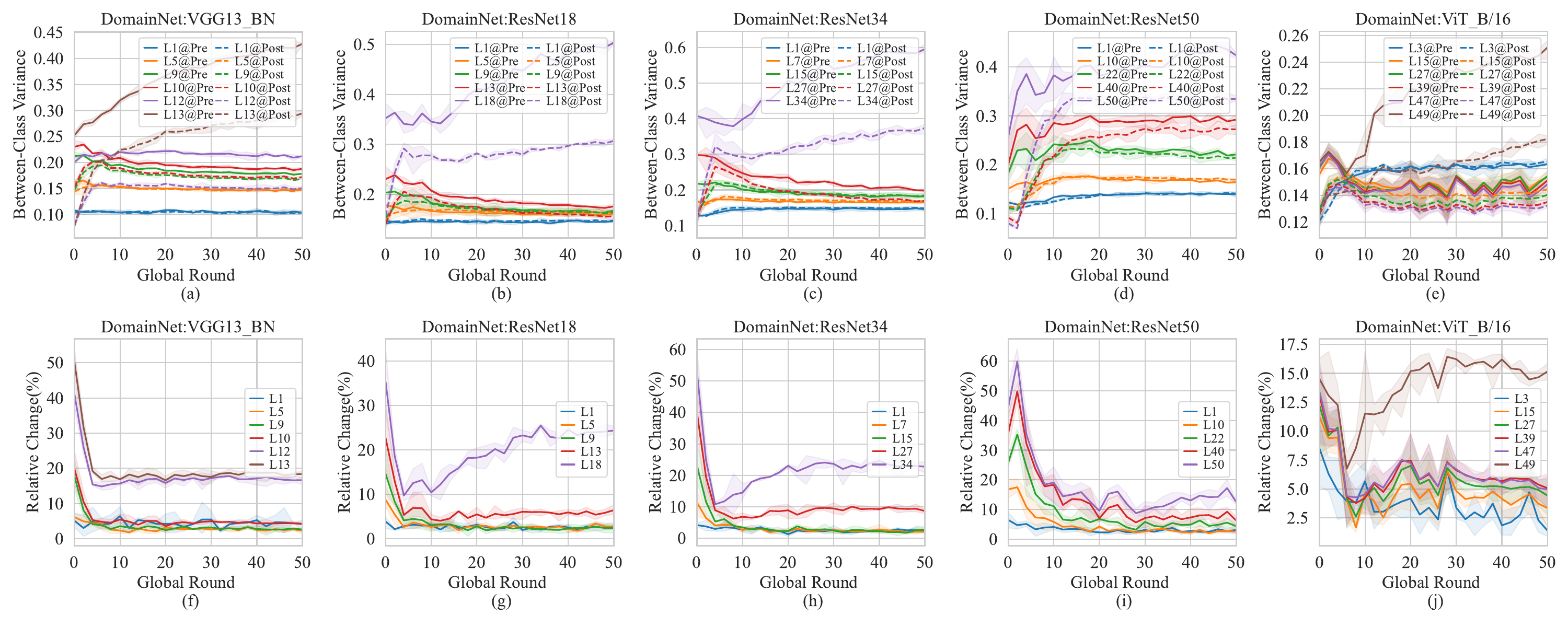}
\caption{Changes in the normalized between-class variance of features across FL training at specific model layers. The model is trained on DomainNet with multiple models that are randomly initialized. The top half of the figure shows the normalized between-class variance, while the bottom half displays the relative change in variance before and after model aggregation.}
\label{FedAvg_AllModel_DomainNet_Layerwise_NC1_between}
\end{figure}

\begin{figure}[H]
\centering
\hspace*{-1.8cm}
\includegraphics[width=6.8in]{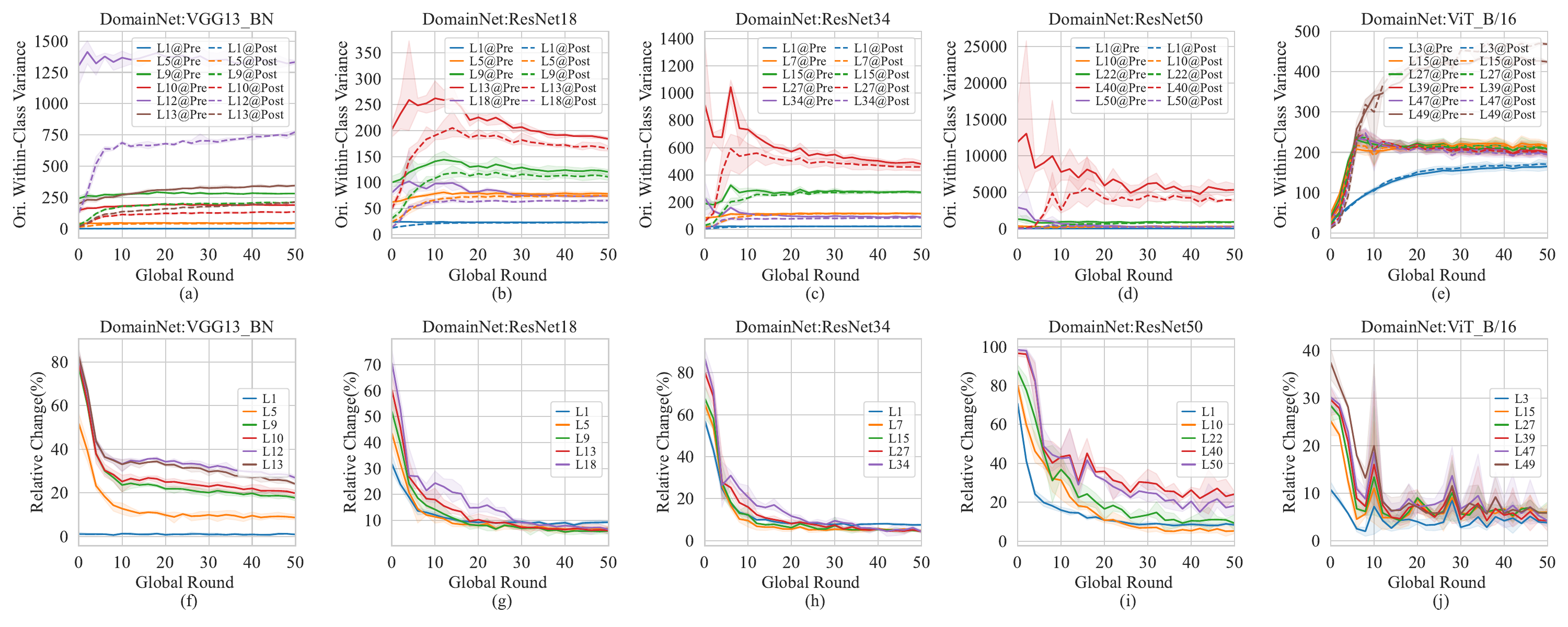}
\caption{Changes in the original unnormalized within-class variance of features across FL training at specific model layers. The model is trained on DomainNet with multiple models that are randomly initialized. The top half of the figure shows the original unnormalized within-class variance, while the bottom half displays the relative change in variance before and after model aggregation.}
\label{FedAvg_AllModel_DomainNet_Layerwise_NC1_trace_within}
\end{figure}

\begin{figure}[H]
\centering
\hspace*{-1.8cm}
\includegraphics[width=6.8in]{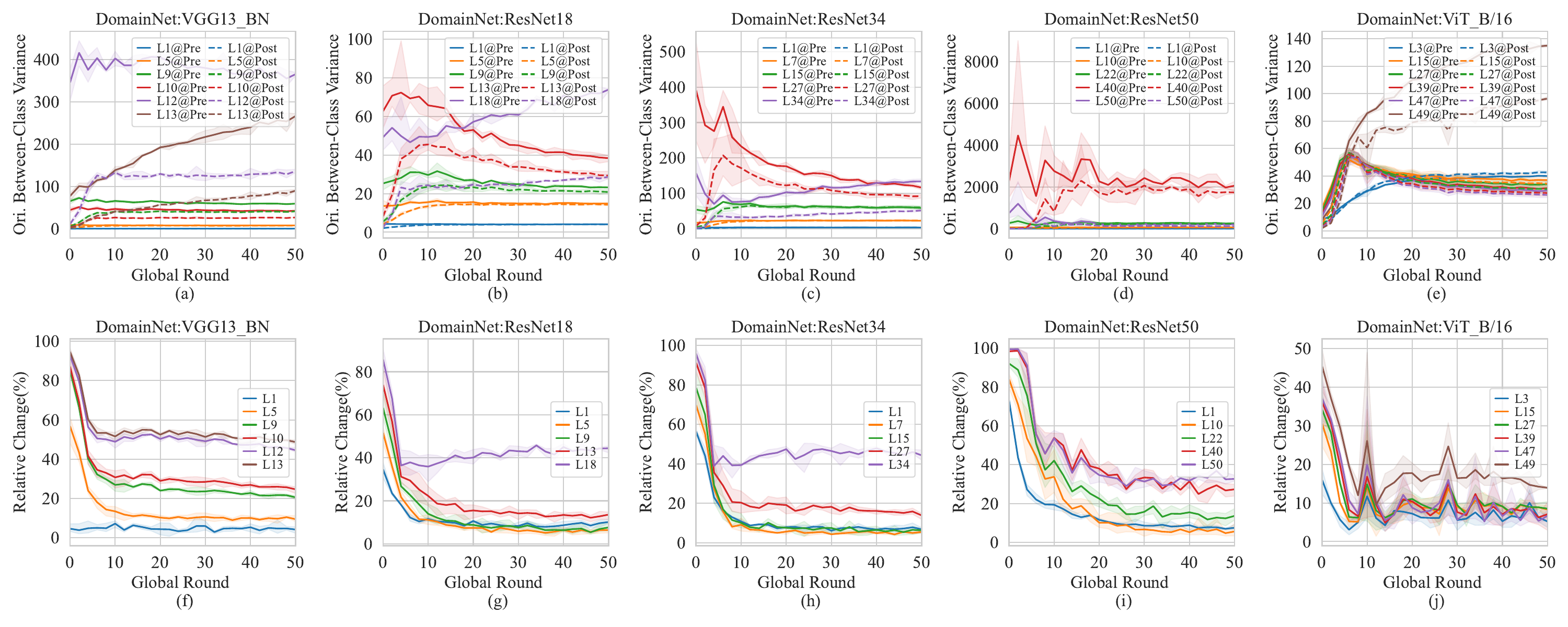}
\caption{Changes in the original unnormalized between-class variance of features across FL training at specific model layers. The model is trained on DomainNet with multiple models that are randomly initialized. The top half of the figure shows the original unnormalized between-class variance, while the bottom half displays the relative change in variance before and after model aggregation.}
\label{FedAvg_AllModel_DomainNet_Layerwise_NC1_trace_between}
\end{figure}

\newpage
\section{Detailed Results of Alignment between Features and Parameters}\label{appendix_alignment}
\vspace{-0.5cm}
In this section, we present the experimental results corresponding to the alignment of features and the parameters of the subsequent layers.
From these results, we observe that the alignment between features and parameters improves as training progresses, with the alignment of penultimate layer features and the classifier increasing more rapidly.

After model aggregation, the alignment between features and parameters tends to decrease. However, the decrease is more pronounced in the classifier. This increased mismatch between penultimate layer features and the classifier, along with the degradation of the penultimate layer features, causes the aggregated model to perform significantly worse when sent back to each client.
\vspace{-0.5cm}
\subsection{Changes of Alignment Across Layers}
\vspace{-0.5cm}
\begin{figure}[H]
\centering
\includegraphics[width=4.0in]{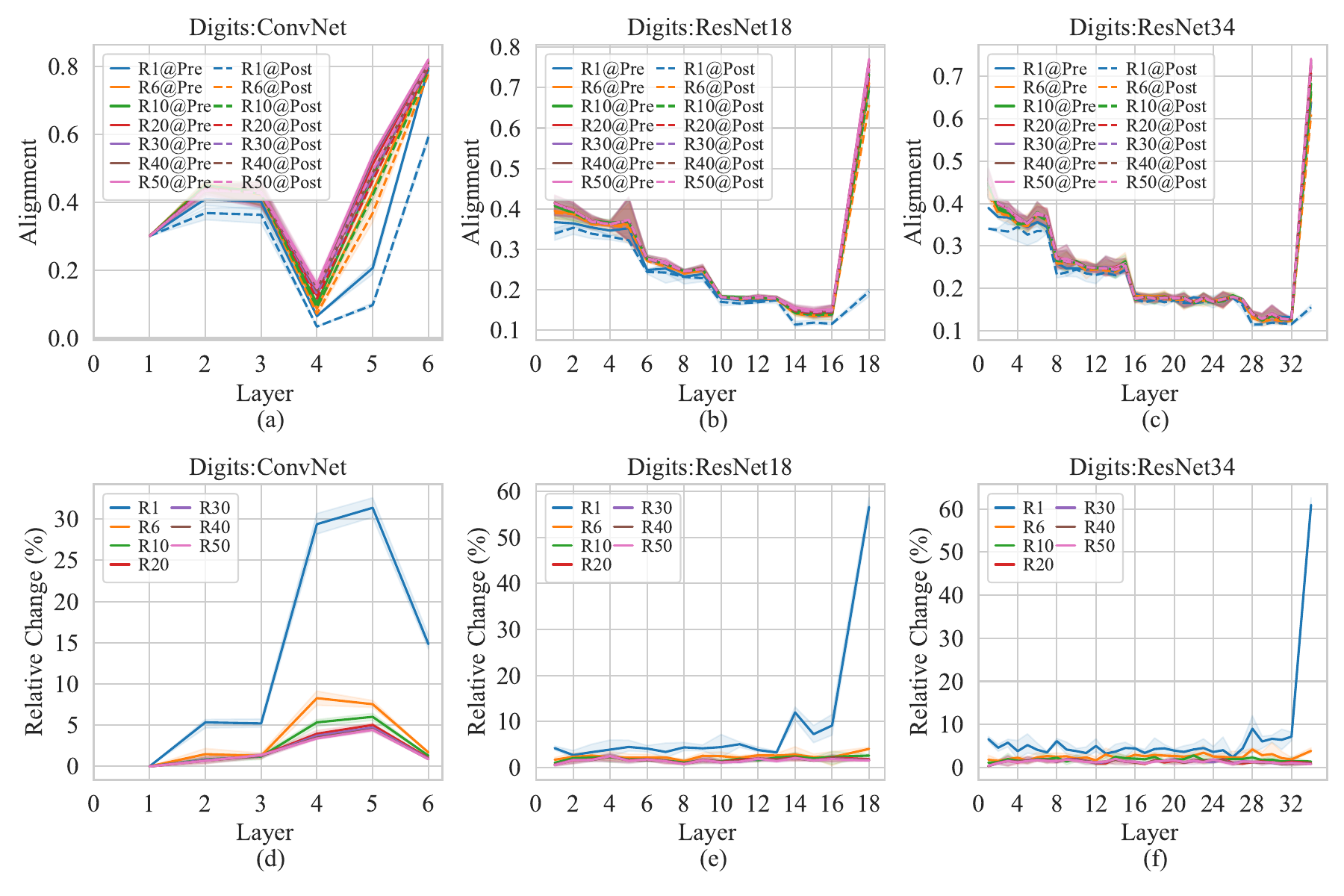}
\caption{Changes in the alignment between features and parameters across model layers for specific global rounds, with larger X-axis values indicating deeper layers. The model is trained on Digit-Five with multiple models that are randomly initialized. The top half of the figure shows the original alignment values between features and parameters, while the bottom half displays the relative change in alignment before and after model aggregation.}
\label{FedAvg_AllModel_Digits_Epochwise_NC3_NC3}
\end{figure}

\vspace{-0.5cm}
\begin{figure}[H]
\centering
\vspace{-0.5cm}
\hspace*{-1.8cm}
\includegraphics[width=6.8in]{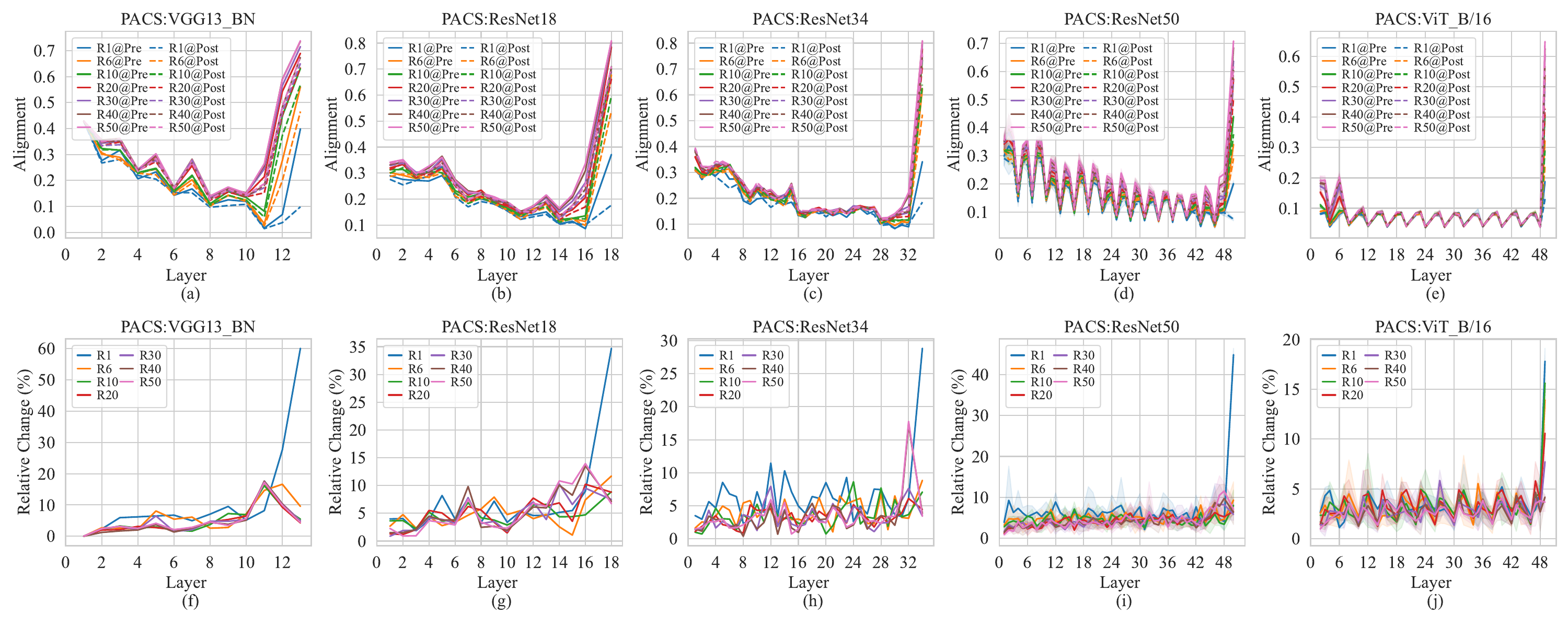}
\vspace{-0.5cm}
\caption{Changes in the alignment between features and parameters across model layers for specific global rounds, with larger X-axis values indicating deeper layers. The model is trained on PACS with multiple models that are randomly initialized. The top half of the figure shows the original alignment values between features and parameters, while the bottom half displays the relative change in alignment before and after model aggregation.}
\label{FedAvg_AllModel_PACS_Epochwise_NC3_NC3}
\end{figure}

\begin{figure}[H]
\centering
\hspace*{-1.8cm}
\includegraphics[width=6.8in]{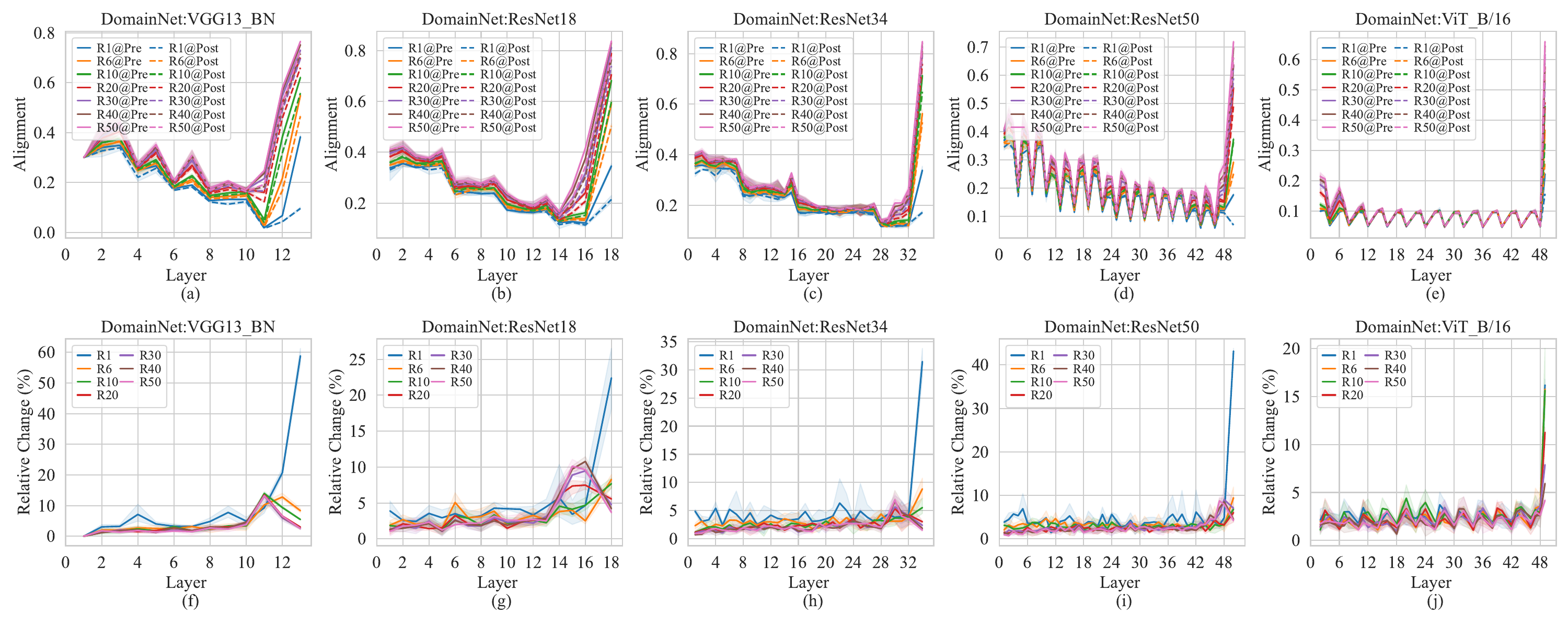}
\caption{Changes in the alignment between features and parameters across model layers for specific global rounds, with larger X-axis values indicating deeper layers. The model is trained on DomainNet with multiple models that are randomly initialized. The top half of the figure shows the original alignment values between features and parameters, while the bottom half displays the relative change in alignment before and after model aggregation.}
\label{FedAvg_AllModel_DomainNet_Epochwise_NC3_NC3}
\end{figure}

\subsection{Changes of Alignment Across Training Rounds}

\begin{figure}[H]
\centering
\includegraphics[width=4.5in]{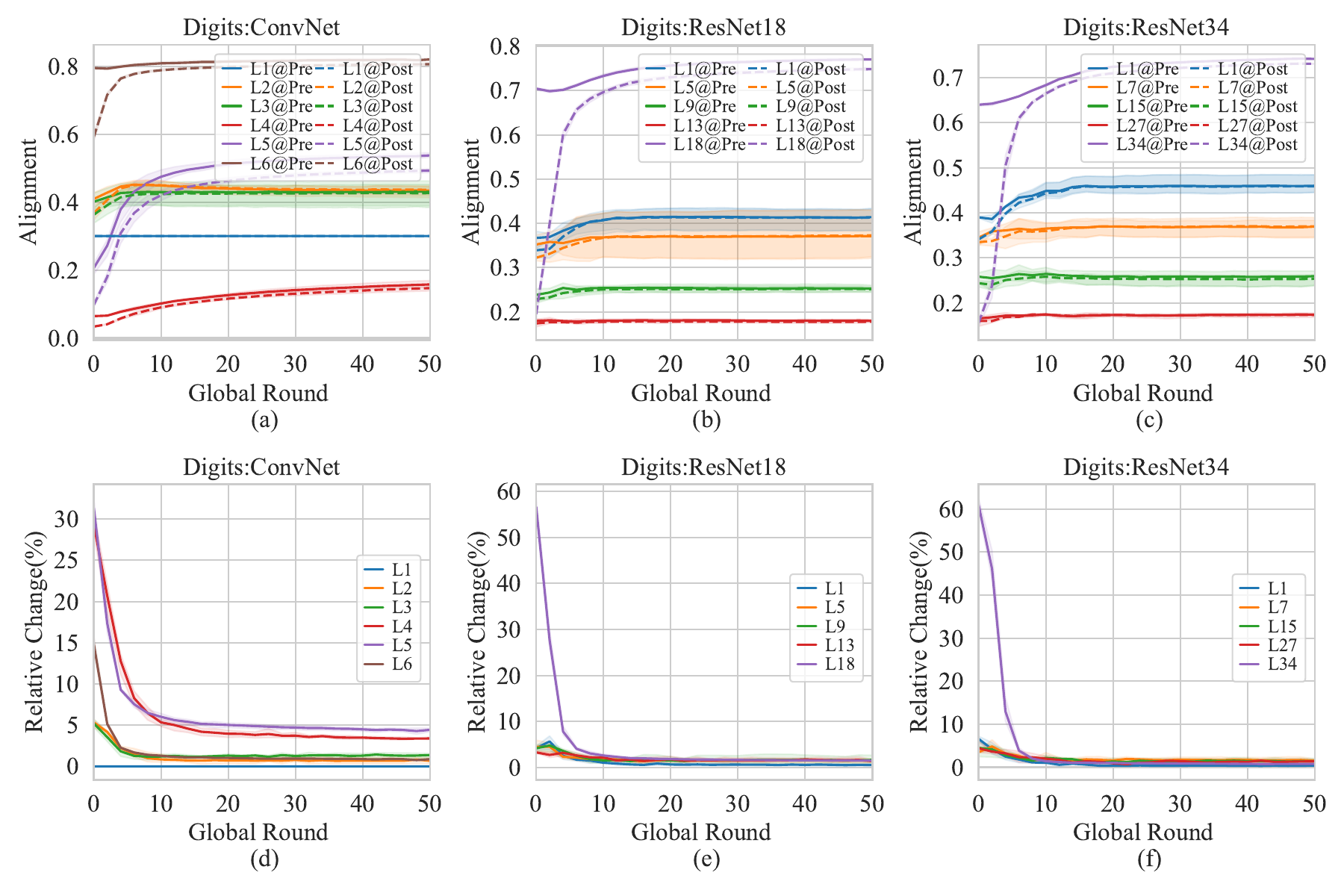}
\caption{Changes in the alignment between features and parameters across FL training at specific model layers. The model is trained on Digit-Five with multiple models that are randomly initialized. The top half of the figure shows the original alignment values between features and parameters, while the bottom half displays the relative change in alignment before and after model aggregation.}
\label{FedAvg_AllModel_Digits_Layerwise_NC3_NC3}
\end{figure}

\begin{figure}[H]
\centering
\hspace*{-1.8cm}
\includegraphics[width=6.8in]{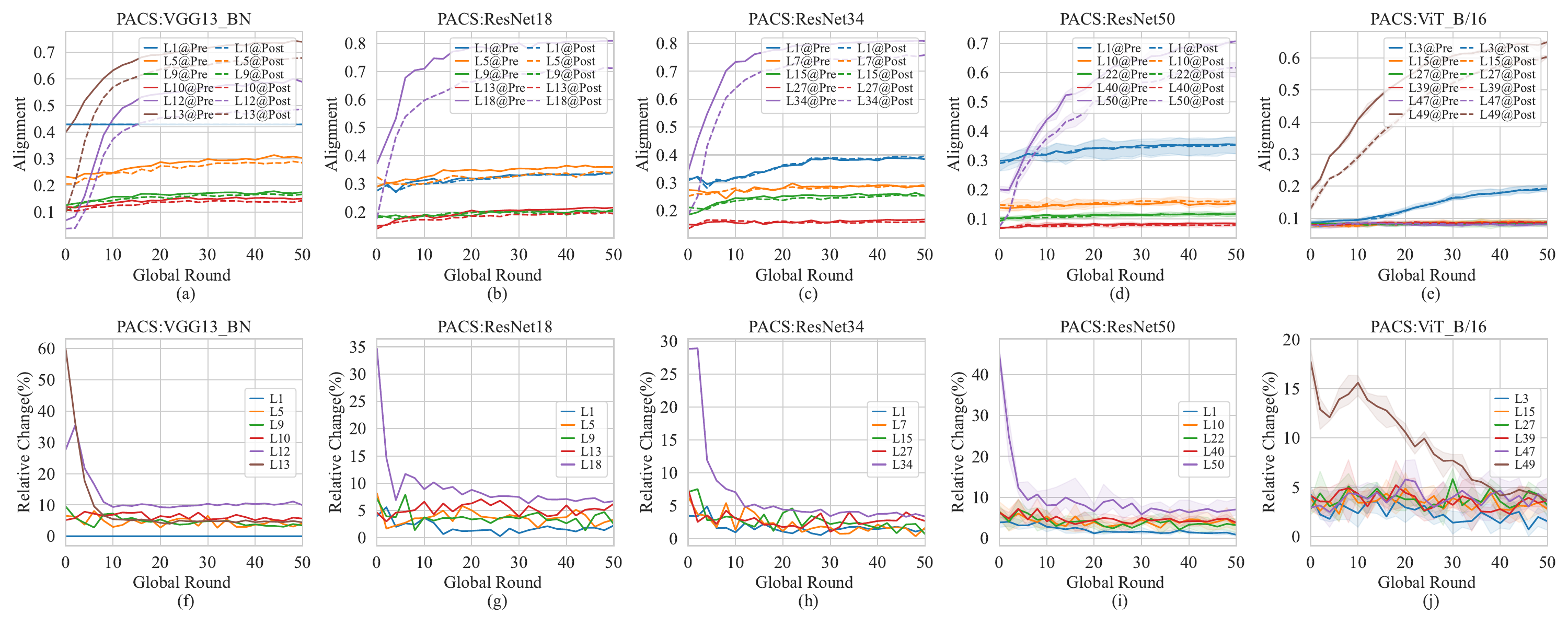}
\caption{Changes in the alignment between features and parameters across FL training at specific model layers. The model is trained on PACS with multiple models that are randomly initialized. The top half of the figure shows the original alignment values between features and parameters, while the bottom half displays the relative change in alignment before and after model aggregation.}
\label{FedAvg_AllModel_PACS_Layerwise_NC3_NC3}
\end{figure}

\begin{figure}[H]
\centering
\hspace*{-1.8cm}
\includegraphics[width=6.8in]{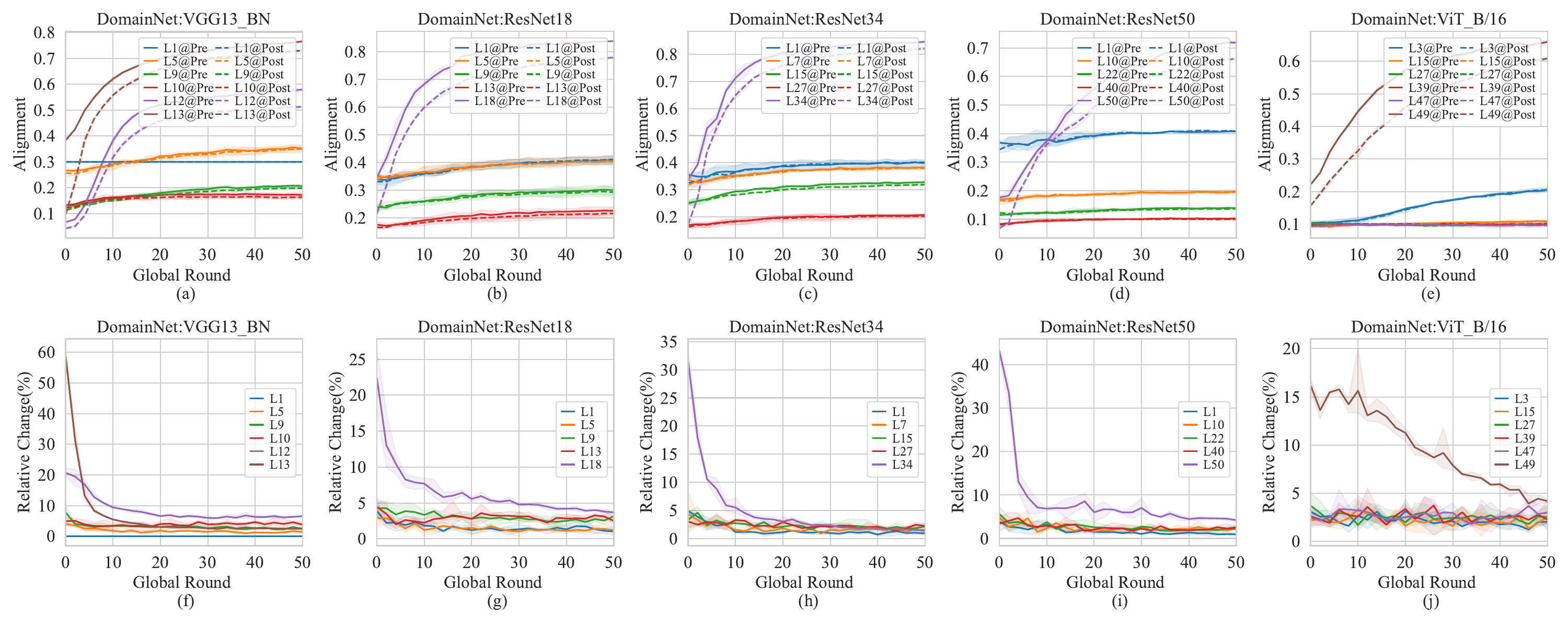}
\caption{Changes in the alignment between features and parameters across FL training at specific model layers. The model is trained on DomainNet with multiple models that are randomly initialized. The top half of the figure shows the original alignment values between features and parameters, while the bottom half displays the relative change in alignment before and after model aggregation.}
\label{FedAvg_AllModel_DomainNet_Layerwise_NC3_NC3}
\end{figure}

\section{Visualization of Pre-aggregated and Post-aggregated Features}

This section visualizes the comparison between pre-aggregated and post-aggregated features.
It can be observed that, as layer depth increases, features become more compressed within the same class and more discriminative across different classes. However, after model aggregation, features become more scattered within the same class and less discriminative across classes. This phenomenon is particularly pronounced in the penultimate layer features (the leftmost subfigure). These observations are consistent with the findings from the quantitative metrics described earlier.

\begin{figure}[H]
\centering
\hspace*{-1.8cm}
\includegraphics[width=6.8in]{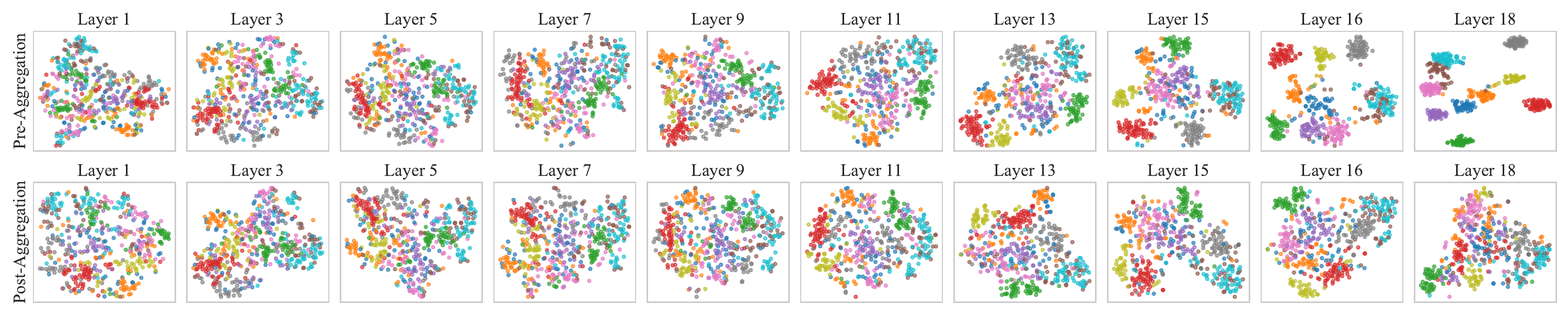}
\caption{T-SNE visualization of features at different layers on the `Quickdraw' domain of DomainNet before and after aggregation.
The features are extracted from ResNet18 in the final global round of FL training, whose parameters are randomly initialized at the beginning.}
\label{TSNE_DomainNet_quickdraw_ResNet18_10Layers}
\end{figure}

\begin{figure}[H]
\centering
\hspace*{-1.8cm}
\includegraphics[width=6.8in]{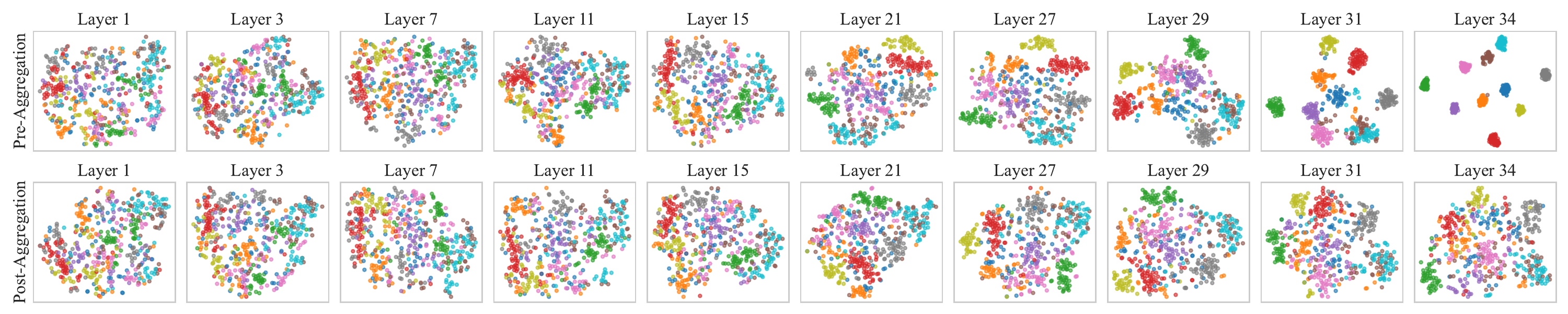}
\caption{T-SNE visualization of features at different layers on the `Quickdraw' domain of DomainNet before and after aggregation.
The features are extracted from ResNet34 in the final global round of FL training, whose parameters are randomly initialized at the beginning.}
\label{TSNE_DomainNet_quickdraw_ResNet34_10Layers}
\end{figure}

\begin{figure}[H]
\centering
\hspace*{-1.8cm}
\includegraphics[width=6.8in]{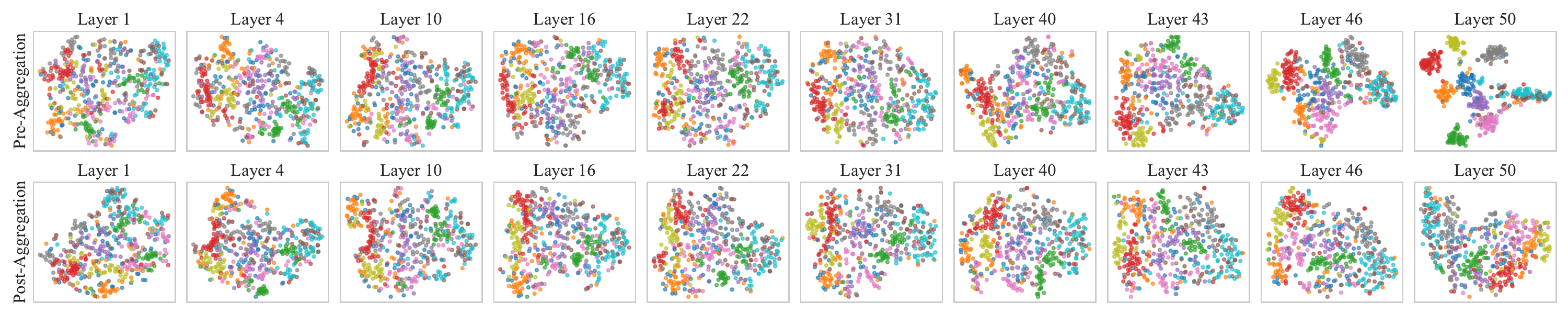}
\caption{T-SNE visualization of features at different layers on the `Quickdraw' domain of DomainNet before and after aggregation.
The features are extracted from ResNet50 in the final global round of FL training, whose parameters are randomly initialized at the beginning.}
\label{TSNE_DomainNet_quickdraw_ResNet50_10Layers}
\end{figure}

\section{Feature Distance and Parameter Distance}

This section compares the distance between features and parameters in the pre-aggregated and post-aggregated models at each layer. It can be observed that, with the exception of the final classifier, the parameter distance between the pre-aggregated and post-aggregated models is significantly smaller than the feature distance.
Furthermore, the distances between parameters and features show different trends across layers. While the parameter distance decreases, the feature distance increases as the layer depth increases.
This demonstrates that even small variations in the parameter space can lead to significant feature variations, as the stacked structure of DNNs tends to magnify errors in feature extraction.
This observation suggests that the `client drift' proposed by previous studies may not be the root cause of performance drops during model aggregation.

\begin{figure}[H]
\centering
\hspace*{-0.8cm}
\includegraphics[width=6.0in]{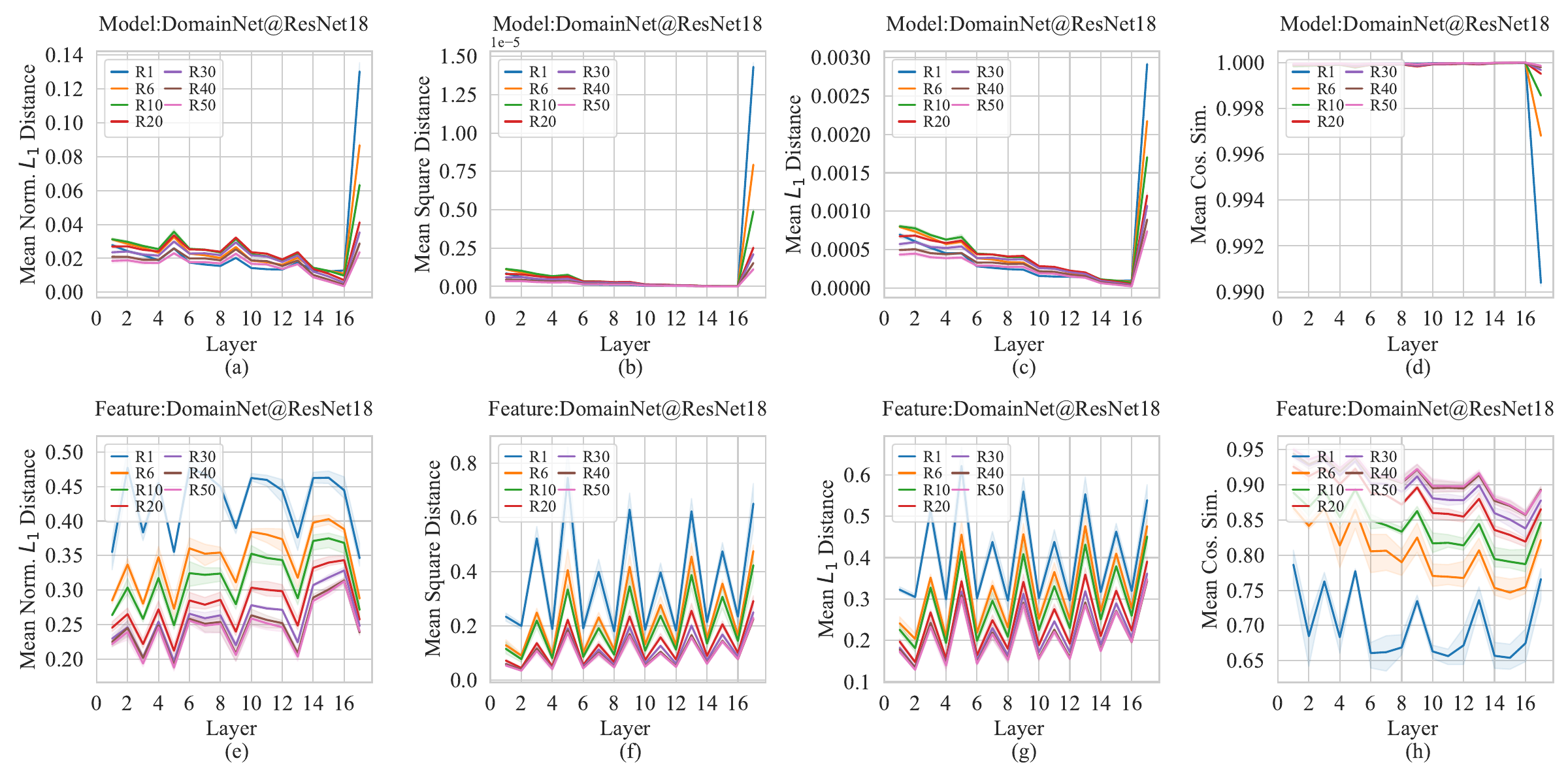}
\caption{Changes in distance of the features and parameters obtained from models before and after aggregation across model layers for specific global rounds, with larger X-axis values indicating deeper layers. The model is trained on DomainNet using ResNet18. The distance is measured by 4 metric including mean normalized distance, mean squared distance, mean $L_1$ distance, and mean cosine similarity. The top half of the figure shows the distance between the parameters of pre-aggregated and post-aggregated models, while the bottom half displays the distance between the features of pre-aggregated and post-aggregated models.
}
\label{FedAvg_Distance_DomainNet_ResNet18_norm_abs_dis_mean_L2_dis_mean_L1_dis_cos_similarity_Epochwise}
\end{figure}

\begin{figure}[H]
\centering
\hspace*{-0.8cm}
\includegraphics[width=6.0in]{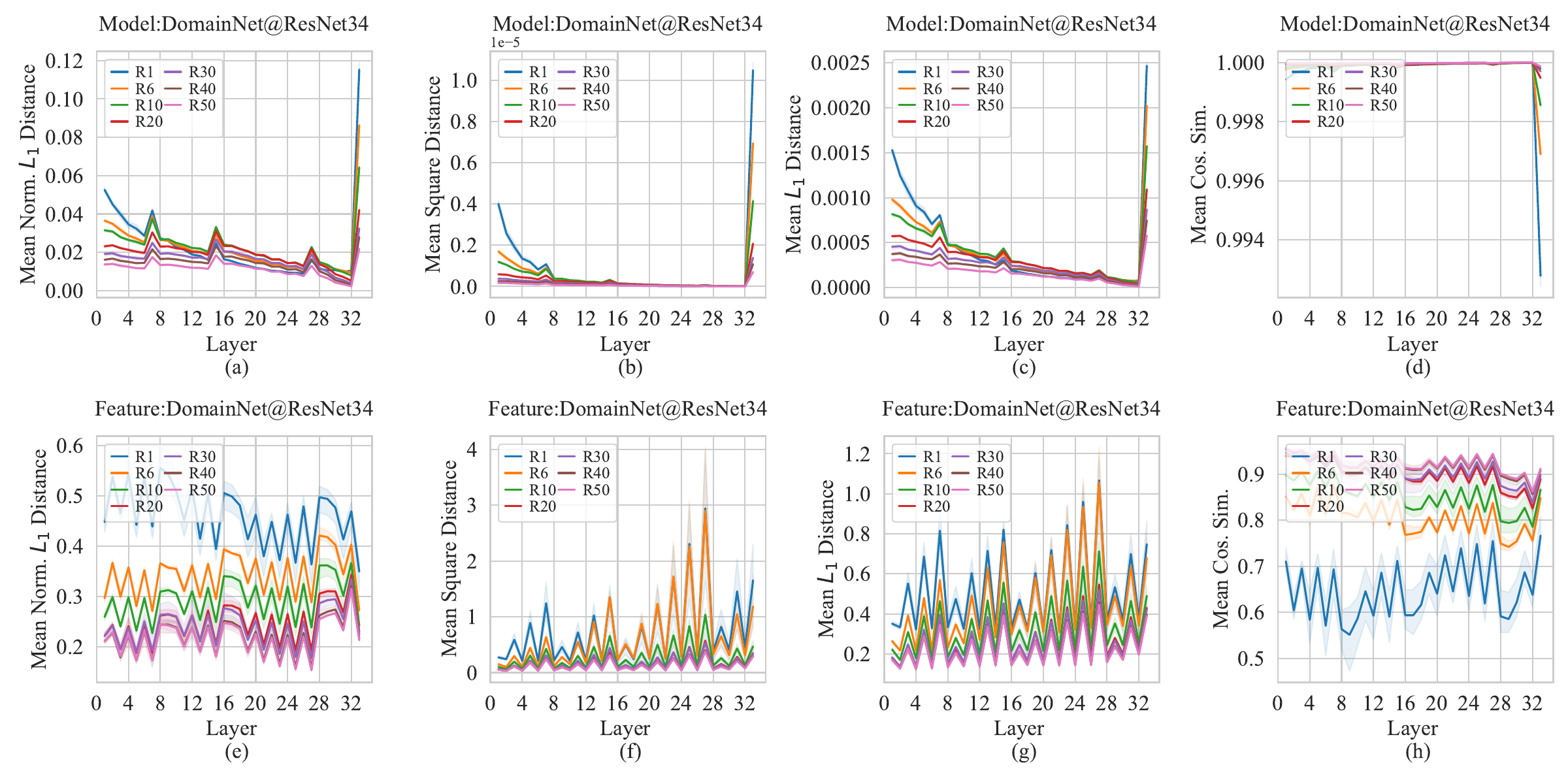}
\caption{Changes in distance of the features and parameters obtained from models before and after aggregation across model layers for specific global rounds, with larger X-axis values indicating deeper layers. The model is trained on DomainNet using ResNet18. The distance is measured by 4 metric including mean normalized distance, mean squared distance, mean $L_1$ distance, and mean cosine similarity. The top half of the figure shows the distance between the parameters of pre-aggregated and post-aggregated models, while the bottom half displays the distance between the features of pre-aggregated and post-aggregated models.
}
\label{FedAvg_Distance_DomainNet_ResNet34_norm_abs_dis_mean_L2_dis_mean_L1_dis_cos_similarity_Epochwise}
\end{figure}

\begin{figure}[H]
\centering
\hspace*{-0.8cm}
\includegraphics[width=6.0in]{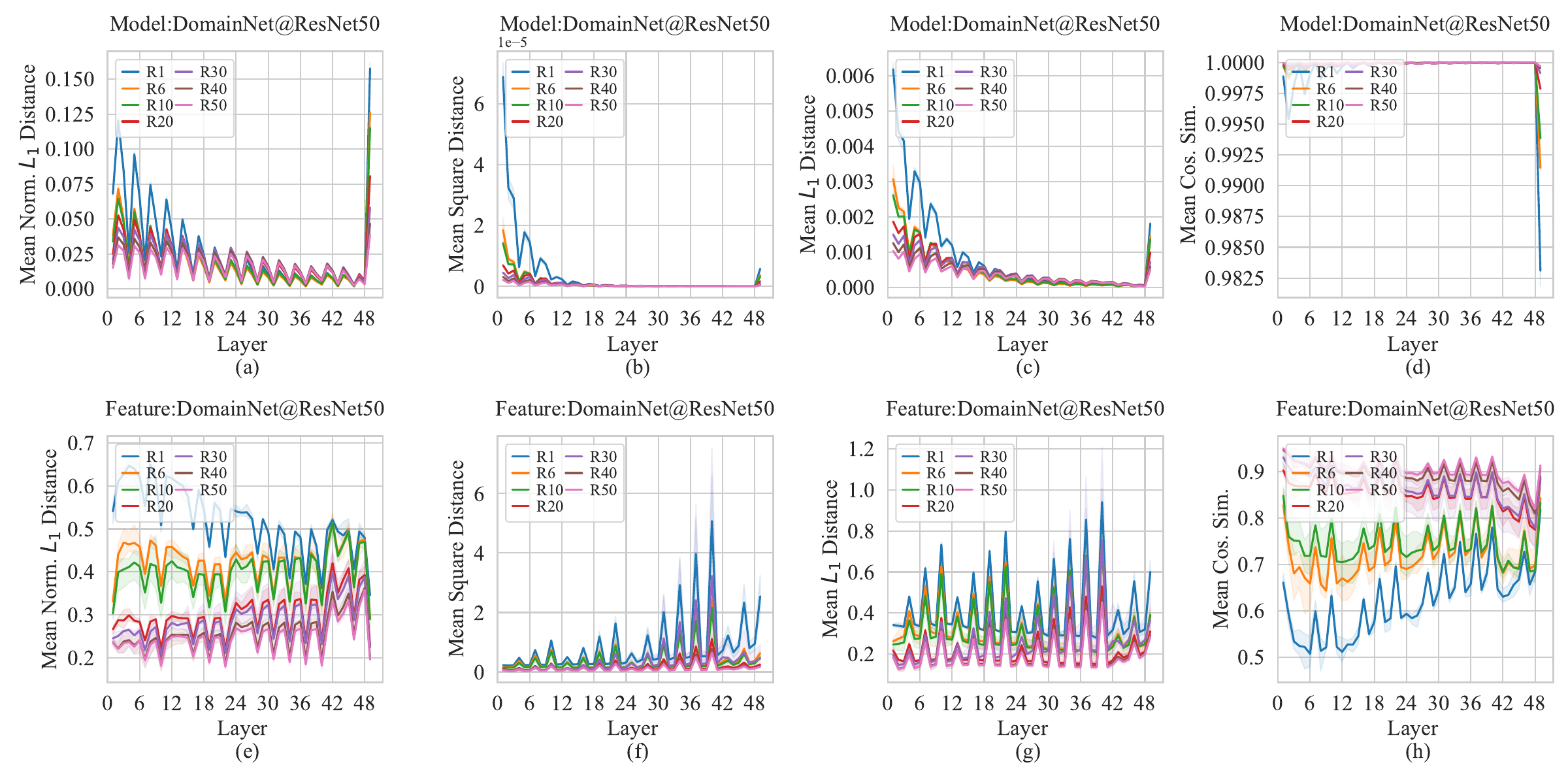}
\vspace{-0.7cm}
\caption{Changes in distance of the features and parameters obtained from models before and after aggregation across model layers for specific global rounds, with larger X-axis values indicating deeper layers. The model is trained on DomainNet using ResNet18. The distance is measured by 4 metric including mean normalized distance, mean squared distance, mean $L_1$ distance, and mean cosine similarity. The top half of the figure shows the distance between the parameters of pre-aggregated and post-aggregated models, while the bottom half displays the distance between the features of pre-aggregated and post-aggregated models.
}
\label{FedAvg_Distance_DomainNet_ResNet50_norm_abs_dis_mean_L2_dis_mean_L1_dis_cos_similarity_Epochwise}
\end{figure}

\vspace{-1.0cm}
\section{Detailed Results of Linear Probing} \label{appendix_linear_probing}
\vspace{-0.5cm}
This section evaluates the accuracy of linear probing across diverse data distributions. In our experiments, we use SGD with a learning rate of 0.01 to optimize the randomly initialized linear layer. The batch size is set to 64, and the total number of epochs is set to 100. We report the best test accuracy as the accuracy of linear probing.

From the experimental results, we observe that the features generated by the post-aggregated model perform well across diverse data distributions. In contrast, the pre-aggregated model only performs well on its local data distribution. This suggests that, while the model after aggregation may not extract task-specific features for local distributions, it is more capable of extracting generalized features that can be applied to different distributions. This is also how model aggregation improves performance compared to purely local training.

\begin{figure}[H]
\centering
\includegraphics[width=3.5in]{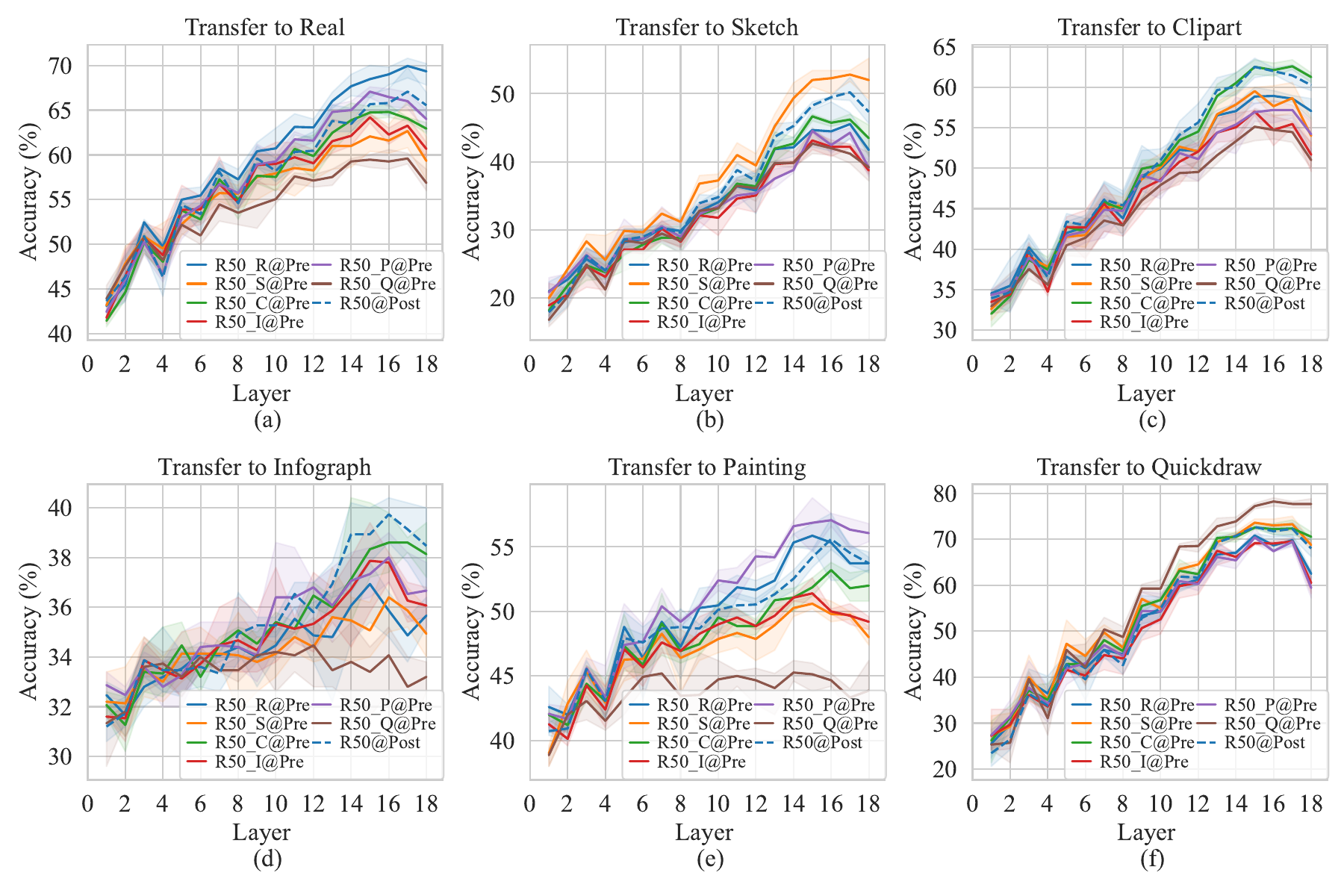}
\vspace{-0.5cm}
\caption{Accuracy of linear probing at different layers across different domains in DomainNet. The experiments are performed in global round 50 using ResNet18 as the backbone model. In the figures, R50@Post represents the results using the model after aggregation, while R50\_*@Pre represents the results using the model before aggregation for domain *.}
\label{Linear_Probing_DomainNet_E50_ResNet18_Long_Epochwise}
\end{figure}

\begin{figure}[H]
\centering
\includegraphics[width=4.0in]{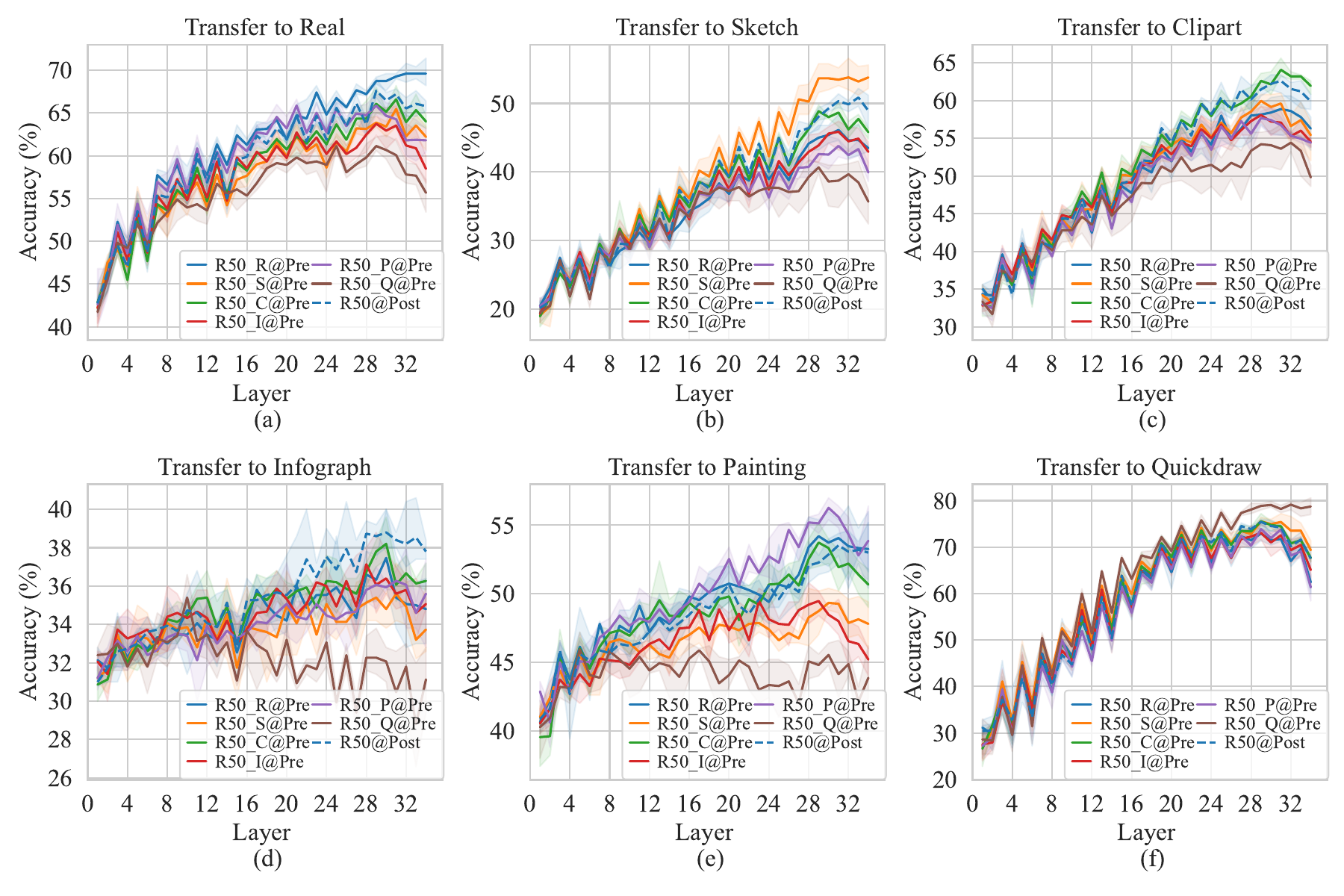}
\caption{Accuracy of linear probing at different layers across different domains in DomainNet. The experiments are performed in global round 50 using ResNet34 as the backbone model. In the figures, R50@Post represents the results using the model after aggregation, while R50\_*@Pre represents the results using the model before aggregation for domain *.}
\label{Linear_Probing_DomainNet_E50_ResNet34_Long_Epochwise}
\end{figure}

\section{Effect of Residual Connection}
Motivated by the observed zigzag pattern linked to residual blocks, this section investigates the role of residual connections in model aggregation. We use ResNet18 and ResNet34 as backbones and remove the residual connections from these models. We then employ these modified architectures for FL training. This section presents a comparison of accuracy and feature metrics between the models with and without residual connections. We observe that the relative change in feature metrics is more pronounced when the residual connections are removed, particularly during the early stages of model training and in the normalized within-class and between-class feature variance metrics. This may be because residual connections help mitigate feature degradation by allowing less disrupted features in lower layers to be propagated to deeper layers.

\begin{figure}[H]
\centering
\includegraphics[width=5.1in]{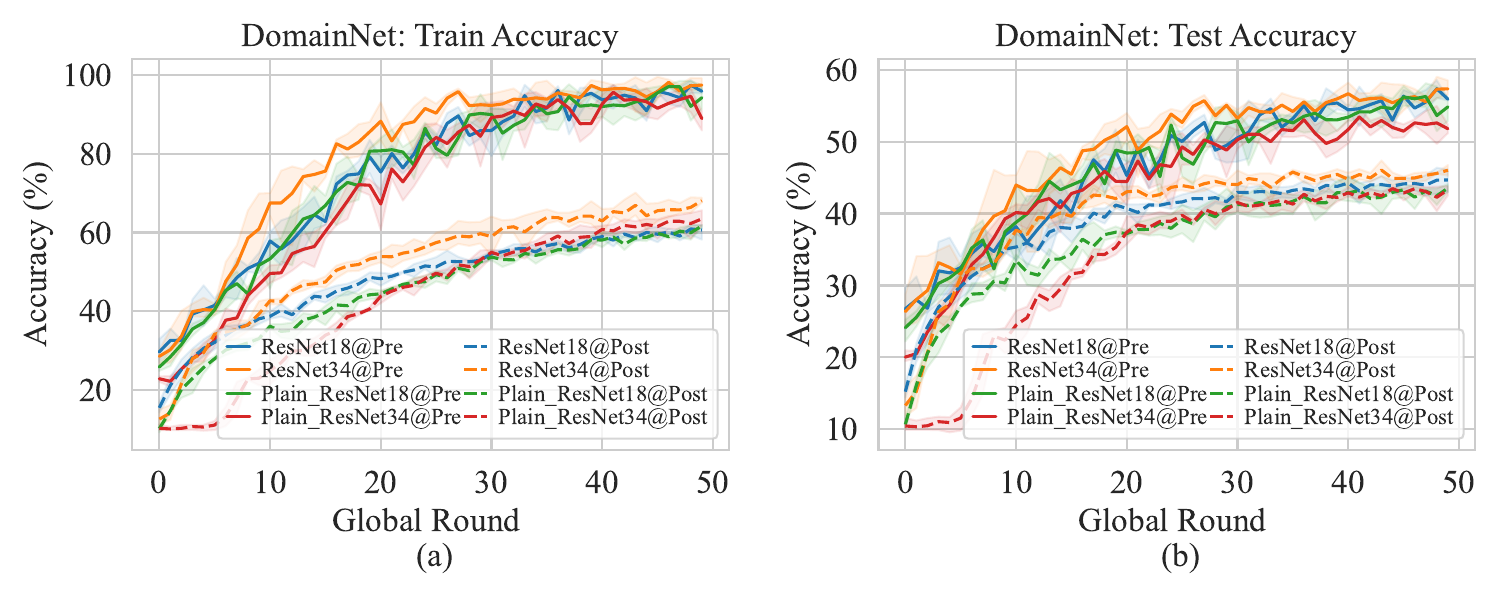}
\caption{Averaged training and testing accuracy curves of the model before and after aggregation, evaluated on the local dataset during FL training. The experiments are conducted on the DomainNet dataset. The models trained include the original ResNet18 and ResNet, as well as their plain versions, with residual connections removed.}
\label{Accuracy_Residual_Connection}
\end{figure}

\section{Detailed Results when Personalizing Specific Parameters}\label{appendix_parameter_personalization}
\begin{figure}[ht]
\centering
\includegraphics[width=4.5in]{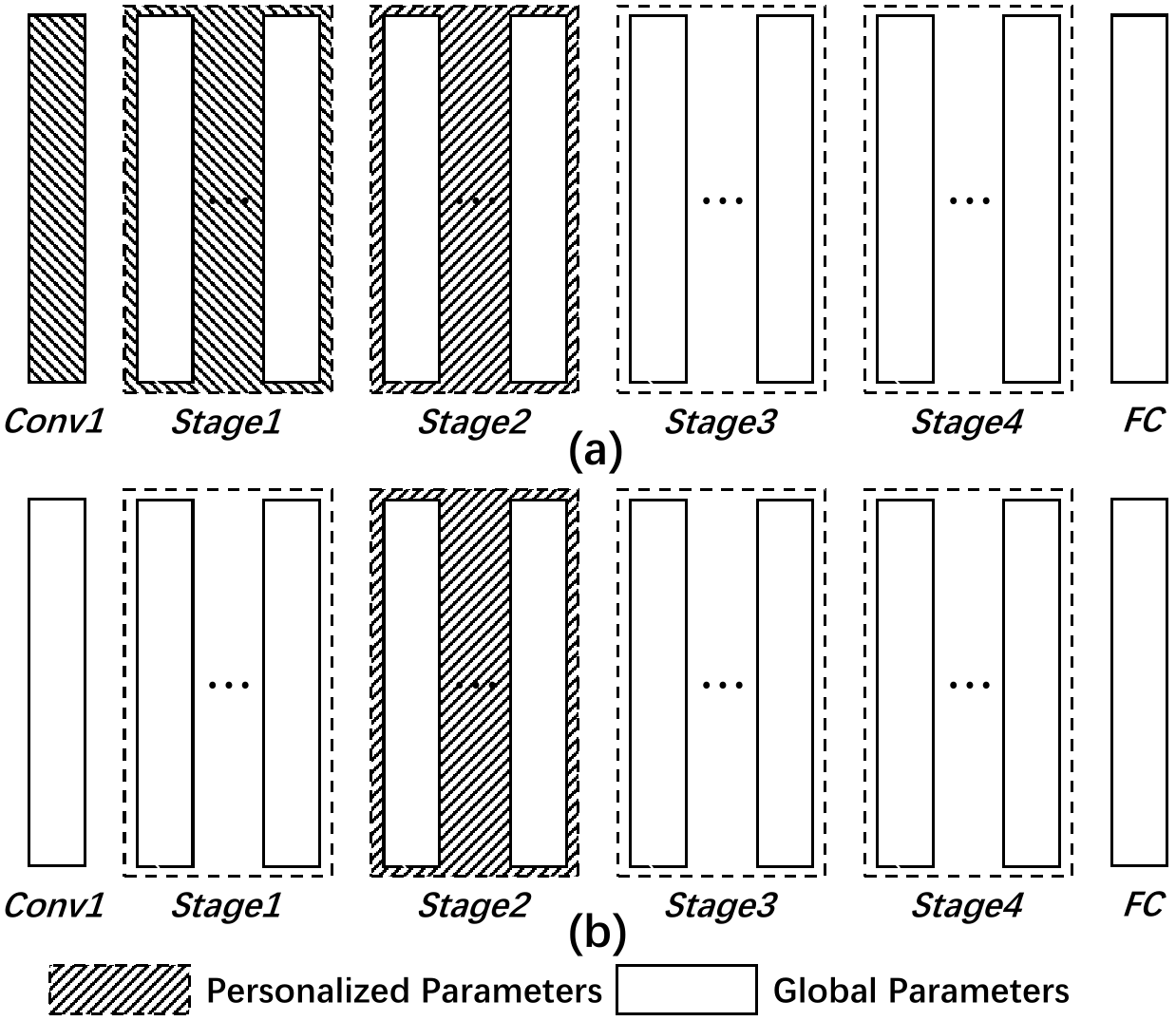}
\caption{Illustration of the architectures of ResNet18 and ResNet34, along with the parameter personalization strategies: (a) Successive parameter personalization strategy, (b) Skip parameter personalization strategy.}
\label{PFL}
\end{figure}

This section presents the detailed experimental results obtained by personalizing specific layers within the model during FL training, a method known as personalized federated learning (PFL).
The experiments are conducted using the ResNet18 and ResNet34 models.
As shown in Figure \ref{PFL}, both models can be divided into several blocks: the first convolutional layer, stages 1-4 (which consist of stacked convolutional layers), and the fully connected (FC) layer used for classification.

In the experiments, we explore PFL methods by personalizing specific layers within the model.
Specifically, we first examine two commonly used PFL methods: FedPer \citep{FedPer}, which personalizes the classifier, and FedBN \citep{FedBN}, which personalizes the batch normalization (BN) layers.
Additionally, motivated by \citep{PartialFed}, we consider two more strategies for parameter personalization: the successive parameter personalization strategy and the skip parameter personalization strategy.
For the successive parameter personalization strategy, we personalize multiple consecutive layers starting from the first layer.
In the skip parameter personalization strategy, we randomly select a layer or block for personalization.

The experimental results show that these parameter personalization methods generally lead to better featured distributions on local data. Moreover, the relative changes introduced by model aggregation decrease as more parameters within the feature extractor are personalized. This improvement is due to the reduction in feature disruption accumulation, which is alleviated by personalizing these parameters, thereby maintaining their ability to extract locally task-specific features without being affected by model aggregation. 

\subsection{Successive Parameter Personalization}

\begin{figure}[H]
\centering
\hspace*{-1.8cm}
\includegraphics[width=6.8in]{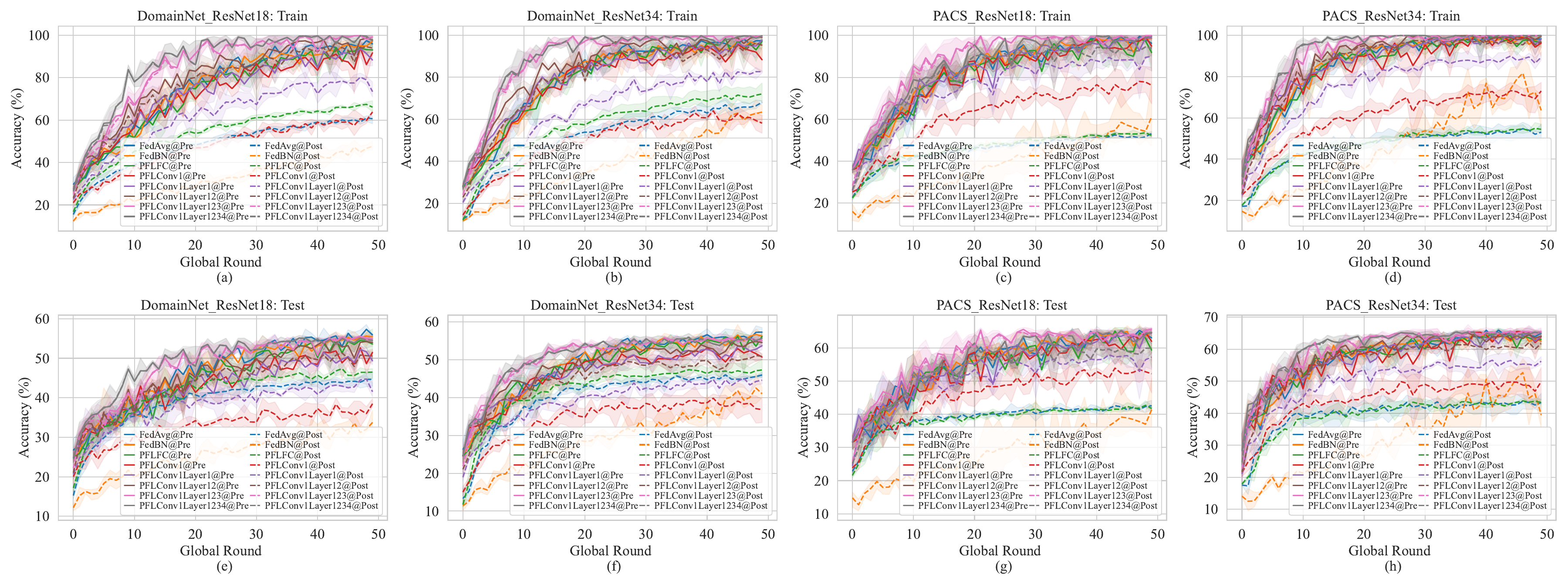}
\caption{Accuracy curves when training FL models with different successive parameter personalization strategies, along with FedAvg, FedPer, and FedBN. In the legend, `C1' is an abbreviation for the Conv1 layer, and `S*' represents the abbreviation for Stage* block.}
\label{PFL_successive_Accuracy}
\end{figure}

\subsection{Skip Parameter Personalization}
\begin{figure}[H]
\centering
\hspace*{-1.8cm}
\includegraphics[width=6.8in]{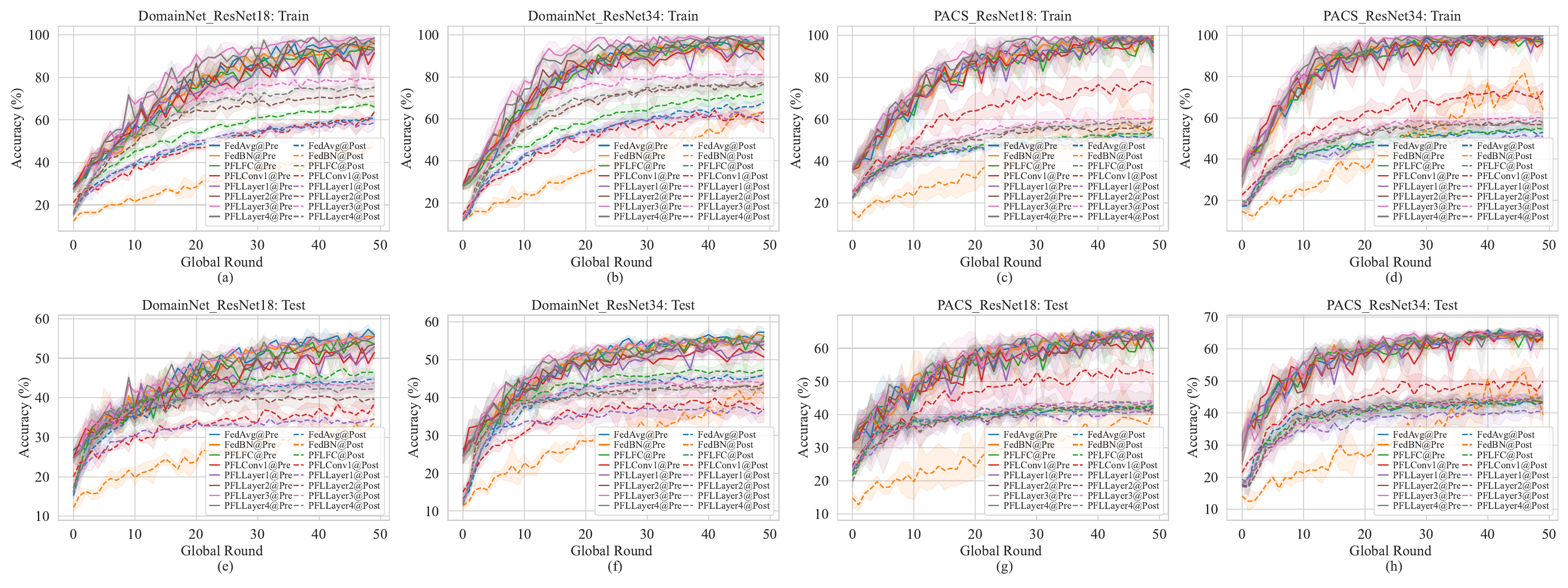}
\caption{Accuracy curves when training FL models with different skip parameter personalization strategies, along with FedAvg, FedPer, and FedBN. In the legend, `C1' is an abbreviation for the Conv1 layer, and `S*' represents the abbreviation for Stage* block.}
\label{PFL_successive_Accuracy}
\end{figure}

\section{Effect of Local Updating Epochs}
\begin{figure}[ht]
\centering
\includegraphics[width=3.2in]{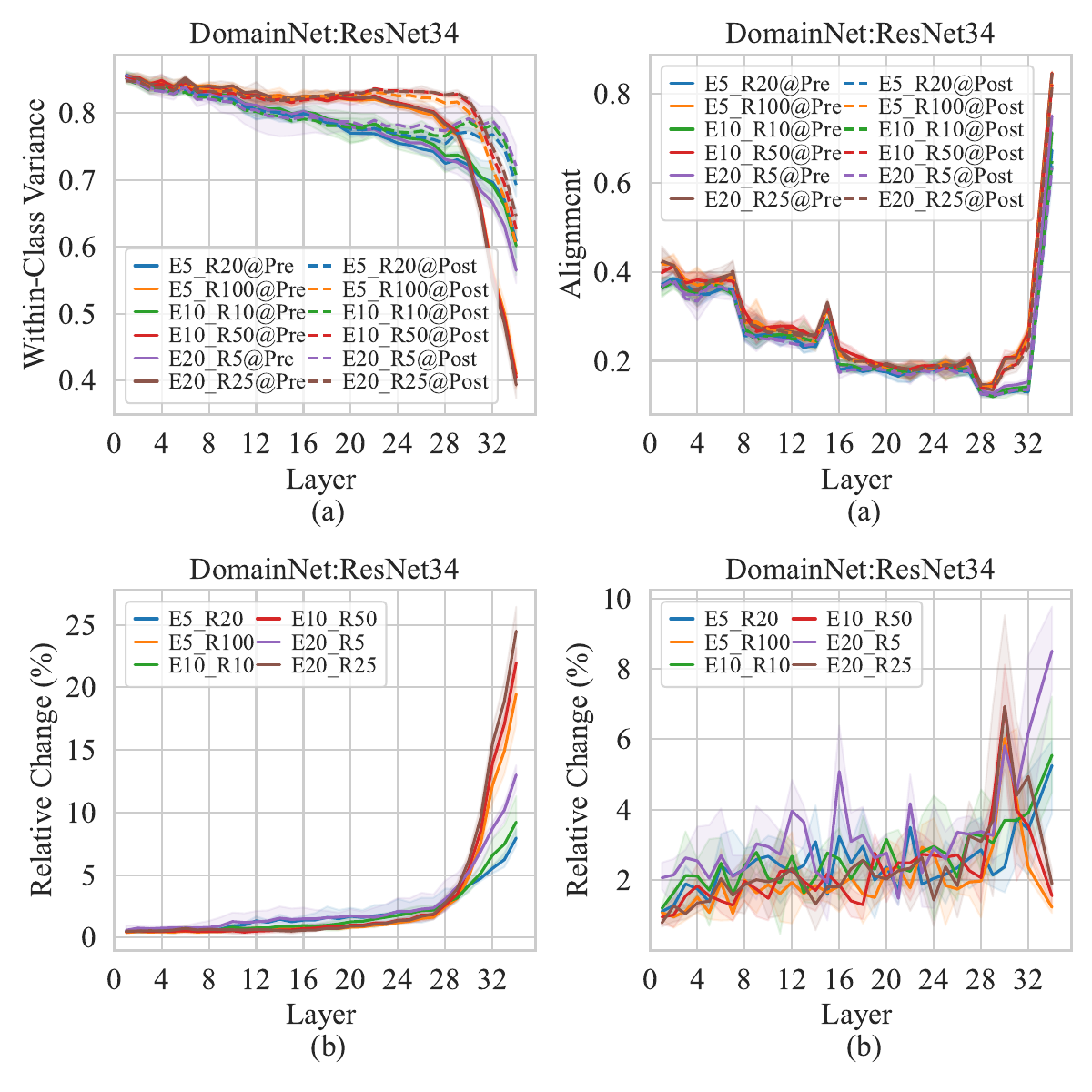}
\vspace{-0.4cm}
\caption{Changes in the normalized within-class feature variance and the alignment between features and parameters when training the FL model with different local update epochs. The model is trained on DomainNet using ResNet34 as backbone. We present two groups of experiments, keeping the total number of updating epochs ($E \times R$) fixed at $100$ and $500$.}
\label{FedAvg_LocalEp_DomainNet_ResNet34_Epochwise_NC1_NC1_NC3_NC3}
\end{figure}

\vspace{-0.45cm}
The local update epoch is a key configuration in federated learning that determines the aggregation frequency.
In this section, we conduct experiments to investigate its impact on model aggregation. 
During experimets, we keep the total number of local updates fixed and vary the number of local epochs.
The experimental results are presented in Figure \ref{FedAvg_LocalEp_DomainNet_ResNet34_Epochwise_NC1_NC1_NC3_NC3}. 
To ensure a fair comparison, we maintain a consistent total number of updating epochs across comparisons. The results show that increasing the number of local epochs will compress the within-class features more effectively.
However, when the models are aggregated, the relative change for larger local update epochs is noticeably greater than for smaller ones. This highlights the sensitivity of the model aggregation process to the number of local updating epochs.
\vspace{-0.8cm}

\section{Detailed Results when Fine-tuning Classifier}\label{appendix_finetune_classifier}
This section provides detailed experimental results on fine-tuning the classifier to reduce the mismatch between the features extracted by the global feature extractor and the classifier. The experiments are conducted on DomainNet using ResNet18 and ResNet34 as backbones. During fine-tuning, we use SGD with momentum as the optimizer, with a learning rate of 0.01 and momentum set to 0.1. We perform only 10 epochs of fine-tuning.

It can be observed that, after fine-tuning the classifier, the alignment between the features and the classifier consistently improves using models at different global rounds during the FL training. As a result, the testing accuracy also improves.

\begin{figure}[H]
\centering
\includegraphics[width=\textwidth]{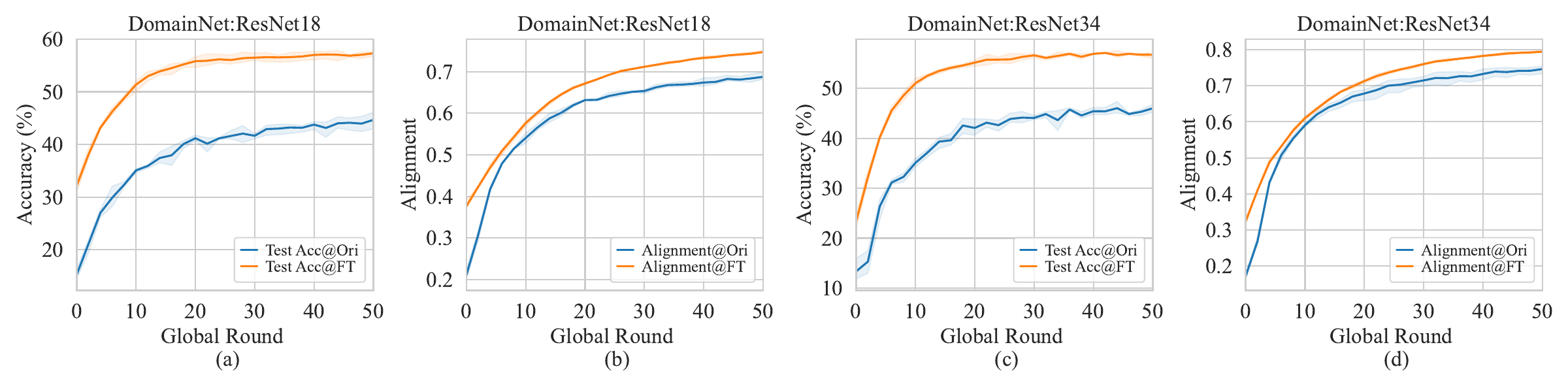}
\vspace{-0.8cm}
\caption{Accuracy and alignment between features and parameters during classifier fine-tuning across global rounds. The experiments are conducted on DomainNet dataset, using ResNet18 and ResNet34 as backbone model.}
\label{FedAvg_ResNet18_ResNet34_DomainNet_Finetune_Classifier}
\end{figure}

\newpage
\section{Detailed Results when Initializing Model with Pre-trained Parameters}\label{appendix_pretrained}
\vspace{-0.8cm}
In this section, we present detailed experimental results from using pre-trained parameters as initialization for FL training.
These experiments follow the same setup as the ones described above, with the only difference being that the model is initialized with parameters pre-trained on large-scale datasets.
The experimental results include the changes in feature variance across layers and training rounds, the alignment between features and parameters across layers and rounds, and the visualization of pre-aggregated and post-aggregated features.
From these experiments, we find that initializing with pre-trained parameters effectively mitigates the accumulation of feature degradation.
This conclusion is based on observations from experiments with randomly initialized parameters, where deeper layers only begin to converge once shallow features have reached a specific state.
This is primarily because the degradation of unconverged features in the shallow layers propagates to the deeper layers, preventing them from converging until the shallow layers are well-trained. This significantly hinders the convergence rate of FL training.
However, when initialized with pre-trained parameters, we find that the features of most shallow layers converge early in training, as the model already possesses strong feature extraction capabilities from being trained on a large-scale dataset.
This significantly reduces the feature degradation accumulation introduced by model aggregation, thereby stabilizing and accelerating the convergence of FL training.
\vspace{-1.0cm}
\subsection{Comparison of Training and Testing Accuracy}
\vspace{-1.0cm}
\begin{figure}[!htbp]
\centering
\hspace*{-1.8cm}
\includegraphics[width=6.8in]{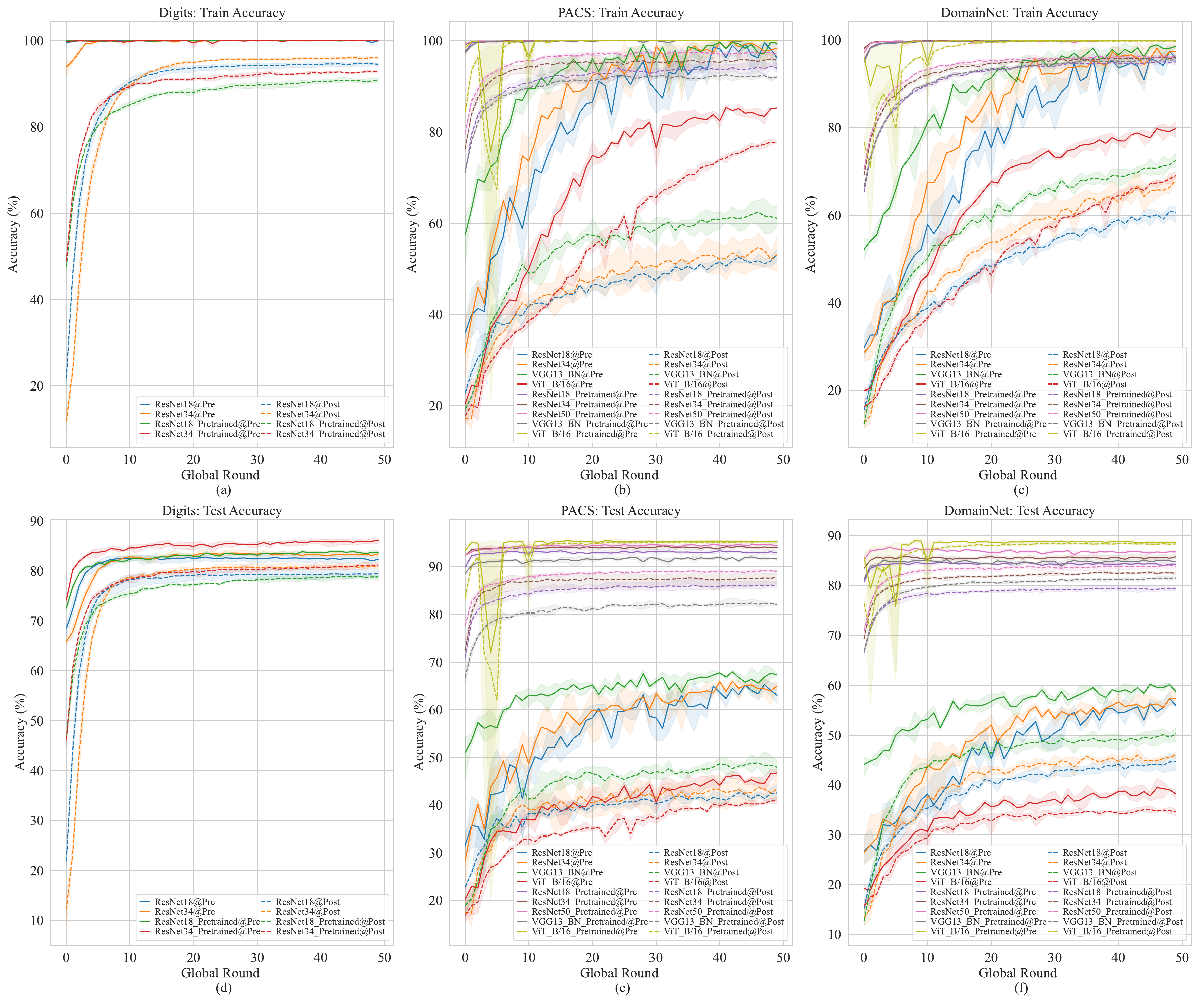}
\vspace{-0.8cm}
\caption{Comparison of accuracy with and without pre-trained parameters as initialization. The performance drop can be significantly mitigated by using pre-trained parameters.}

\label{FedAvg_AllModel_Pretrained_Accuracy}
\end{figure}

\subsection{Changes of Feature Variance Across Layers}

\begin{figure}[H]
\centering
\includegraphics[width=3.0in]{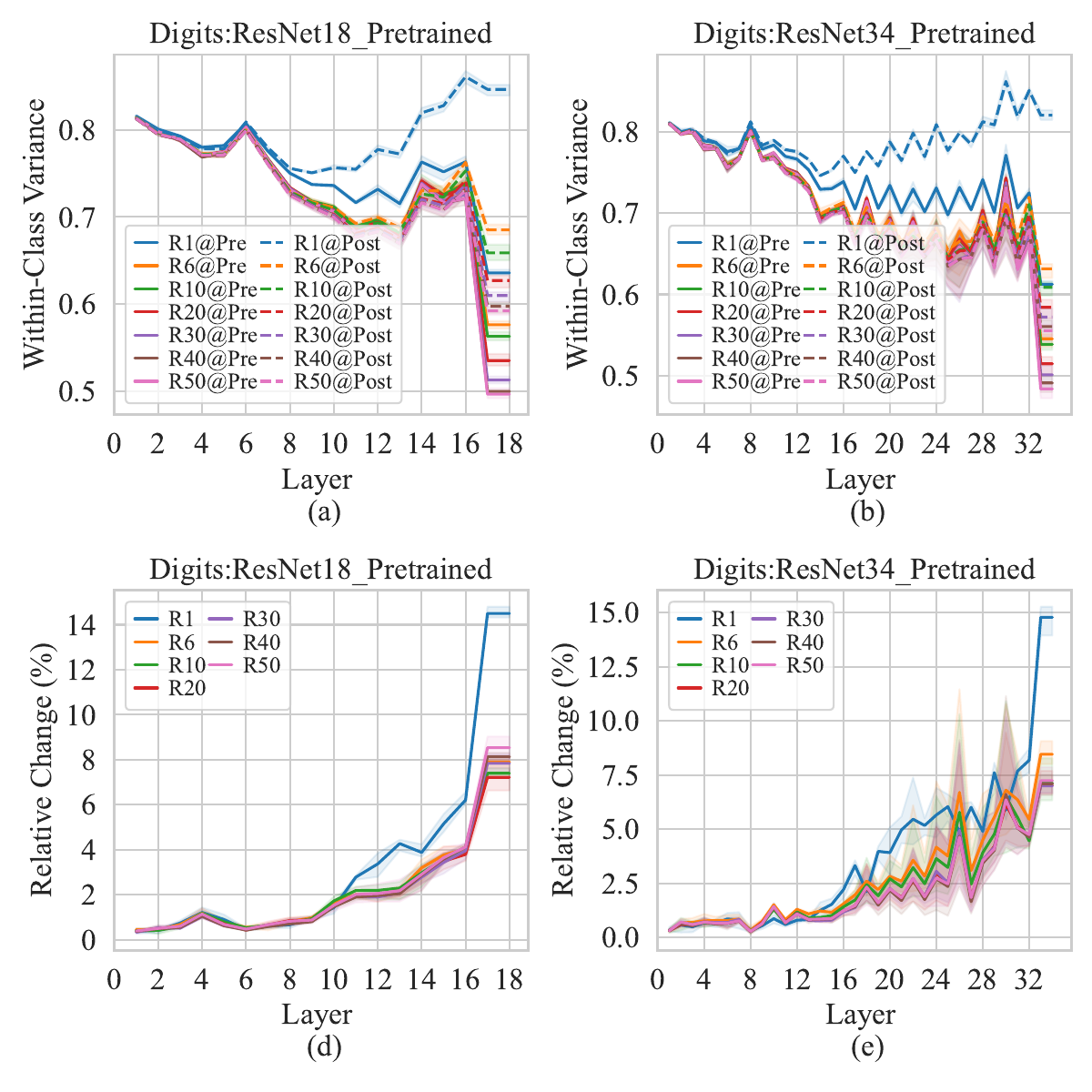}
\caption{Changes in the normalized within-class variance of features across model layers for specific global rounds, with larger X-axis values indicating deeper layers. The model is trained on Digit-Five with multiple models that are initialized by parameters pre-trained on large-scaled datasets. The top half of the figure shows the normalized within-class variance, while the bottom half displays the relative change in variance before and after model aggregation.}
\label{FedAvg_AllModel_Pretrained_Digits_Epochwise_NC1_NC1}
\end{figure}

\begin{figure}[H]
\centering
\includegraphics[width=3.0in]{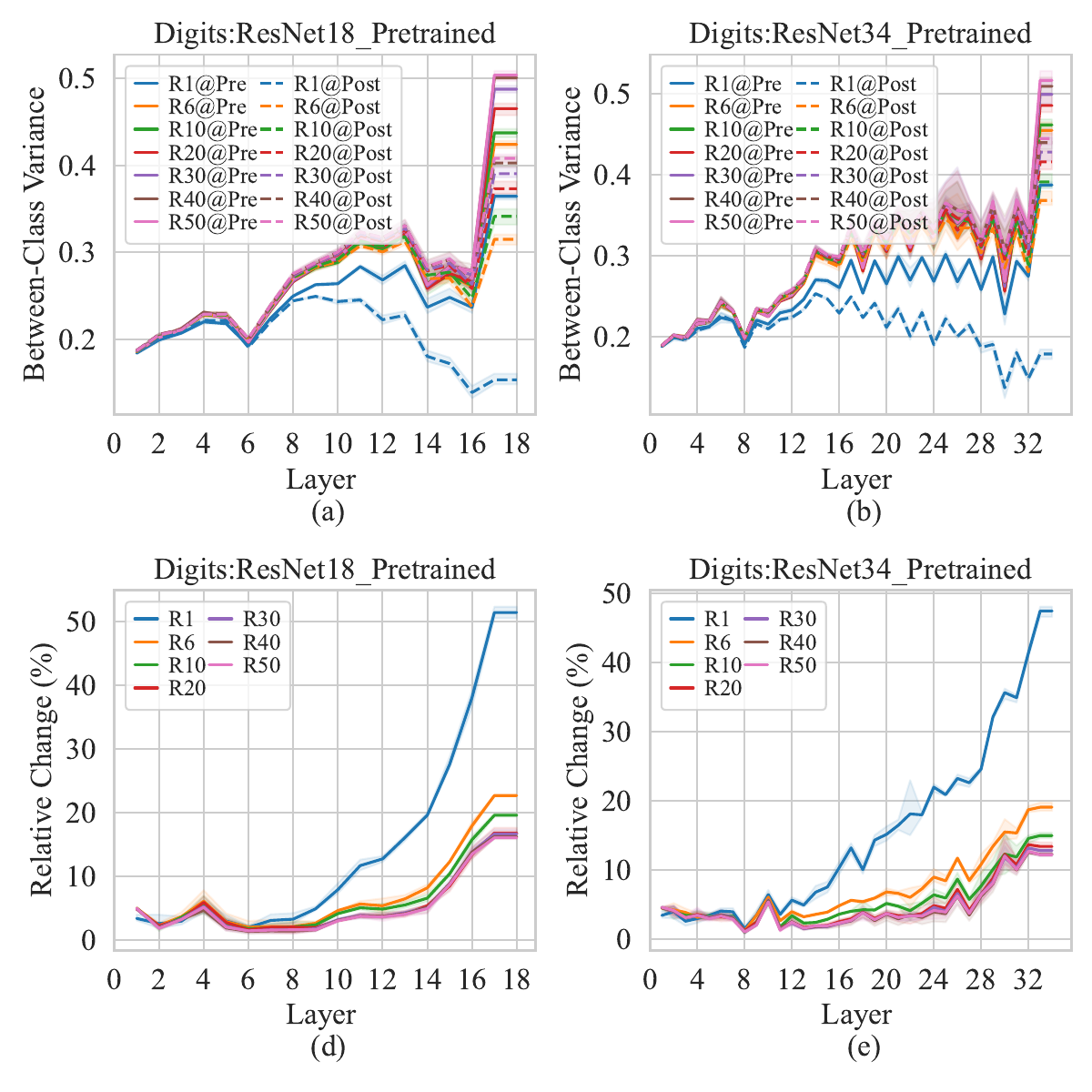}
\caption{Changes in the normalized between-class variance of features across model layers for specific global rounds, with larger X-axis values indicating deeper layers. The model is trained on Digit-Five with multiple models that are initialized by parameters pre-trained on large-scaled datasets. The top half of the figure shows the normalized between-class variance, while the bottom half displays the relative change in variance before and after model aggregation.}
\label{FedAvg_AllModel_Pretrained_Digits_Epochwise_NC1_between}
\end{figure}

\begin{figure}[H]
\centering
\includegraphics[width=3.0in]{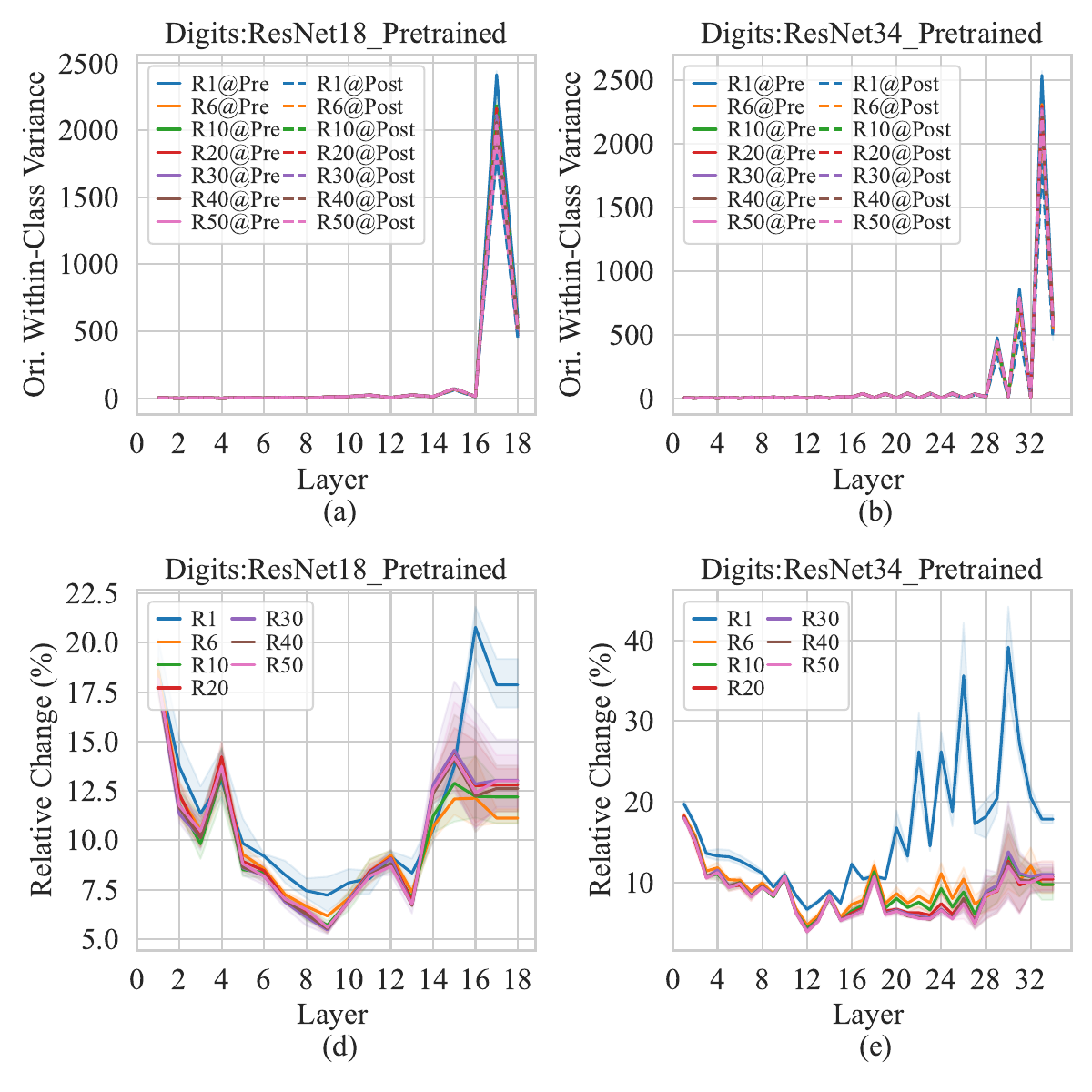}
\caption{Changes in the original unnormalized within-class variance of features across model layers for specific global rounds, with larger X-axis values indicating deeper layers. The model is trained on Digit-Five with multiple models that are initialized by parameters pre-trained on large-scaled datasets. The top half of the figure shows the original unnormalized within-class variance, while the bottom half displays the relative change in variance before and after model aggregation.}
\label{FedAvg_AllModel_Pretrained_Digits_Epochwise_NC1_trace_within}
\end{figure}

\begin{figure}[H]
\centering
\includegraphics[width=3.0in]{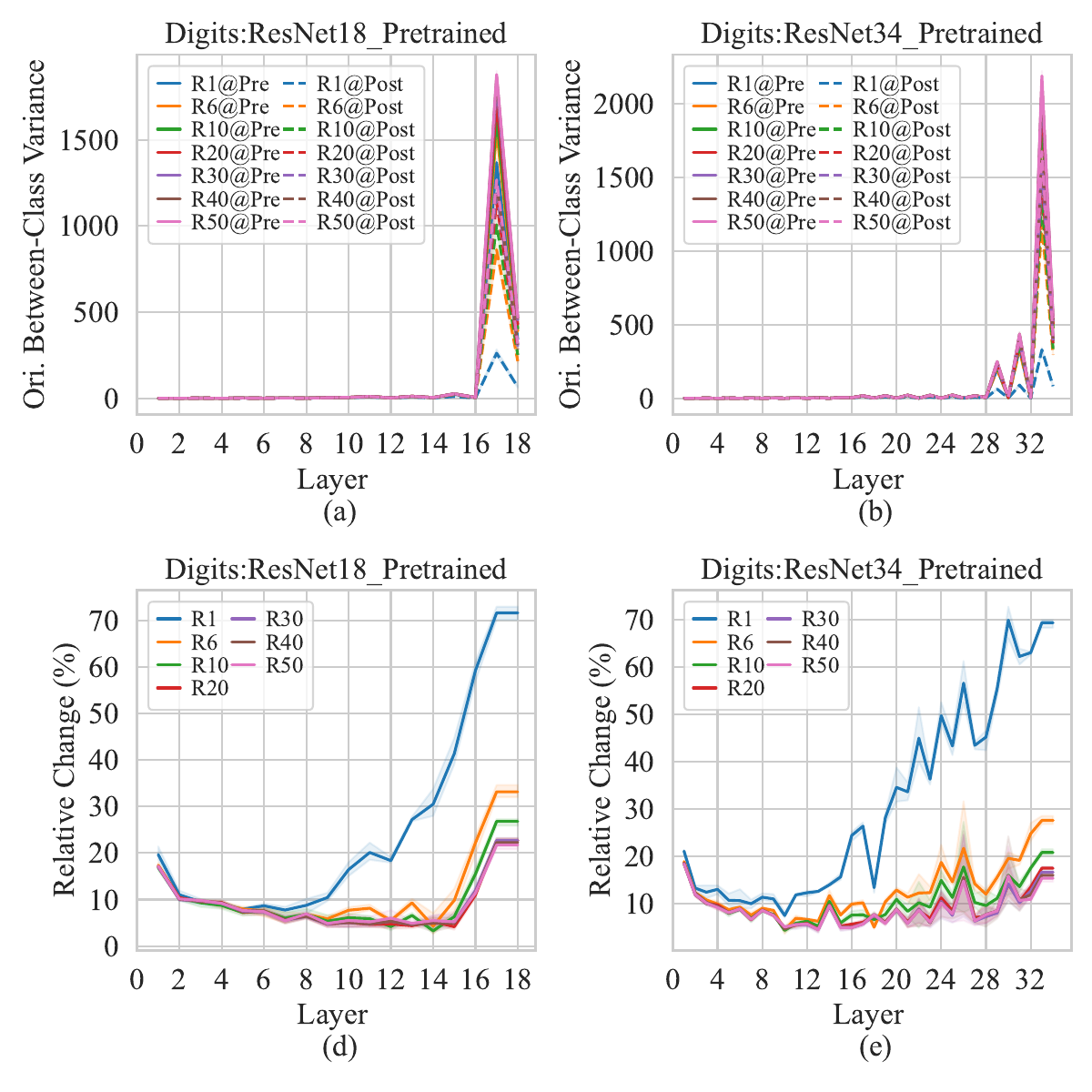}
\caption{Changes in the original unnormalized between-class variance of features across model layers for specific global rounds, with larger X-axis values indicating deeper layers. The model is trained on Digit-Five with multiple models that are initialized by parameters pre-trained on large-scaled datasets. The top half of the figure shows the original unnormalized between-class variance, while the bottom half displays the relative change in variance before and after model aggregation.}
\label{FedAvg_AllModel_Pretrained_Digits_Epochwise_NC1_trace_between}
\end{figure}

\begin{figure}[H]
\centering
\hspace*{-1.8cm}
\includegraphics[width=6.8in]{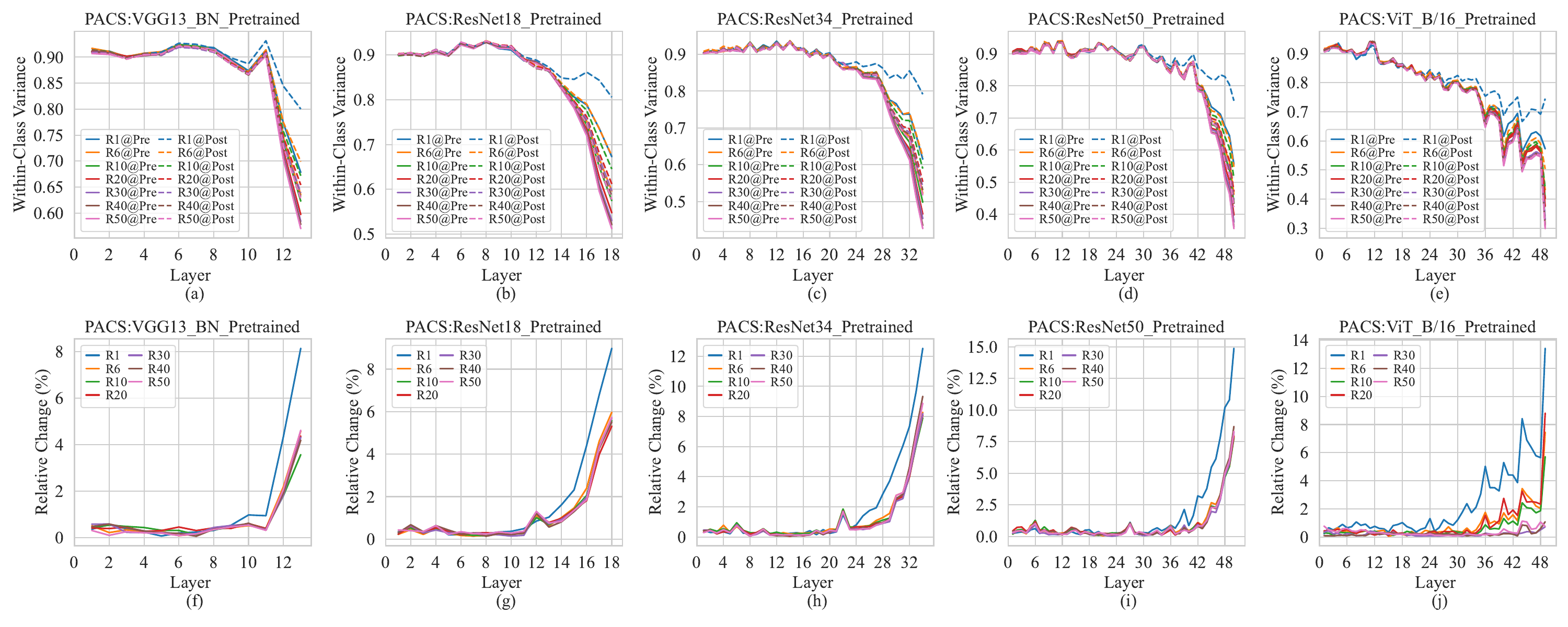}
\caption{Changes in the normalized within-class variance of features across model layers for specific global rounds, with larger X-axis values indicating deeper layers. The model is trained on PACS with multiple models that are initialized by parameters pre-trained on large-scaled datasets. The top half of the figure shows the normalized within-class variance, while the bottom half displays the relative change in variance before and after model aggregation.}
\label{FedAvg_AllModel_Pretrained_PACS_Epochwise_NC1_NC1}
\end{figure}

\begin{figure}[H]
\centering
\hspace*{-1.8cm}
\includegraphics[width=6.8in]{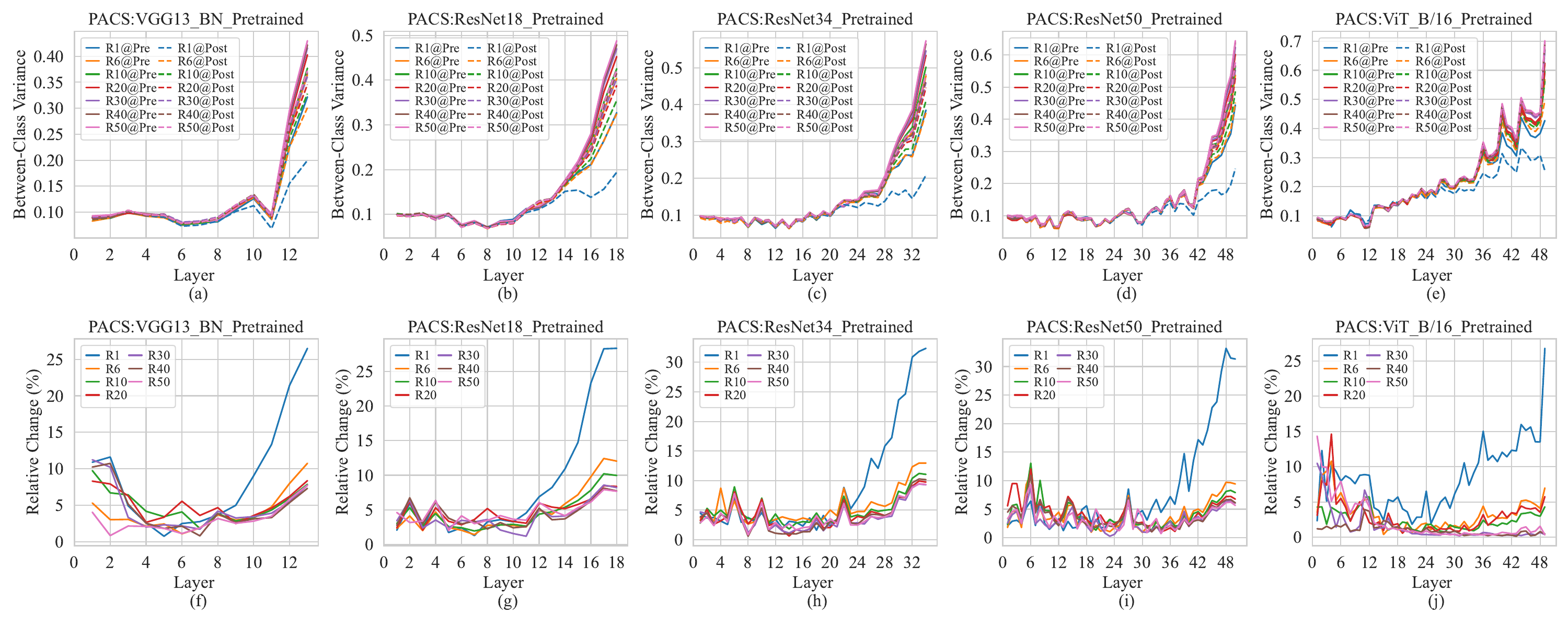}
\caption{Changes in the normalized between-class variance of features across model layers for specific global rounds, with larger X-axis values indicating deeper layers. The model is trained on PACS with multiple models that are initialized by parameters pre-trained on large-scaled datasets. The top half of the figure shows the normalized between-class variance, while the bottom half displays the relative change in variance before and after model aggregation.}
\label{FedAvg_AllModel_Pretrained_PACS_Epochwise_NC1_between}
\end{figure}

\begin{figure}[H]
\centering
\hspace*{-1.8cm}
\includegraphics[width=6.8in]{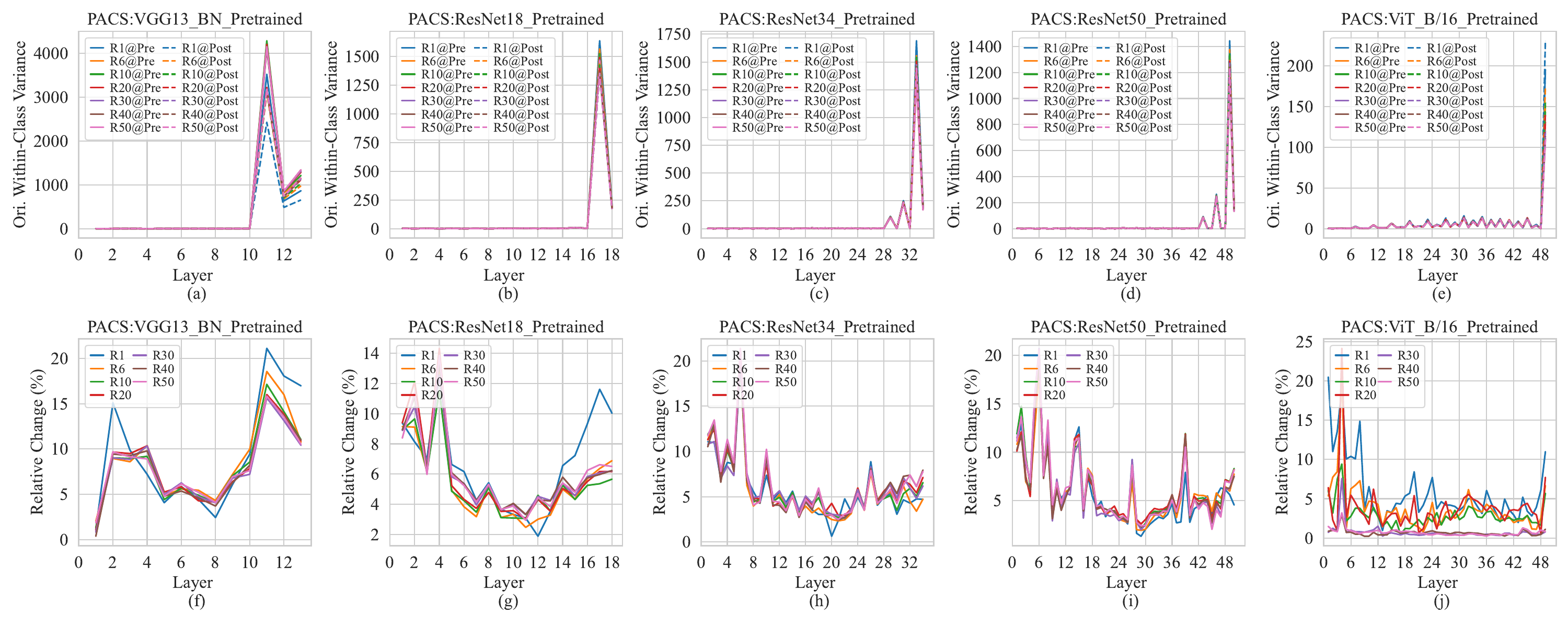}
\caption{Changes in the original unnormalized within-class variance of features across model layers for specific global rounds, with larger X-axis values indicating deeper layers. The model is trained on PACS with multiple models that are initialized by parameters pre-trained on large-scaled datasets. The top half of the figure shows the original unnormalized within-class variance, while the bottom half displays the relative change in variance before and after model aggregation.}
\label{FedAvg_AllModel_Pretrained_PACS_Epochwise_NC1_trace_within}
\end{figure}

\begin{figure}[H]
\centering
\hspace*{-1.8cm}
\includegraphics[width=6.8in]{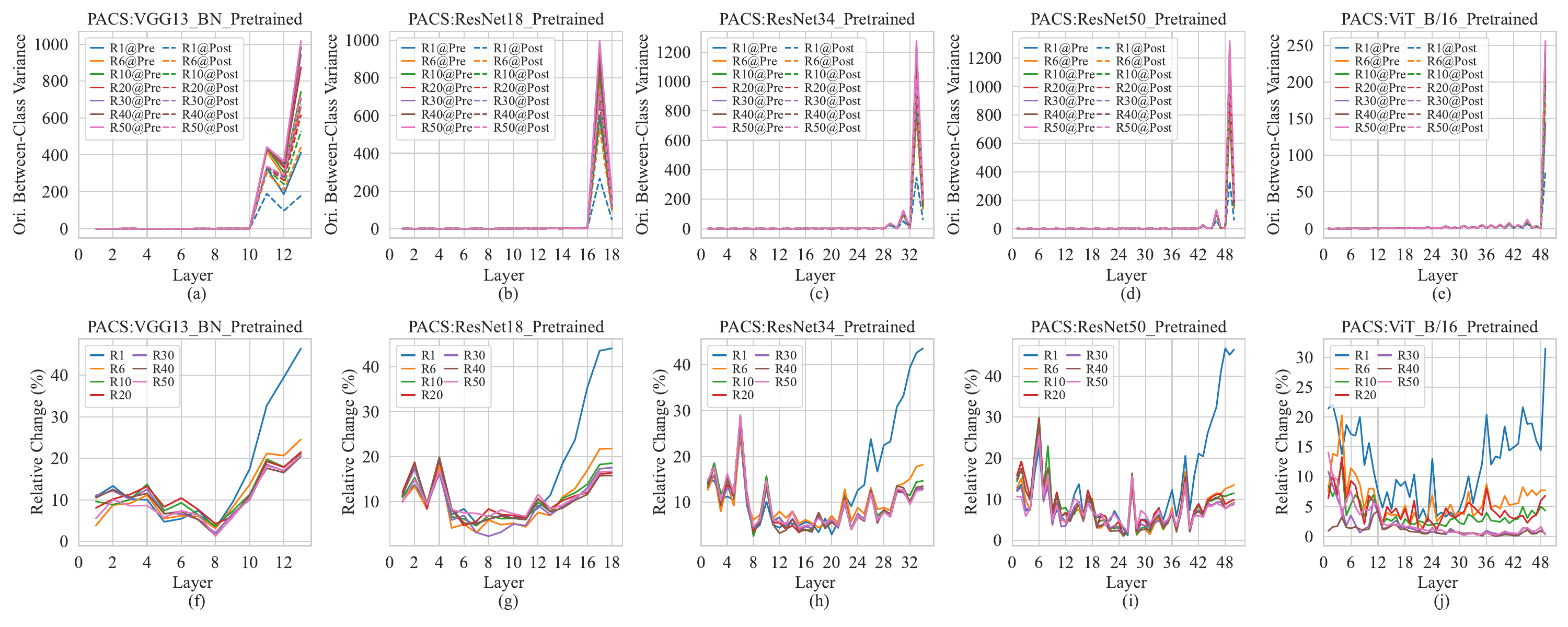}
\caption{Changes in the original unnormalized between-class variance of features across model layers for specific global rounds, with larger X-axis values indicating deeper layers. The model is trained on PACS with multiple models that are initialized by parameters pre-trained on large-scaled datasets. The top half of the figure shows the original unnormalized between-class variance, while the bottom half displays the relative change in variance before and after model aggregation.}
\label{FedAvg_AllModel_Pretrained_PACS_Epochwise_NC1_trace_between}
\end{figure}

\begin{figure}[H]
\centering
\hspace*{-1.8cm}
\includegraphics[width=6.8in]{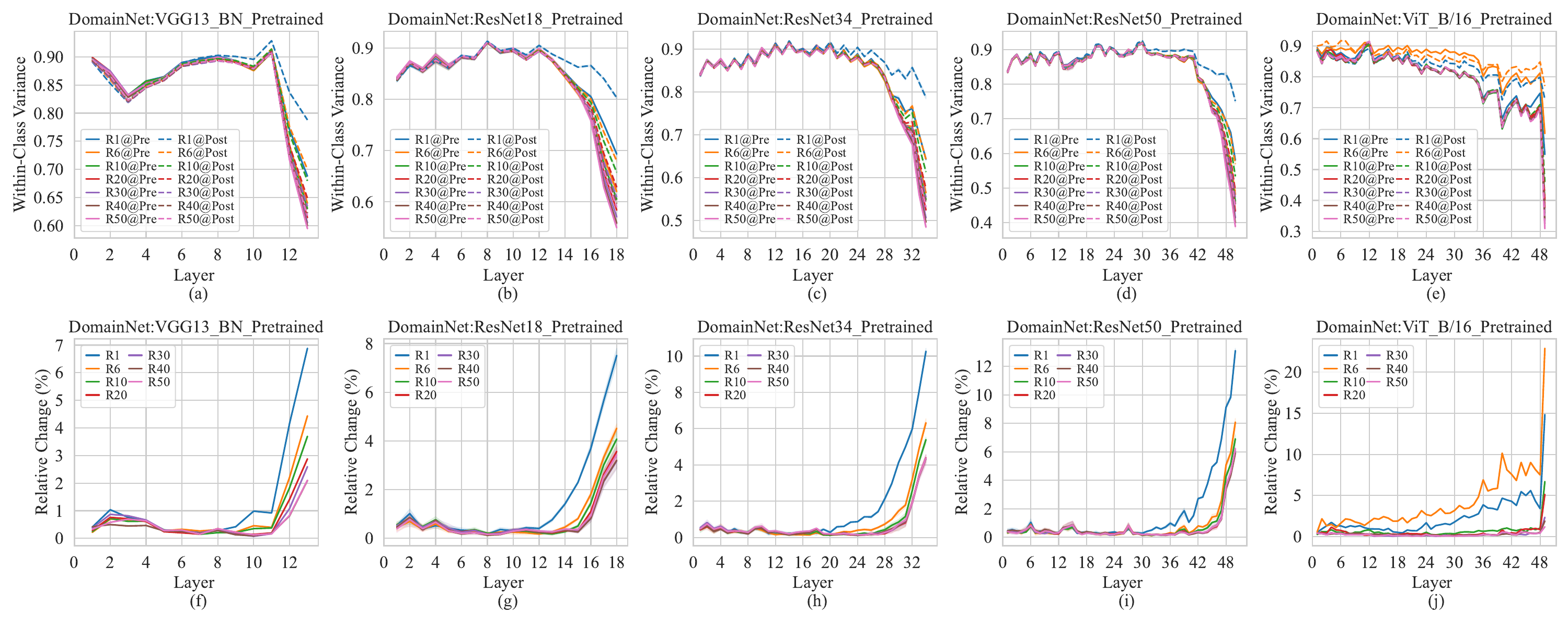}
\caption{Changes in the normalized within-class variance of features across model layers for specific global rounds, with larger X-axis values indicating deeper layers. The model is trained on DomainNet with multiple models that are initialized by parameters pre-trained on large-scaled datasets. The top half of the figure shows the normalized within-class variance, while the bottom half displays the relative change in variance before and after model aggregation.}
\label{FedAvg_AllModel_Pretrained_DomainNet_Epochwise_NC1_NC1}
\end{figure}

\begin{figure}[H]
\centering
\hspace*{-1.8cm}
\includegraphics[width=6.8in]{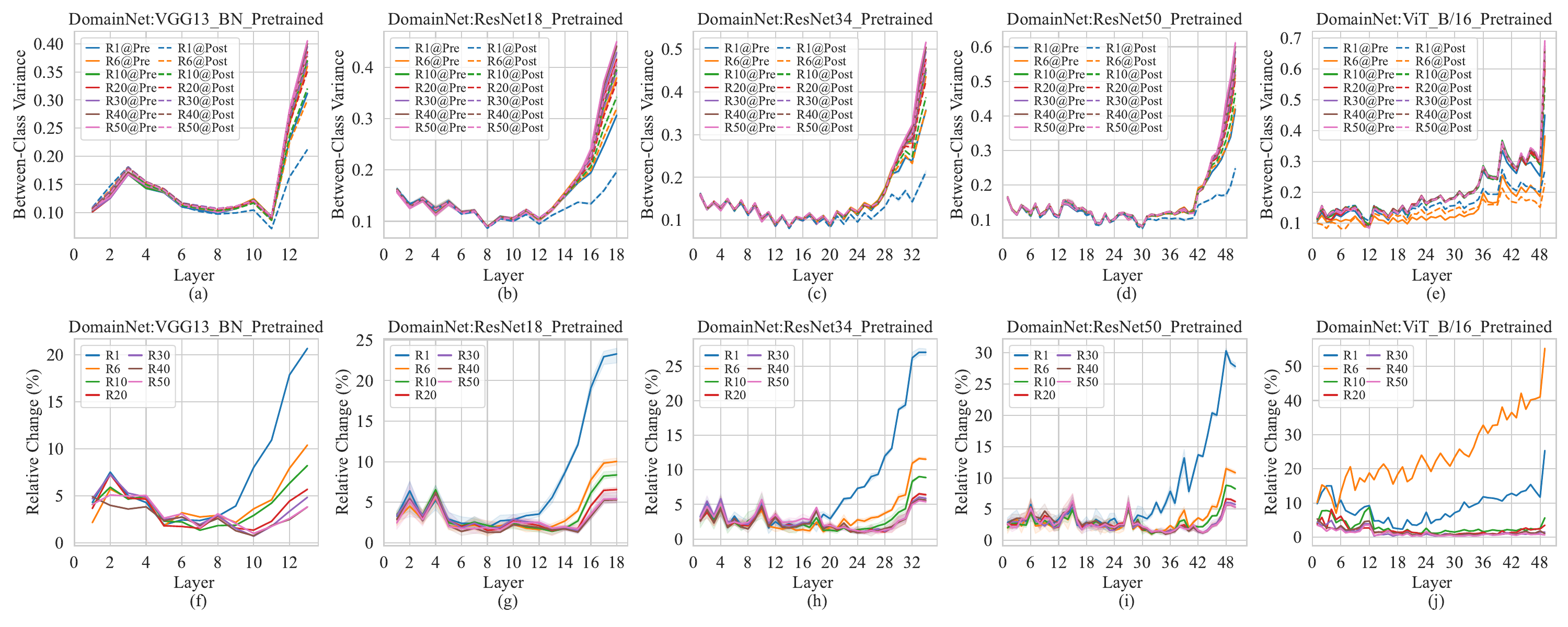}
\caption{Changes in the normalized between-class variance of features across model layers for specific global rounds, with larger X-axis values indicating deeper layers. The model is trained on DomainNet with multiple models that are initialized by parameters pre-trained on large-scaled datasets. The top half of the figure shows the normalized between-class variance, while the bottom half displays the relative change in variance before and after model aggregation.}
\label{FedAvg_AllModel_Pretrained_DomainNet_Epochwise_NC1_between}
\end{figure}

\begin{figure}[H]
\centering
\hspace*{-1.8cm}
\includegraphics[width=6.8in]{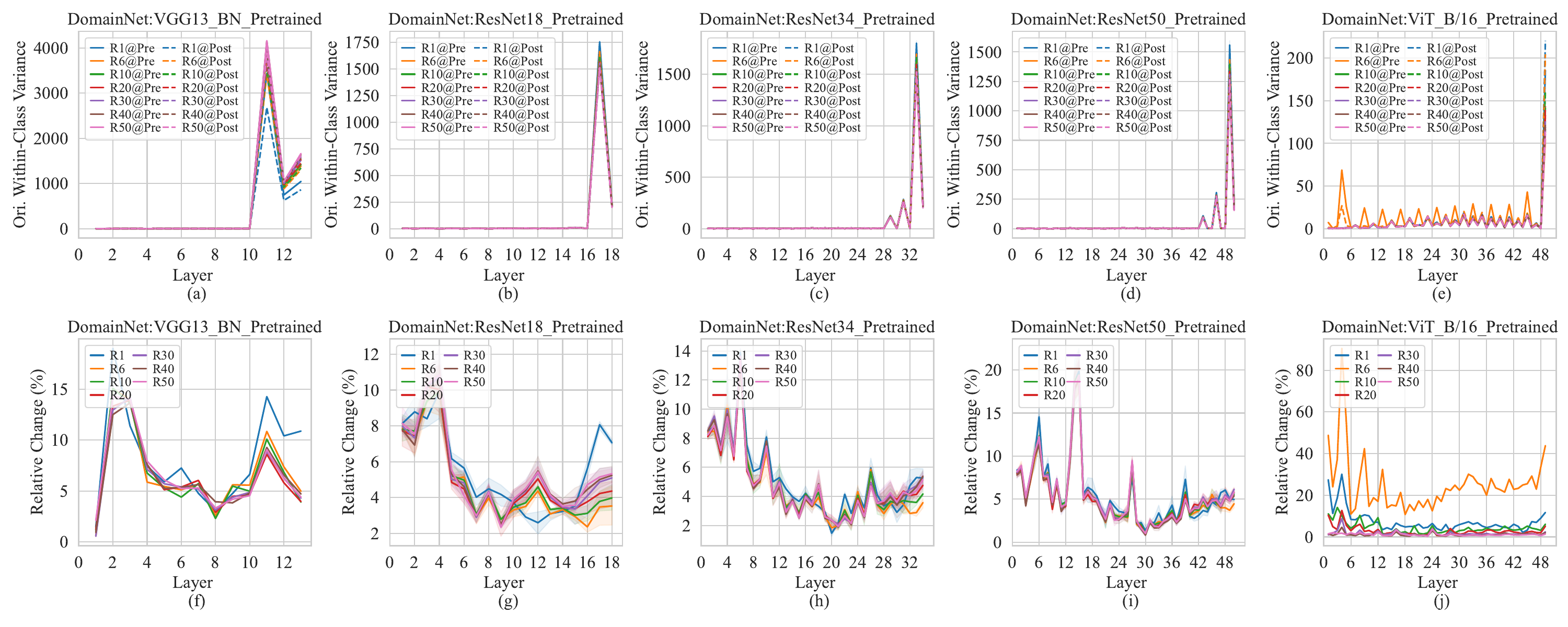}
\caption{Changes in the unnormalized within-class variance of features across model layers for specific global rounds, with larger X-axis values indicating deeper layers. The model is trained on DomainNet with multiple models that are initialized by parameters pre-trained on large-scaled datasets. The top half of the figure shows the original unnormalized within-class variance, while the bottom half displays the relative change in variance before and after model aggregation.}
\label{FedAvg_AllModel_Pretrained_DomainNet_Epochwise_NC1_trace_within}
\end{figure}

\begin{figure}[H]
\centering
\hspace*{-1.8cm}
\includegraphics[width=6.8in]{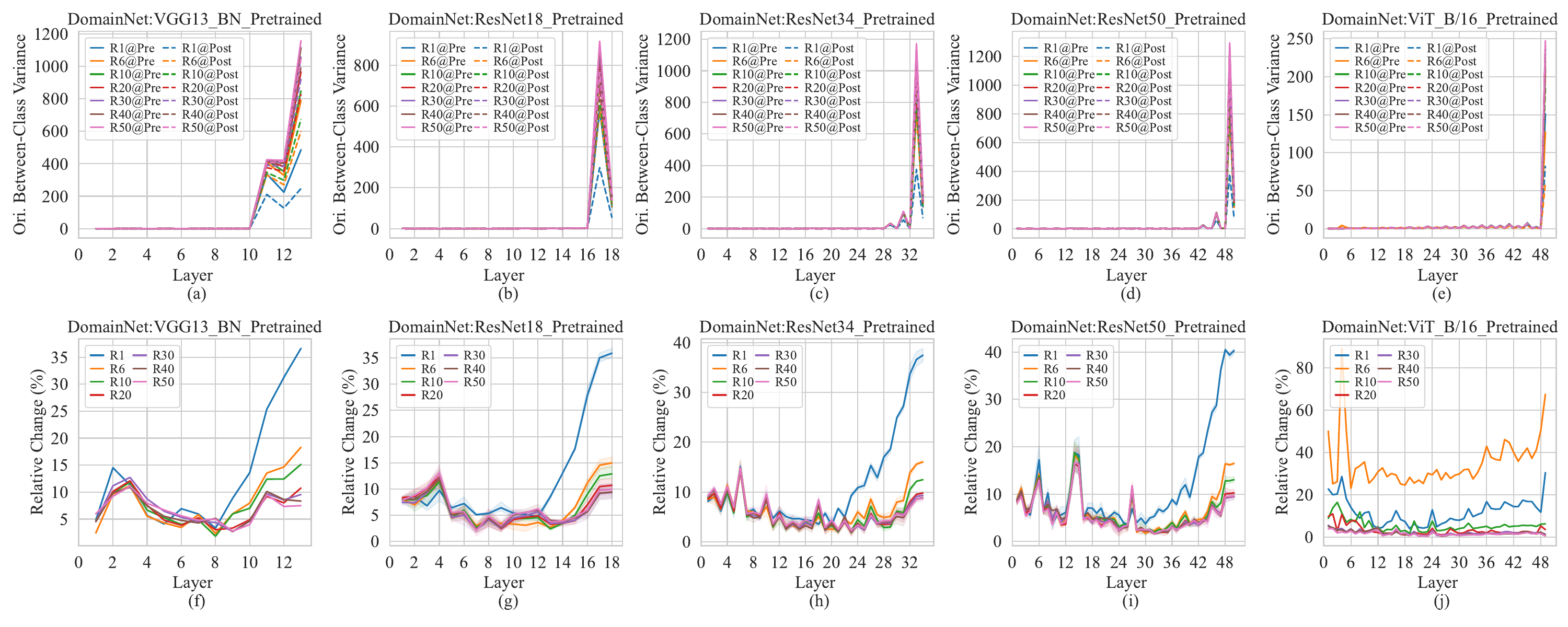}
\caption{Changes in the unnormalized between-class variance of features across model layers for specific global rounds, with larger X-axis values indicating deeper layers. The model is trained on DomainNet with multiple models that are initialized by parameters pre-trained on large-scaled datasets. The top half of the figure shows the original unnormalized between-class variance, while the bottom half displays the relative change in variance before and after model aggregation.}
\label{FedAvg_AllModel_Pretrained_DomainNet_Epochwise_NC1_trace_between}
\end{figure}

\subsection{Changes of Feature Variance Across Training Rounds}

\begin{figure}[H]
\centering
\includegraphics[width=3.0in]{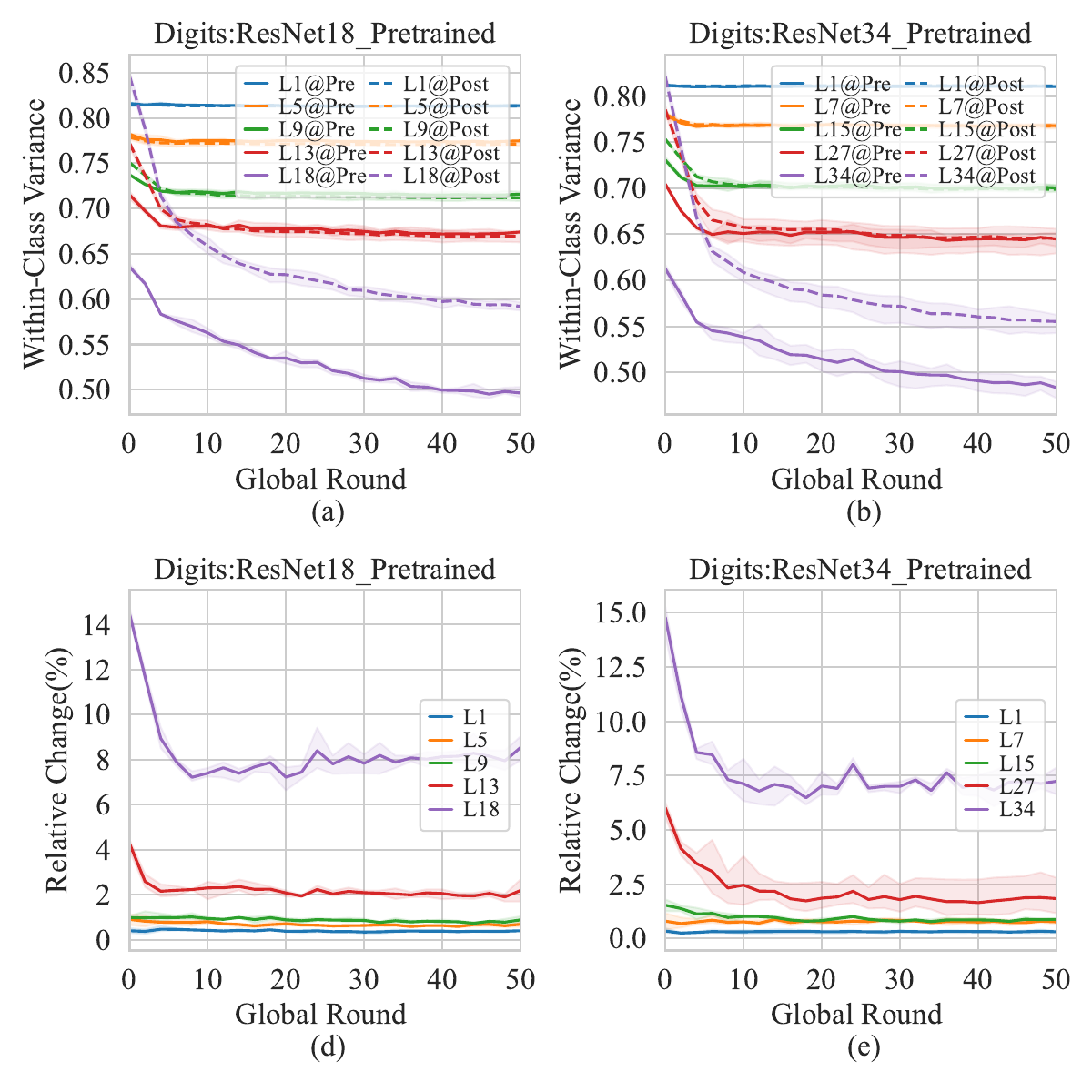}
\caption{Changes in the normalized within-class variance of features across FL training at specific model layers. The model is trained on Digit-Five with multiple models that are initialized by parameters pre-trained on large-scaled datasets. The top half of the figure shows the original normalized within-class variance, while the bottom half displays the relative change in variance before and after model aggregation.}
\label{FedAvg_AllModel_Pretrained_Digits_Layerwise_NC1_NC1}
\end{figure}

\begin{figure}[H]
\centering
\includegraphics[width=3.0in]{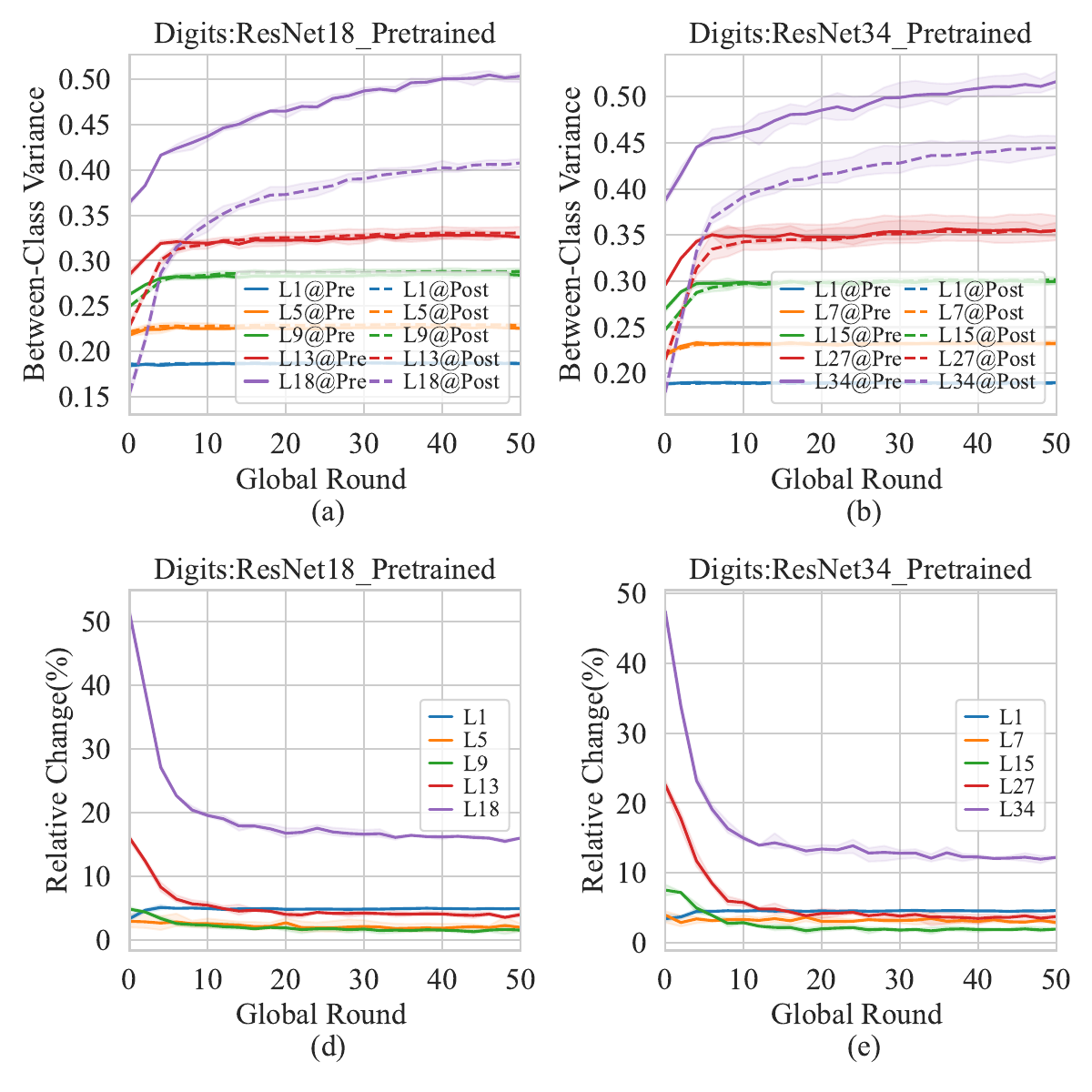}
\caption{Changes in the normalized between-class variance of features across FL training at specific model layers. The model is trained on Digit-Five with multiple models that are initialized by parameters pre-trained on large-scaled datasets. The top half of the figure shows the normalized within-class variance, while the bottom half displays the relative change in variance before and after model aggregation.}
\label{FedAvg_AllModel_Pretrained_Digits_Layerwise_NC1_between}
\end{figure}

\begin{figure}[H]
\centering
\includegraphics[width=3.0in]{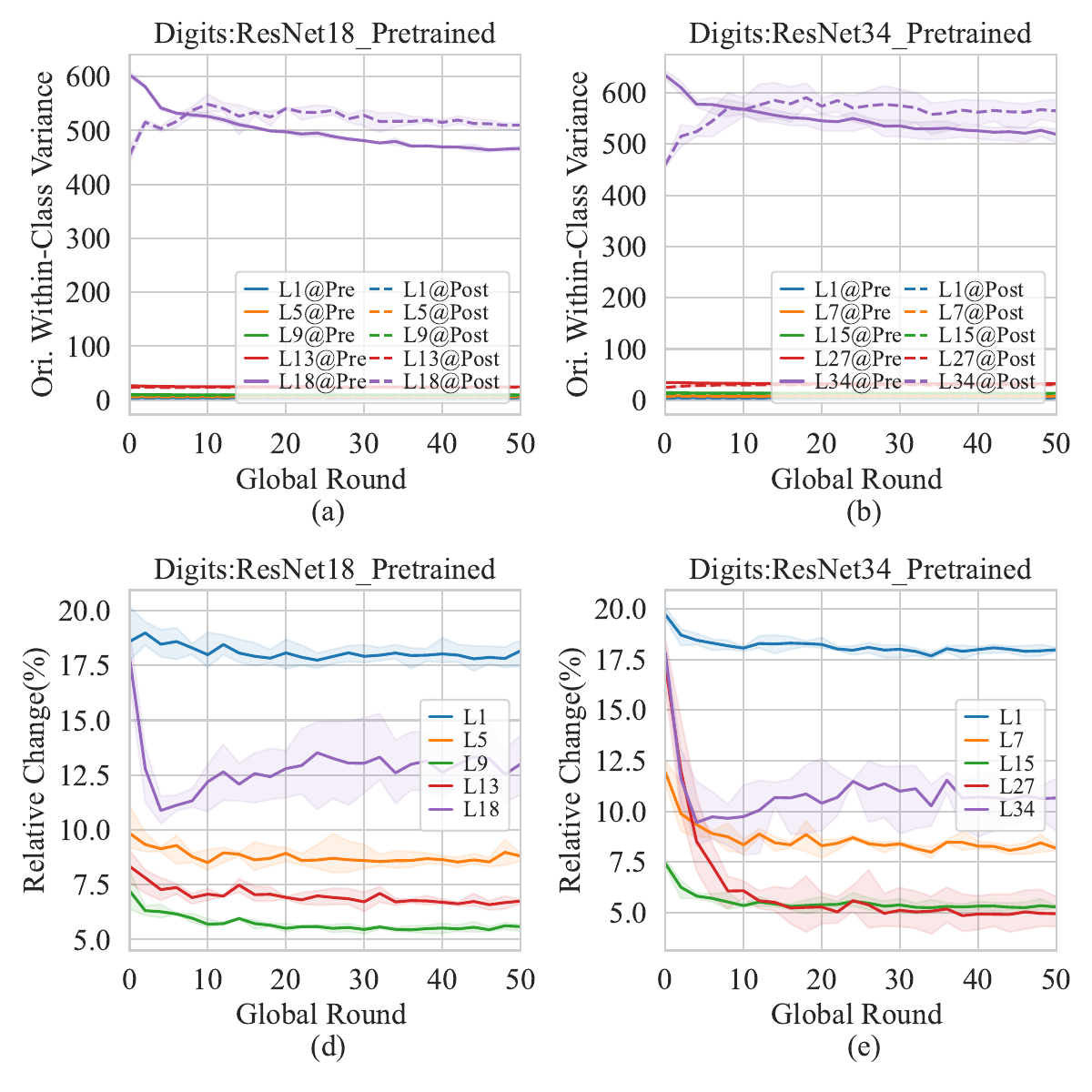}
\caption{Changes in the original unnormalized within-class variance of features across FL training at specific model layers. The model is trained on Digit-Five with multiple models that are initialized by parameters pre-trained on large-scaled datasets. The top half of the figure shows the original unnormalized within-class variance, while the bottom half displays the relative change in variance before and after model aggregation.}
\label{FedAvg_AllModel_Pretrained_Digits_Layerwise_NC1_trace_within}
\end{figure}

\begin{figure}[H]
\centering
\includegraphics[width=3.0in]{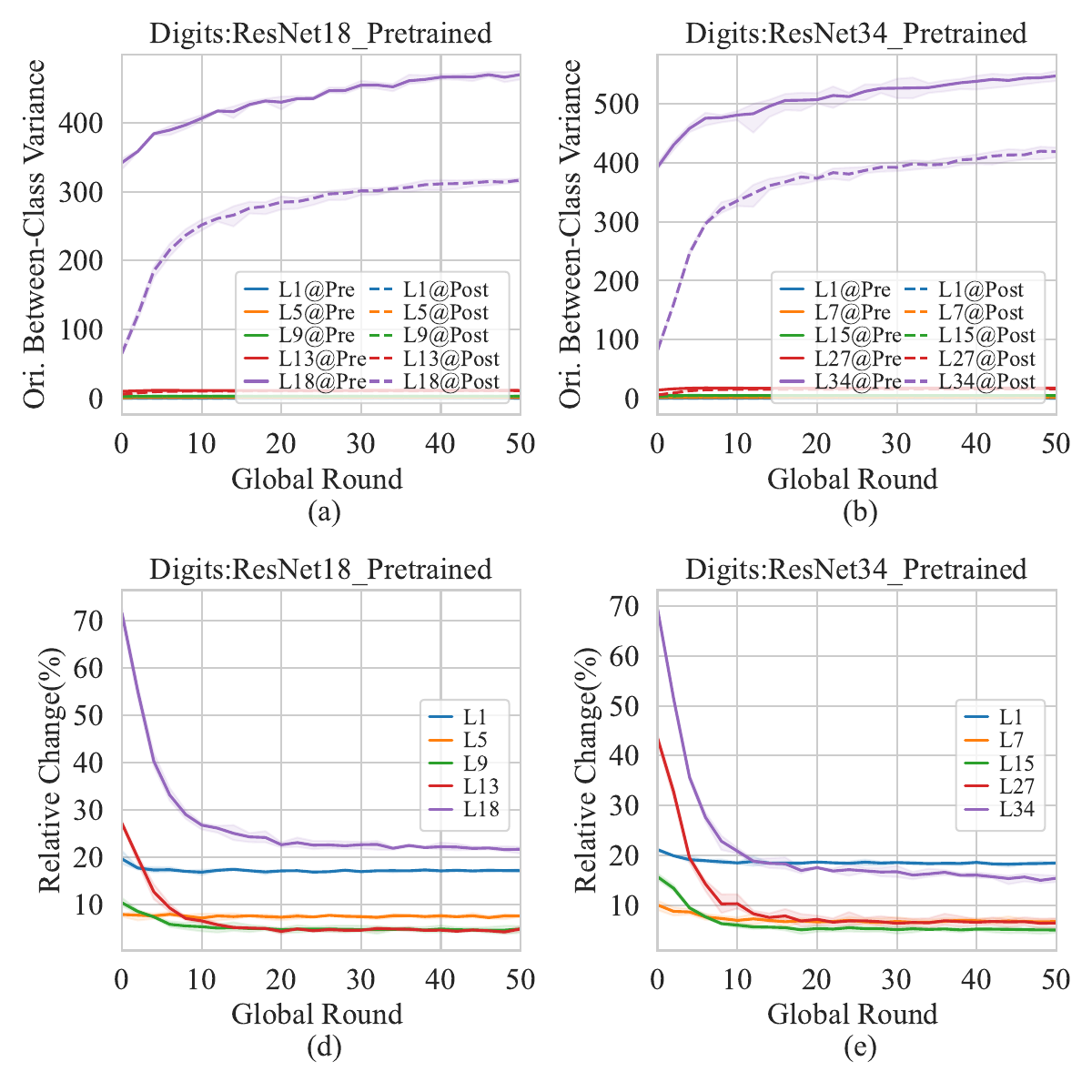}
\caption{Changes in the original unnormalized between-class variance of features across FL training at specific model layers. The model is trained on Digit-Five with multiple models that are initialized by parameters pre-trained on large-scaled datasets. The top half of the figure shows the original unnormalized between-class variance, while the bottom half displays the relative change in variance before and after model aggregation.}
\label{FedAvg_AllModel_Pretrained_Digits_Layerwise_NC1_trace_between}
\end{figure}

\begin{figure}[H]
\centering
\hspace*{-1.8cm}
\includegraphics[width=6.8in]{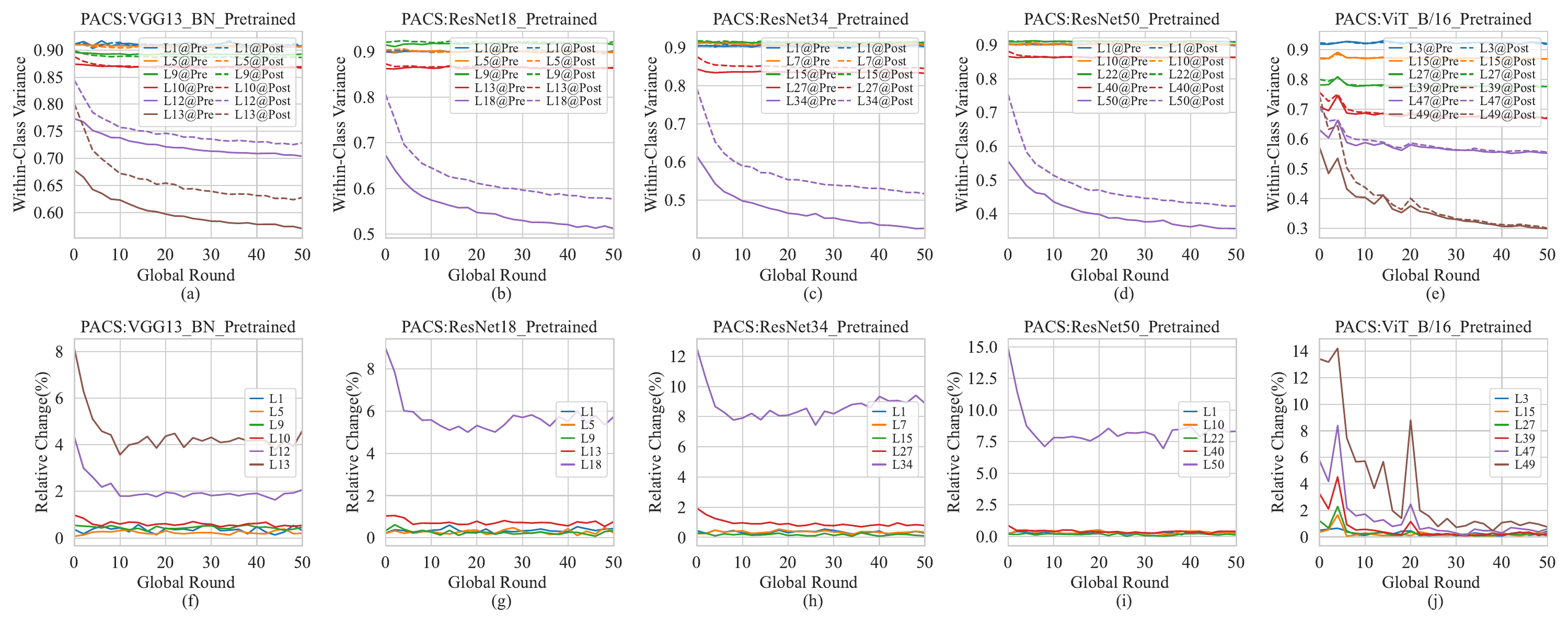}
\caption{Changes in the normalized within-class variance of features across FL training at specific model layers. The model is trained on PACS with multiple models that are initialized by parameters pre-trained on large-scaled datasets. The top half of the figure shows the normalized within-class variance, while the bottom half displays the relative change in variance before and after model aggregation.}
\label{FedAvg_AllModel_Pretrained_PACS_Layerwise_NC1_NC1}
\end{figure}

\begin{figure}[H]
\centering
\hspace*{-1.8cm}
\includegraphics[width=6.8in]{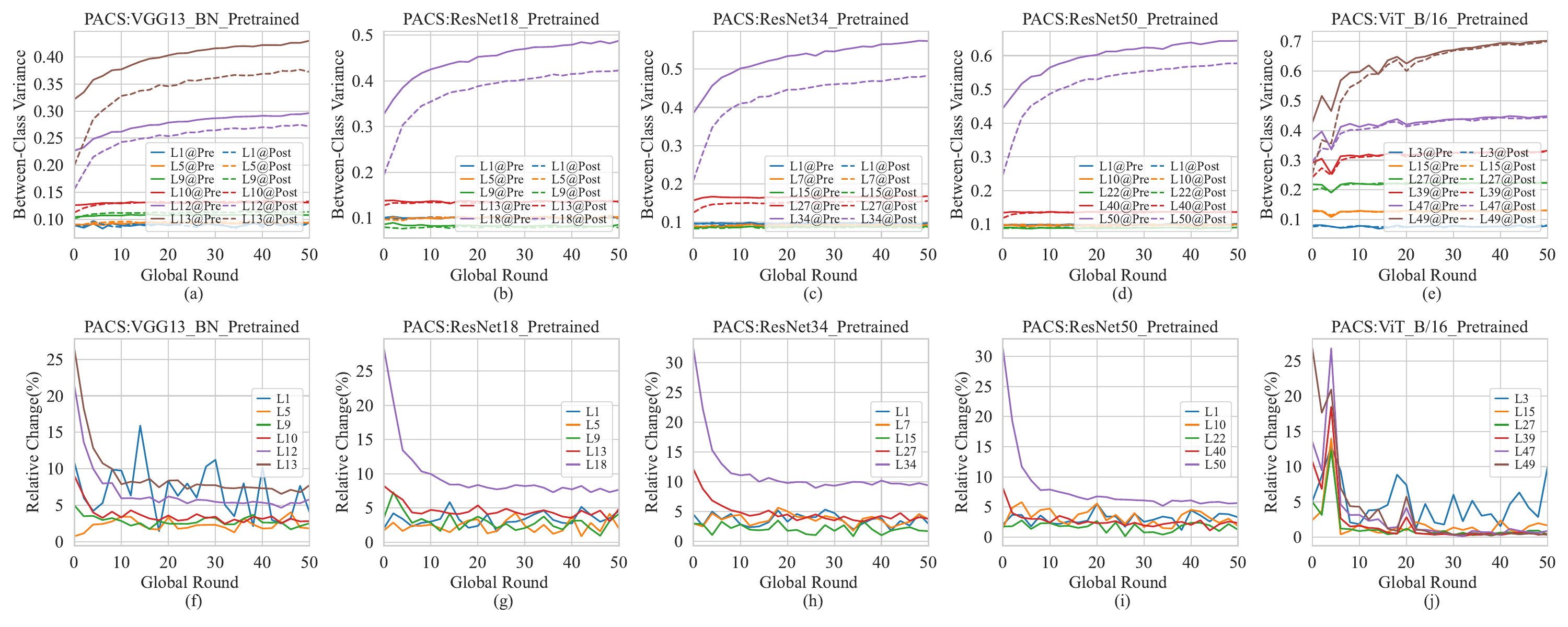}
\caption{Changes in the normalized between-class variance of features across FL training at specific model layers. The model is trained on PACS with multiple models that are initialized by parameters pre-trained on large-scaled datasets. The top half of the figure shows the normalized between-class variance, while the bottom half displays the relative change in variance before and after model aggregation.}
\label{FedAvg_AllModel_Pretrained_PACS_Layerwise_NC1_between}
\end{figure}

\begin{figure}[H]
\centering
\hspace*{-1.8cm}
\includegraphics[width=6.8in]{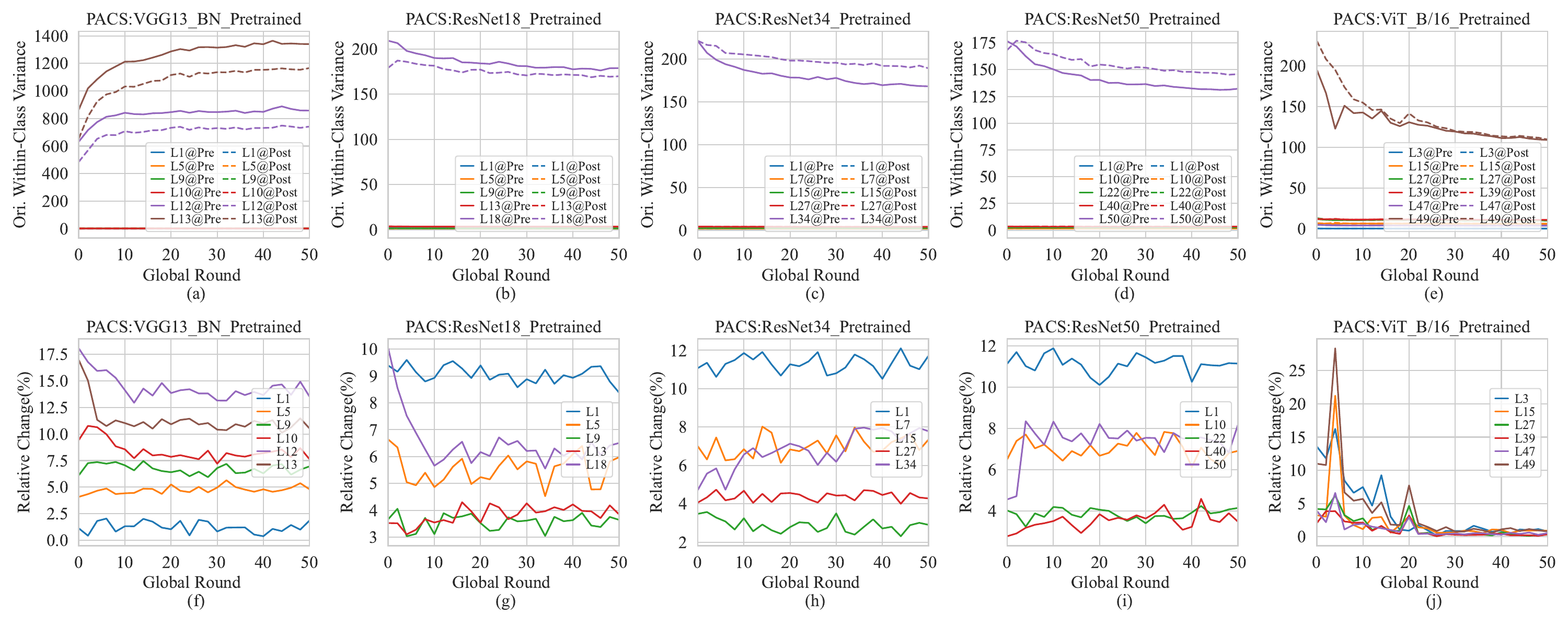}
\caption{Changes in the original unnormalized within-class variance of features across FL training at specific model layers. The model is trained on PACS with multiple models that are initialized by parameters pre-trained on large-scaled datasets. The top half of the figure shows the original unnormalized within-class variance, while the bottom half displays the relative change in variance before and after model aggregation.}
\label{FedAvg_AllModel_Pretrained_PACS_Layerwise_NC1_trace_within}
\end{figure}

\begin{figure}[H]
\centering
\hspace*{-1.8cm}
\includegraphics[width=6.8in]{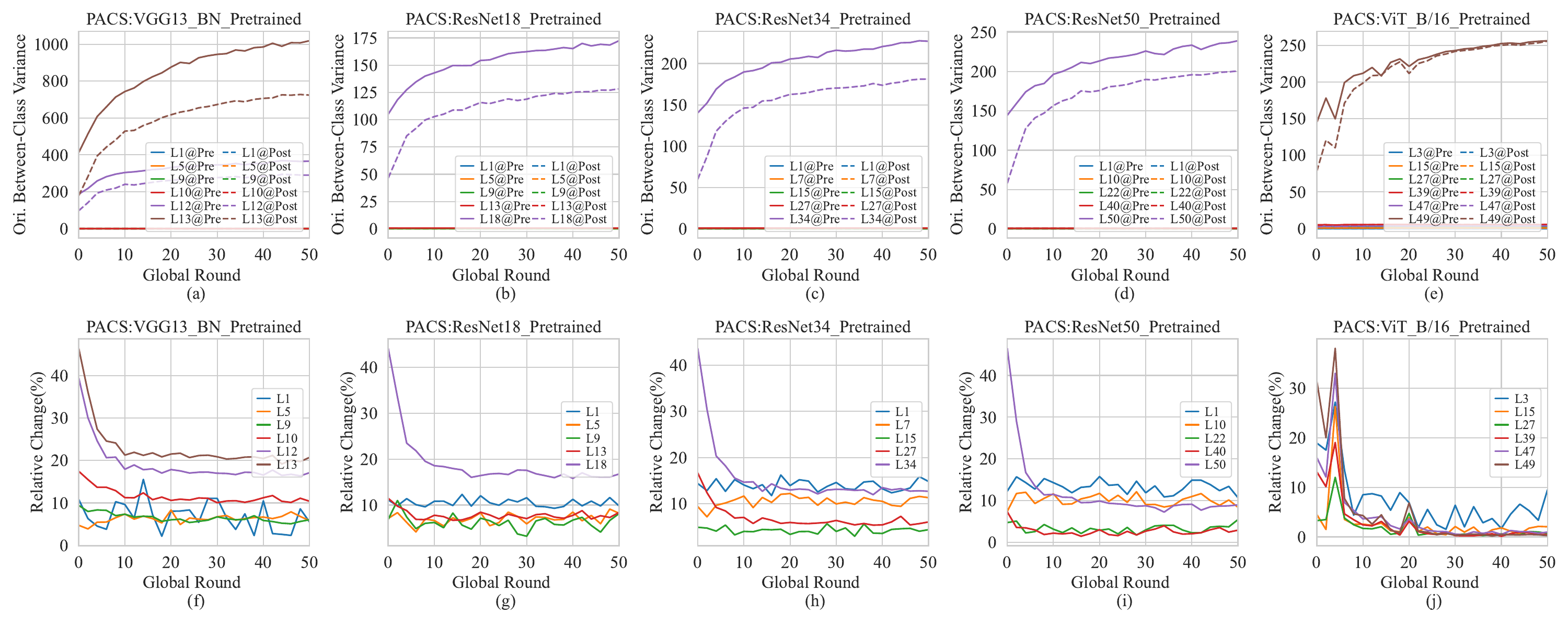}
\caption{Changes in the original unnormalized between-class variance of features across FL training at specific model layers. The model is trained on PACS with multiple models that are initialized by parameters pre-trained on large-scaled datasets. The top half of the figure shows the original unnormalized between-class variance, while the bottom half displays the relative change in variance before and after model aggregation.}
\label{FedAvg_AllModel_Pretrained_PACS_Layerwise_NC1_trace_between}
\end{figure}

\begin{figure}[H]
\centering
\hspace*{-1.8cm}
\includegraphics[width=6.8in]{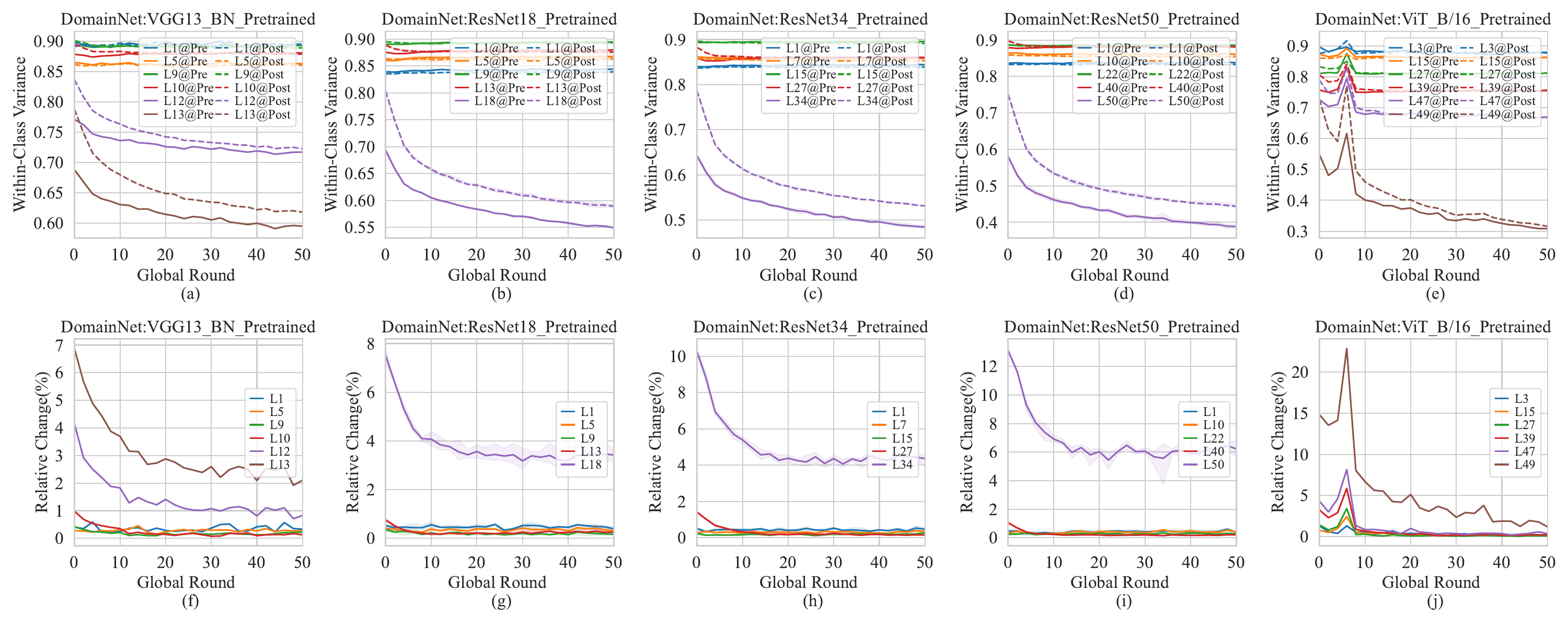}
\caption{Changes in the normalized within-class variance of features across FL training at specific model layers. The model is trained on DomainNet with multiple models that are initialized by parameters pre-trained on large-scaled datasets. The top half of the figure shows the normalized within-class variance, while the bottom half displays the relative change in variance before and after model aggregation.}
\label{FedAvg_AllModel_Pretrained_DomainNet_Layerwise_NC1_NC1}
\end{figure}

\begin{figure}[H]
\centering
\hspace*{-1.8cm}
\includegraphics[width=6.8in]{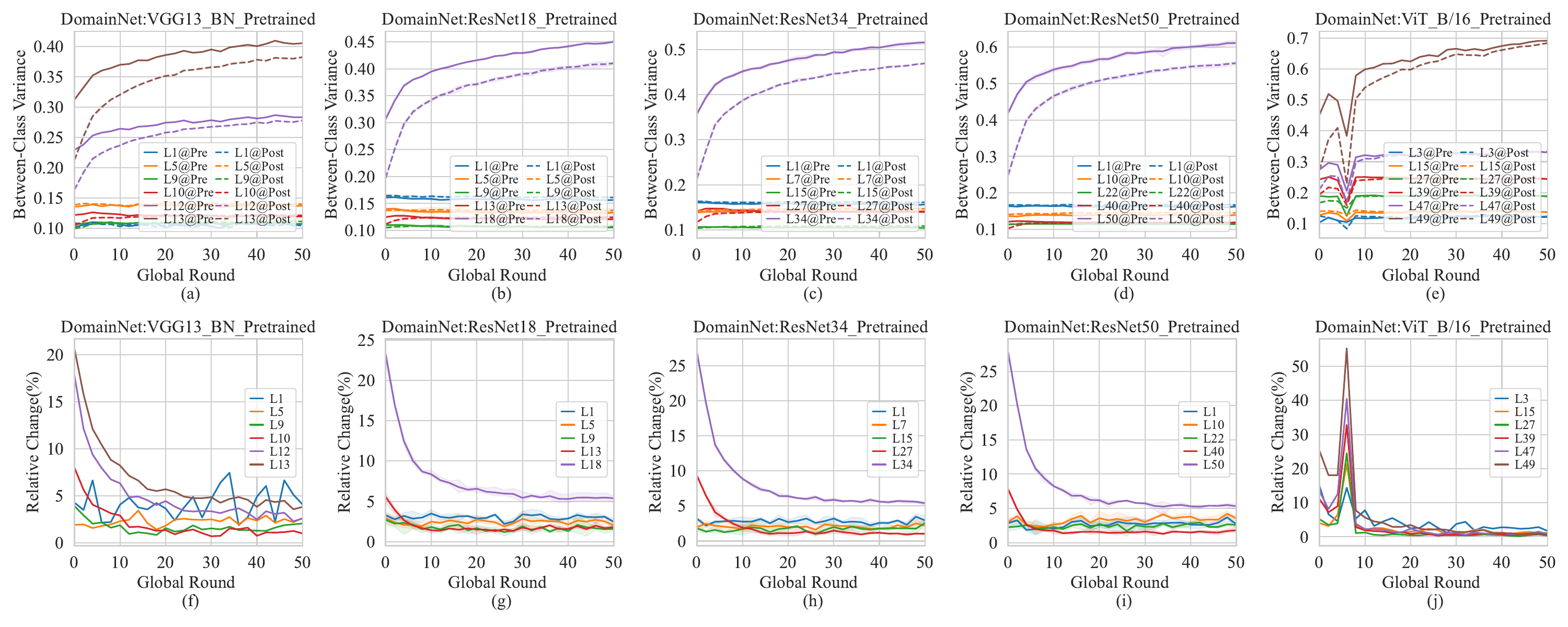}
\caption{Changes in the normalized between-class variance of features across FL training at specific model layers. The model is trained on DomainNet with multiple models that are initialized by parameters pre-trained on large-scaled datasets. The top half of the figure shows the normalized between-class variance, while the bottom half displays the relative change in variance before and after model aggregation.}
\label{FedAvg_AllModel_Pretrained_DomainNet_Layerwise_NC1_between}
\end{figure}

\begin{figure}[H]
\centering
\hspace*{-1.8cm}
\includegraphics[width=6.8in]{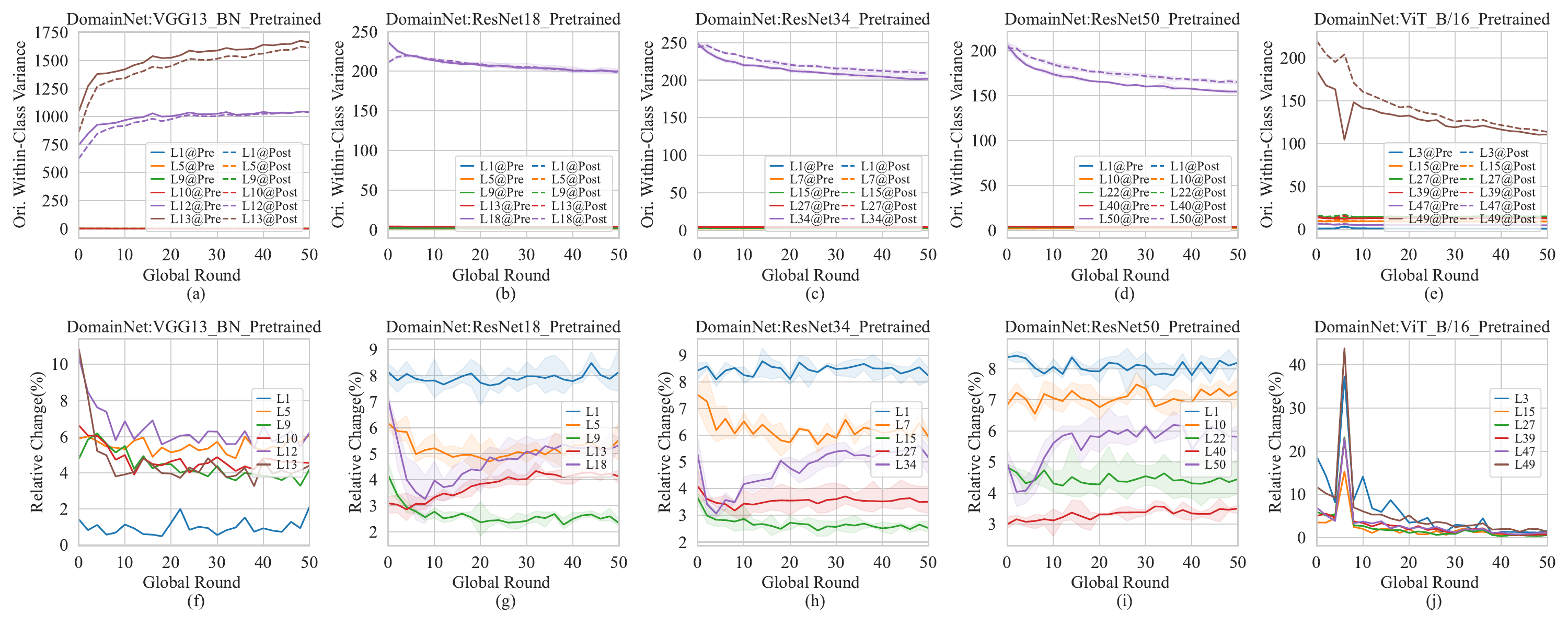}
\caption{Changes in the original unnormalized within-class variance of features across FL training at specific model layers. The model is trained on DomainNet with multiple models that are initialized by parameters pre-trained on large-scaled datasets. The top half of the figure shows the original unnormalized within-class variance, while the bottom half displays the relative change in variance before and after model aggregation.}
\label{FedAvg_AllModel_Pretrained_DomainNet_Layerwise_NC1_trace_within}
\end{figure}

\begin{figure}[H]
\centering
\hspace*{-1.8cm}
\includegraphics[width=6.8in]{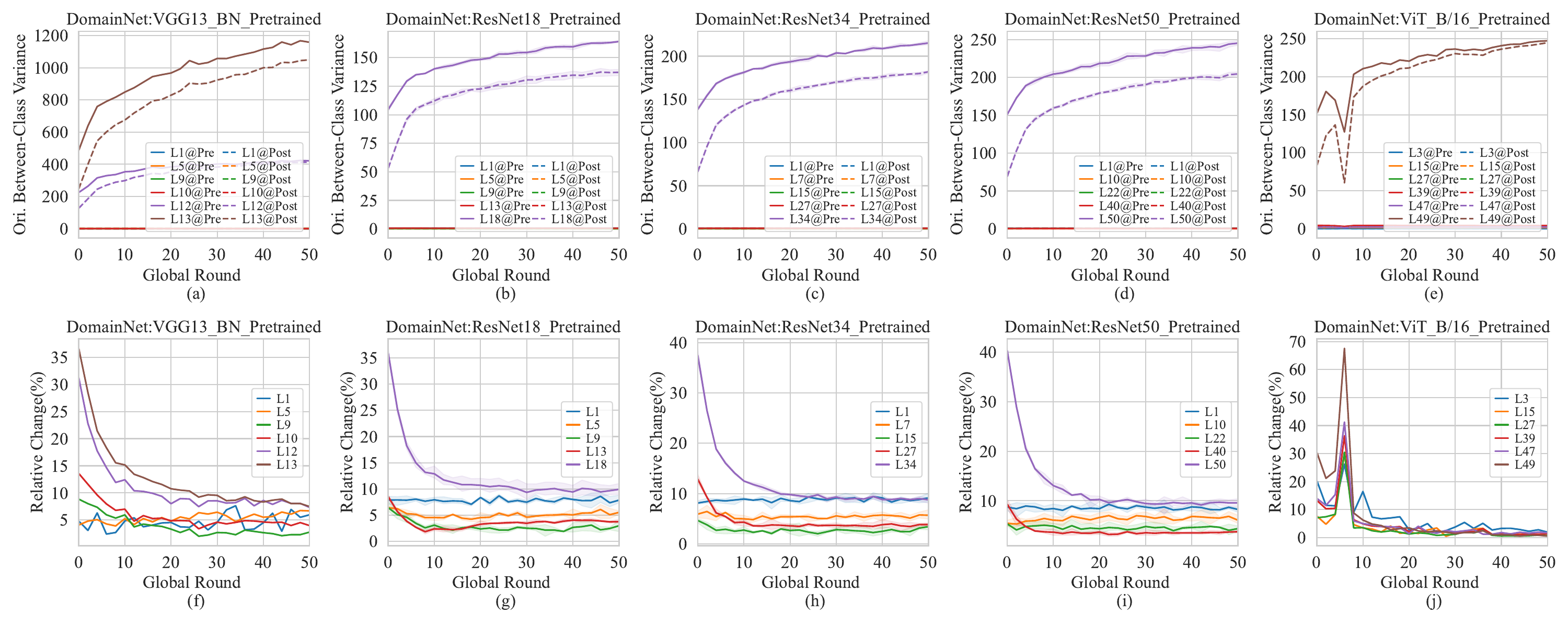}
\caption{Changes in the original unnormalized between-class variance of features across FL training at specific model layers. The model is trained on DomainNet with multiple models that are initialized by parameters pre-trained on large-scaled datasets. The top half of the figure shows the original unnormalized between-class variance, while the bottom half displays the relative change in variance before and after model aggregation.}
\label{FedAvg_AllModel_Pretrained_DomainNet_Layerwise_NC1_trace_between}
\end{figure}

\subsection{Changes of Alignment Across Layers}

\begin{figure}[H]
\centering
\includegraphics[width=3.0in]{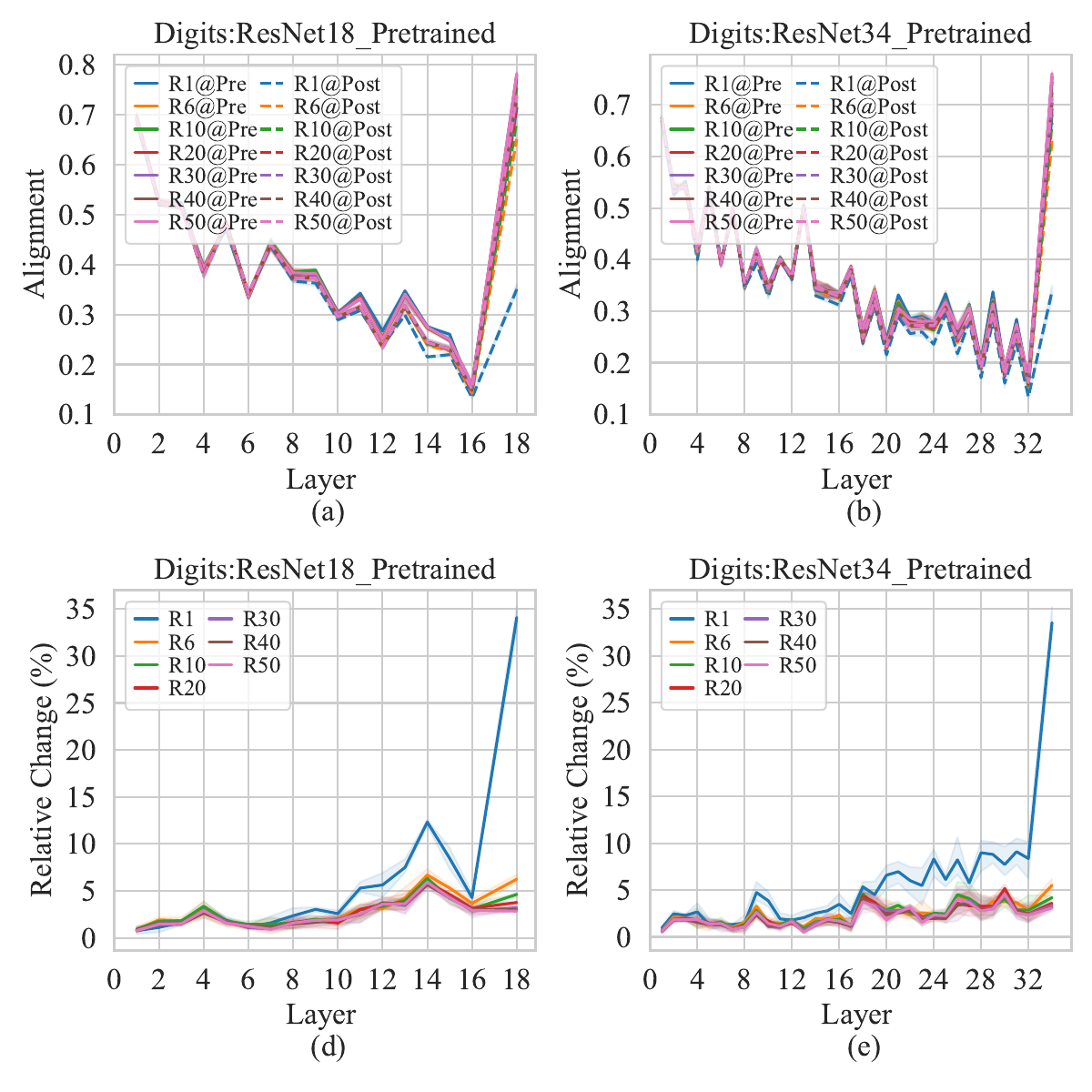}
\caption{Changes in the alignment between features and parameters across model layers for specific global rounds, with larger X-axis values indicating deeper layers. The model is trained on Digit-Five with multiple models that are initialized by parameters pre-trained on large-scaled datasets. The top half of the figure shows the original alignment values between features and parameters, while the bottom half displays the relative change in alignment values before and after model aggregation.}
\label{FedAvg_AllModel_Pretrained_Digits_Epochwise_NC3_NC3}
\end{figure}

\begin{figure}[H]
\centering
\hspace*{-1.8cm}
\includegraphics[width=6.8in]{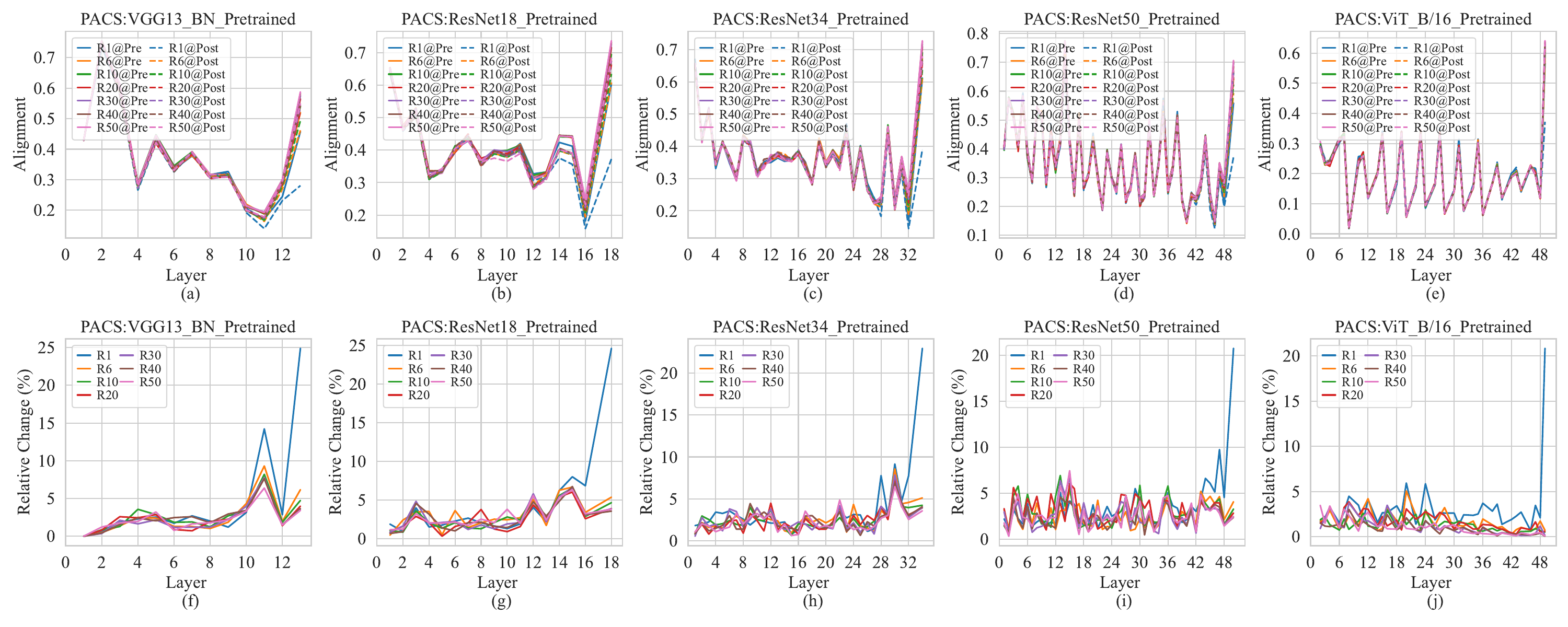}
\caption{Changes in the alignment between features and parameters across model layers for specific global rounds, with larger X-axis values indicating deeper layers. The model is trained on PACS with multiple models that are initialized by parameters pre-trained on large-scaled datasets. The top half of the figure shows the original alignment values between features and parameters, while the bottom half displays the relative change in alignment before and after model aggregation.}
\label{FedAvg_AllModel_Pretrained_PACS_Epochwise_NC3_NC3}
\end{figure}

\begin{figure}[H]
\centering
\hspace*{-1.8cm}
\includegraphics[width=6.8in]{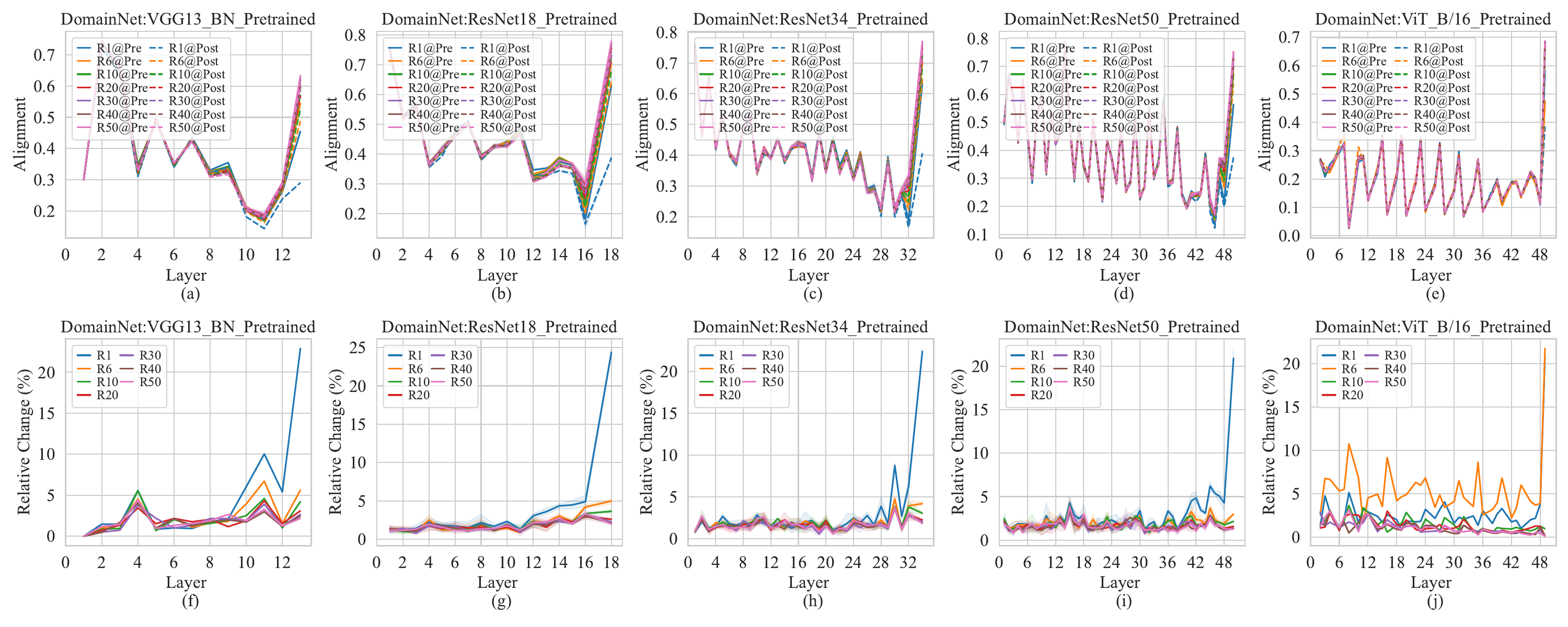}
\caption{Changes in the alignment between features and parameters across model layers for specific global rounds, with larger X-axis values indicating deeper layers. The model is trained on DomainNet with multiple models that are initialized by parameters pre-trained on large-scaled datasets. The top half of the figure shows the original alignment values between features and parameters, while the bottom half displays the relative change in alignment before and after model aggregation.}
\label{FedAvg_AllModel_Pretrained_DomainNet_Epochwise_NC3_NC3}
\end{figure}

\subsection{Changes of Alignment Across Training Rounds}

\begin{figure}[H]
\centering
\includegraphics[width=3.0in]{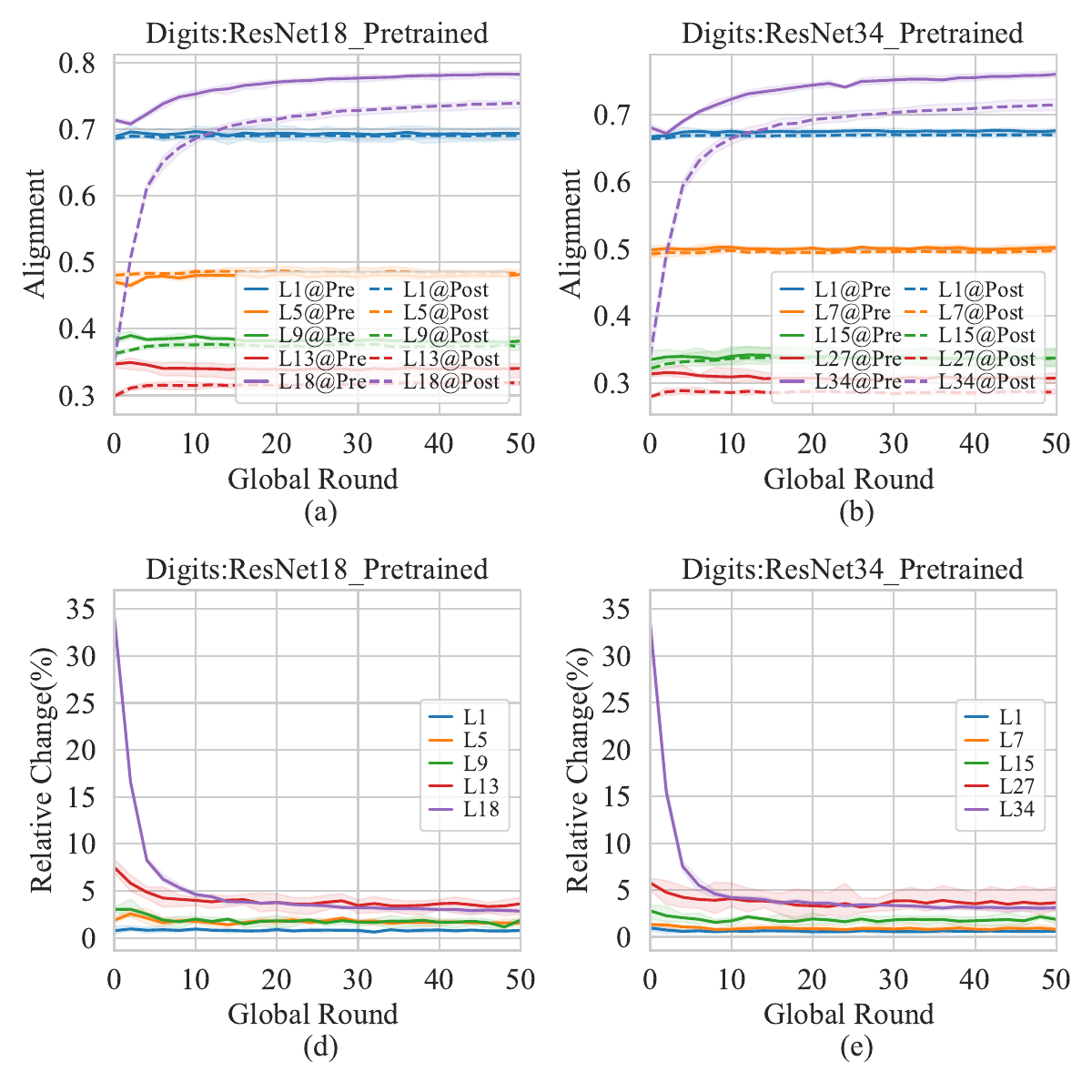}
\caption{Changes in the alignment between features and parameters across FL training at specific model layers. The model is trained on Digit-Five with multiple models that are initialized by parameters pre-trained on large-scaled datasets. The top half of the figure shows the original alignment values between features and parameters, while the bottom half displays the relative change in alignment before and after model aggregation.}
\label{FedAvg_AllModel_Pretrained_Digits_Layerwise_NC3_NC3}
\end{figure}

\begin{figure}[H]
\centering
\hspace*{-1.8cm}
\includegraphics[width=6.8in]{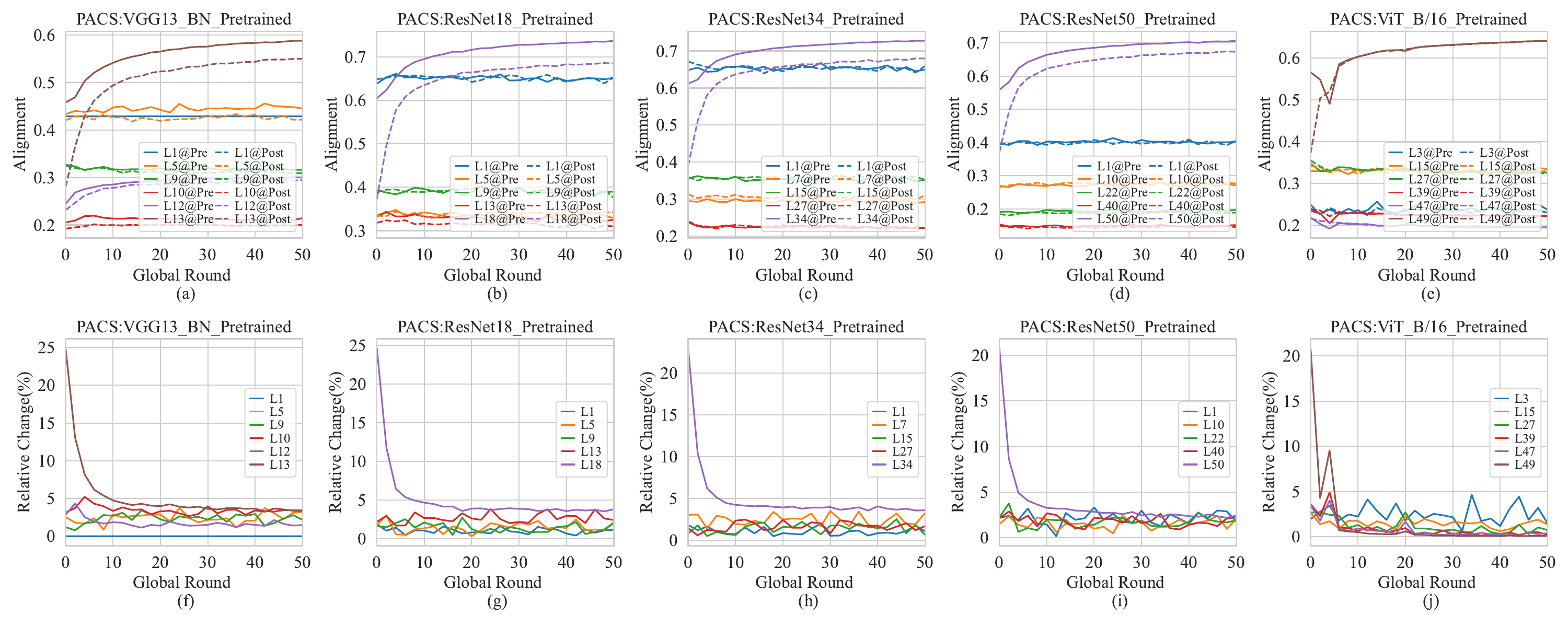}
\caption{Changes in the alignment between features and parameters across FL training at specific model layers. The model is trained on PACS with multiple models that are initialized by parameters pre-trained on large-scaled datasets. The top half of the figure shows the original alignment values between features and parameters, while the bottom half displays the relative change in alignment before and after model aggregation.}
\label{FedAvg_AllModel_Pretrained_PACS_Layerwise_NC3_NC3}
\end{figure}

\begin{figure}[H]
\centering
\hspace*{-1.8cm}
\includegraphics[width=6.8in]{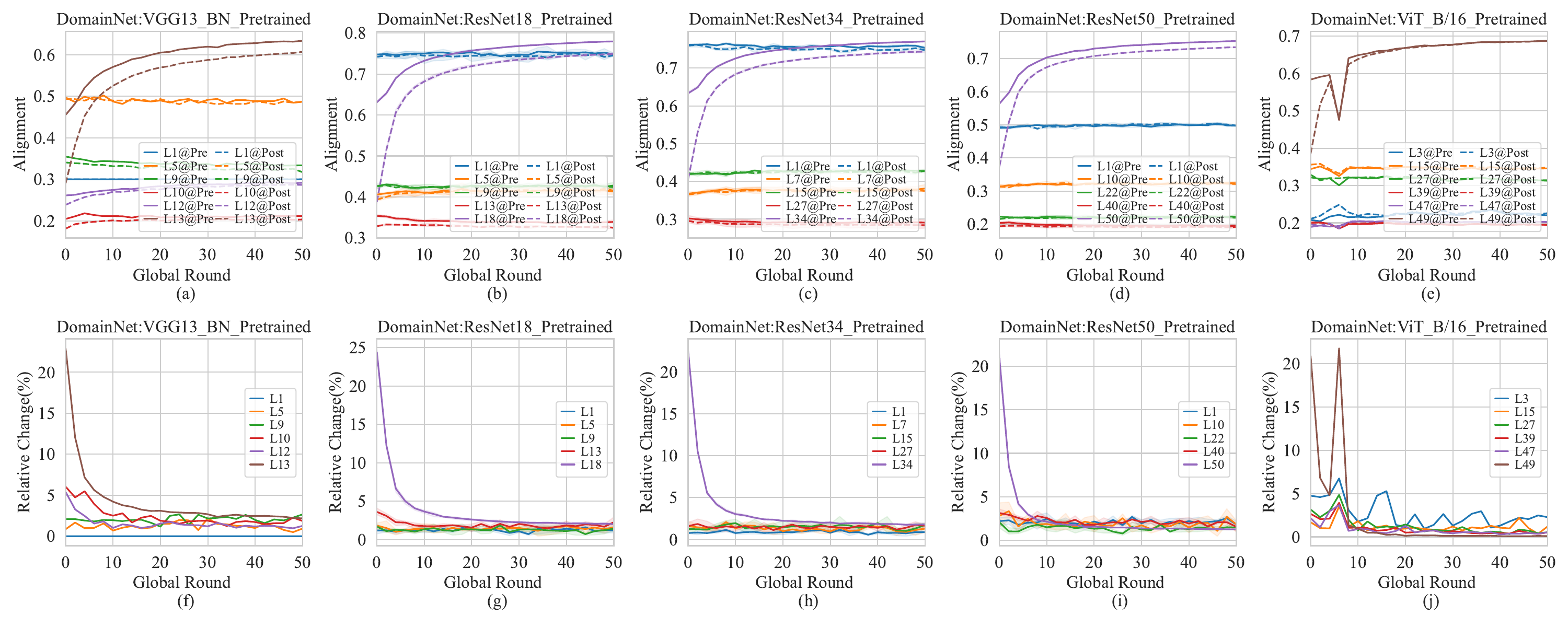}
\caption{Changes in the alignment between features and parameters across FL training at specific model layers. The model is trained on DomainNet with multiple models that are initialized by parameters pre-trained on large-scaled datasets. The top half of the figure shows the original alignment values between features and parameters, while the bottom half displays the relative change in alignment before and after model aggregation.}
\label{FedAvg_AllModel_Pretrained_DomainNet_Layerwise_NC3_NC3}
\end{figure}

\subsection{Visualization of Pre-aggregated and Post-aggregated Features}
\begin{figure}[H]
\centering
\hspace*{-1.8cm}
\includegraphics[width=6.8in]{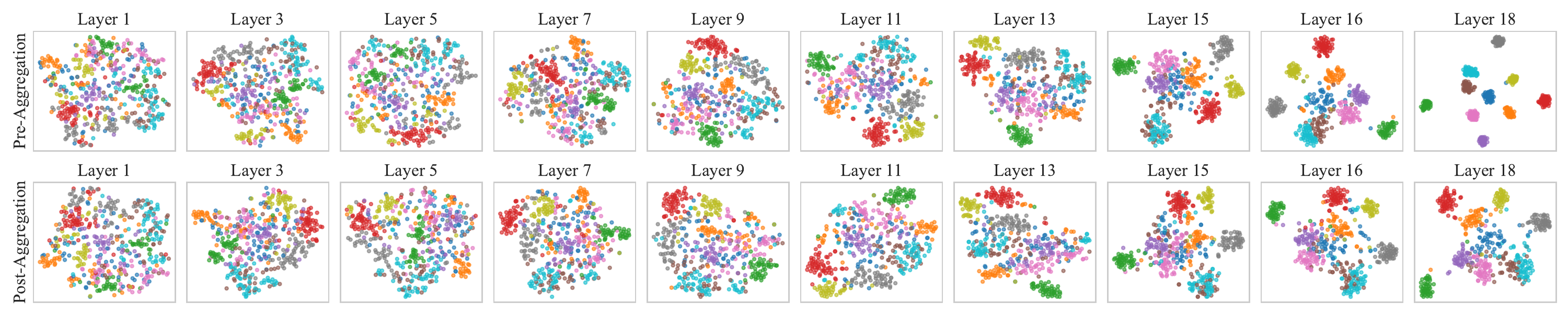}
\caption{T-SNE visualization of features at different layers on the `Quickdraw' domain of DomainNet before and after aggregation.
The features are extracted from ResNet18 in the final global round of FL training, whose parameters are initialized by the parameters pre-trained on large-scaled datasets at the beginning.}
\label{TSNE_DomainNet_quickdraw_ResNet18_Pretrained_10Layers}
\end{figure}

\begin{figure}[H]
\centering
\hspace*{-1.8cm}
\includegraphics[width=6.8in]{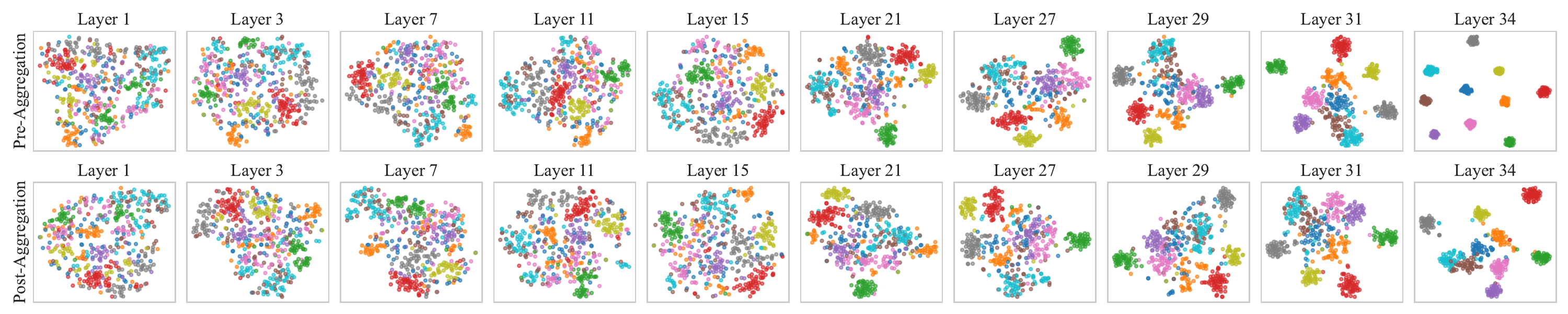}
\caption{T-SNE visualization of features at different layers on the `Quickdraw' domain of DomainNet before and after aggregation.
The features are extracted from ResNet34 in the final global round of FL training, whose parameters are initialized by the parameters pre-trained on large-scaled datasets at the beginning.}
\label{TSNE_DomainNet_quickdraw_ResNet34_Pretrained_10Layers}
\end{figure}

\section{Discussion and Limitations}
This work presents a layer-wise feature analysis framework to better understand the impact of model aggregation on feature representations in FL.
Our findings reveal that model aggregation disrupts within-class compactness, between-class separability, and feature-parameter alignment—three key objectives of representation learning during FL training.
Moreover, this degradation accumulates across network depth, resulting in more severe feature distortion in deeper layers.
We further show that widely adopted techniques---such as parameter personalization, pretrained initialization, and classifier fine-tuning---can be reinterpreted through the lens of our framework as strategies that mitigate the degradation and accumulation effects introduced by model aggregation.

Despite its generality, our framework has several limitations:

First, our experiments are primarily conducted on image classification tasks using architectures such as convolutional neural networks and vision transformers. Evaluating the framework on other modalities, such as text data, remains an important direction for future work.
Second, the feature-parameter alignment metric used in this study is more effective for linear layers than for convolutional layers. Developing more suitable alignment metrics tailored to convolutional structures would enhance the precision of our analysis.
Lastly, although we evaluate several classical FL methods like FedBN \citep{FedBN} and FedPer\citep{FedPer}, applying the framework to a broader range of state-of-the-art algorithms could further validate its generality and reveal new insights into their effectiveness.



\newpage
\section*{NeurIPS Paper Checklist}

\begin{enumerate}

\item {\bf Claims}
    \item[] Question: Do the main claims made in the abstract and introduction accurately reflect the paper's contributions and scope?
    \item[] Answer: \answerYes{} 
    \item[] Justification: The abstract and introduction clearly state the paper's contributions, including the layer-wise analysis of feature behavior under model aggregation in FL and reinterpretation of existing techniques using this framework.

\item {\bf Limitations}
    \item[] Question: Does the paper discuss the limitations of the work performed by the authors?
    \item[] Answer: \answerYes{} 
    \item[] Justification: We include a dedicated “Discussions and Limitations” section at the end of the paper.
Specifically, we acknowledge that: (1) our framework is currently evaluated only on vision tasks, and future work should extend it to other modalities such as text or time series;
(2) the feature-parameter alignment metric used is more interpretable and reliable for linear layers, while its effectiveness on convolutional layers remains limited;
and (3) although we analyze several classical FL methods, a broader evaluation on state-of-the-art algorithms is needed to further validate the generality of the framework.

\item {\bf Theory assumptions and proofs}
    \item[] Question: For each theoretical result, does the paper provide the full set of assumptions and a complete (and correct) proof?
    \item[] Answer:  \answerNA{} 
    \item[] Justification: This paper is empirical in nature and does not include formal theorems or proofs.

    \item {\bf Experimental result reproducibility}
    \item[] Question: Does the paper fully disclose all the information needed to reproduce the main experimental results of the paper to the extent that it affects the main claims and/or conclusions of the paper (regardless of whether the code and data are provided or not)?
    \item[] Answer: \answerYes{} 
    \item[] Justification: The paper describes all training details, models, datasets, and feature evaluation metrics used (see Section \ref{evaluation_setup} and Appendix \ref{appendix_dataset_description}-\ref{appendix_metric}). Details for model backbones, optimizers, and federated learning training settings are included.

\item {\bf Open access to data and code}
    \item[] Question: Does the paper provide open access to the data and code, with sufficient instructions to faithfully reproduce the main experimental results, as described in supplemental material?
    \item[] Answer: \answerYes{}
    \item[] Justification: We will include the code used for this paper in the supplemental material, covering all key experiments such as federated training and layer-peeled feature evaluation. All datasets used are publicly available.

\item {\bf Experimental setting/details}
    \item[] Question: Does the paper specify all the training and test details (e.g., data splits, hyperparameters, how they were chosen, type of optimizer, etc.) necessary to understand the results?
    \item[] Answer: \answerYes{} 
    \item[] Justification: Section \ref{evaluation_setup} and Appendix \ref{appendix_dataset_description}-\ref{appendix_metric}) specify data partiation, model architecture, optimizer (including learning rate and momentum), batch size, training rounds, and aggregation settings used in our paper.

\item {\bf Experiment statistical significance}
    \item[] Question: Does the paper report error bars suitably and correctly defined or other appropriate information about the statistical significance of the experiments?
    \item[] Answer: \answerYes{} 
    \item[] Justification: All experimental plots include error bars, which represent the range (minimum to maximum) across three independent runs with different random seeds. These error bars help visualize the stability of feature evaluation metrics under different settings.

\item {\bf Experiments compute resources}
    \item[] Question: For each experiment, does the paper provide sufficient information on the computer resources (type of compute workers, memory, time of execution) needed to reproduce the experiments?
    \item[] Answer: \answerYes{} 
    \item[] Justification: All experiments are conducted using the PyTorch framework \citep{Pytorch} and implemented on a cluster with four NVIDIA V100 (32GB) GPUs. Details are provided in the main paper and Appendix~\ref{implementation_details_appendix}.
    
\item {\bf Code of ethics}
    \item[] Question: Does the research conducted in the paper conform, in every respect, with the NeurIPS Code of Ethics \url{https://neurips.cc/public/EthicsGuidelines}?
    \item[] Answer: \answerYes{} 
    \item[] Justification: The research uses only publicly available datasets and does not involve human subjects or personal data.

\item {\bf Broader impacts}
    \item[] Question: Does the paper discuss both potential positive societal impacts and negative societal impacts of the work performed?
    \item[] Answer: \answerNA{} 
    \item[] Justification: This work focuses on analyzing feature behaviors in federated learning from a representational perspective. It does not involve any direct applications or deployments with immediate societal impact, such as human subjects, sensitive data, or policy-related use cases.
    
\item {\bf Safeguards}
    \item[] Question: Does the paper describe safeguards that have been put in place for responsible release of data or models that have a high risk for misuse (e.g., pretrained language models, image generators, or scraped datasets)?
    \item[] Answer: \answerNA{} 
    \item[] Justification: Our work does not involve high-risk assets such as pretrained generative models or scraped datasets.

\item {\bf Licenses for existing assets}
    \item[] Question: Are the creators or original owners of assets (e.g., code, data, models), used in the paper, properly credited and are the license and terms of use explicitly mentioned and properly respected?
    \item[] Answer: \answerYes{} 
    \item[] Justification: All datasets (Digit-Five, PACS, DomainNet) and pre-trained models are used in accordance with their licenses, and sources are cited appropriately.

\item {\bf New assets}
    \item[] Question: Are new assets introduced in the paper well documented and is the documentation provided alongside the assets?
    \item[] Answer: \answerNo{} 
    \item[] Justification: The paper does not introduce new datasets or pretrained models.

\item {\bf Crowdsourcing and research with human subjects}
    \item[] Question: For crowdsourcing experiments and research with human subjects, does the paper include the full text of instructions given to participants and screenshots, if applicable, as well as details about compensation (if any)? 
    \item[] Answer: \answerNA{} 
    \item[] Justification: This research does not involve crowdsourcing or human participants.

\item {\bf Institutional review board (IRB) approvals or equivalent for research with human subjects}
    \item[] Question: Does the paper describe potential risks incurred by study participants, whether such risks were disclosed to the subjects, and whether Institutional Review Board (IRB) approvals (or an equivalent approval/review based on the requirements of your country or institution) were obtained?
    \item[] Answer: \answerNA{} 
    \item[] Justification: Not applicable as the paper does not involve human subjects.

\item {\bf Declaration of LLM usage}
    \item[] Question: Does the paper describe the usage of LLMs if it is an important, original, or non-standard component of the core methods in this research? Note that if the LLM is used only for writing, editing, or formatting purposes and does not impact the core methodology, scientific rigorousness, or originality of the research, declaration is not required.
    \item[] Answer: \answerNA{} 
    \item[] Justification: LLMs were not used as part of the research methodology; they were used only for writing assistance in non-substantive ways.

\end{enumerate}

\end{document}